\documentclass{article}

\PassOptionsToPackage{numbers,sort&compress}{natbib}

\usepackage[final]{neurips_2022}

\usepackage[dvipsnames]{xcolor}
\usepackage[colorlinks,linkcolor=black,citecolor=blue,urlcolor=blue,bookmarks=true]{hyperref}

\usepackage{amsmath,amsfonts,bm,mathtools}

\def\eqref#1{equation~\ref{#1}}

\def\1{\bm{1}}

\def\rvs{{\mathbf{s}}}

\def\rvv{{\mathbf{v}}}
\def\rvw{{\mathbf{w}}}
\def\rvx{{\mathbf{x}}}
\def\rvy{{\mathbf{y}}}
\def\rvz{{\mathbf{z}}}

\def\vmu{{\bm{\mu}}}
\def\vtheta{{\bm{\theta}}}
\def\vepsilon{{\bm{\epsilon}}}
\def\vpsi{{\bm{\psi}}}

\def\vf{{\bm{f}}}

\def\vk{{\bm{k}}}

\def\vo{{\bm{o}}}

\def\vs{{\bm{s}}}

\def\vv{{\bm{v}}}

\def\mI{{\bm{I}}}

\def\mS{{\bm{S}}}

\DeclareMathAlphabet{\mathsfit}{\encodingdefault}{\sfdefault}{m}{sl}
\SetMathAlphabet{\mathsfit}{bold}{\encodingdefault}{\sfdefault}{bx}{n}

\def\gN{{\mathcal{N}}}
\def\gO{{\mathcal{O}}}

\def\gU{{\mathcal{U}}}

\newcommand{\E}{\mathbb{E}}

\usepackage[utf8]{inputenc} %
\usepackage[T1]{fontenc}    %
\usepackage{hyperref}       %
\usepackage{url}            %
\usepackage{booktabs}       %
\usepackage{amsfonts}       %
\usepackage{nicefrac}       %
\usepackage{microtype}      %
\usepackage{xcolor}         %

\usepackage{cleveref}
\Crefname{figure}{Fig.}{Figs.}
\Crefname{table}{Tab.}{Tabs.}
\Crefname{section}{Sec.}{Secs.}
\Crefname{appendix}{App.}{Apps.}
\Crefname{equation}{Eq.}{Eqs.}
\Crefname{algorithm}{Alg.}{Algs.}

\usepackage{pifont}
\usepackage{tikz}
\usepackage{pgfplots}
\pgfplotsset{compat=1.17}
\newcommand{\cmark}{\ding{51}}
\newcommand{\xmark}{\ding{55}}
\usepackage{multirow}
\usepackage{subcaption}
\usepackage{wrapfig}
\usepackage{lipsum}
\usepackage{rotating}
\usepackage{algorithm}
\usepackage{algpseudocode}

\usepackage{pgfkeys}
\newenvironment{customlegend}[1][]{%
    \begingroup
    \csname pgfplots@init@cleared@structures\endcsname
    \pgfplotsset{#1}%
}{%
    \csname pgfplots@createlegend\endcsname
    \endgroup
}%
\def\addlegendimage{\csname pgfplots@addlegendimage\endcsname}

\definecolor{set21}{RGB}{102.0 194.0 165.0}
\definecolor{set22}{RGB}{252.0 141.0 98.0}
\definecolor{set23}{RGB}{141.0 160.0 203.0}
\definecolor{set24}{RGB}{231.0 138.0 195.0}
\definecolor{set25}{RGB}{166.0 216.0 84.0}
\definecolor{set26}{RGB}{255.0 217.0 47.0}
\definecolor{set27}{RGB}{229.0 196.0 148.0}
\definecolor{set28}{RGB}{179.0 179.0 179.0}

\tikzset{every picture/.style=semithick}

\definecolor{nvgreen}{HTML}{76B900}
\definecolor{myred}{HTML}{A03F77}

\newcommand{\ARASH}[1]{}
\newcommand{\KK}[1]{}
\newcommand{\TIM}[1]{}

\definecolor{nvgreen}{HTML}{76B900}
\definecolor{myred}{HTML}{A03F77}

\title{\textcolor{myred}{GENIE}: Hi\textcolor{myred}{g}her-Ord\textcolor{myred}{e}r De\textcolor{myred}{n}oising Diffus\textcolor{myred}{i}on Solv\textcolor{myred}{e}rs}

\author{Tim Dockhorn\textsuperscript{1,2,3,}\thanks{Work done during internship at NVIDIA.}
\qquad\qquad Arash Vahdat\textsuperscript{1}
\qquad\qquad Karsten Kreis\textsuperscript{1} \\
\\
{\textsuperscript{1}NVIDIA \quad \textsuperscript{2}University of Waterloo \quad \textsuperscript{3}Vector Institute \vspace{1pt}}
\\
\texttt{\scriptsize tim.dockhorn@uwaterloo.ca,} \quad \texttt{\scriptsize   \{avahdat,kkreis\}@nvidia.com}
}

\begin{document}

\maketitle

\begin{abstract}
Denoising diffusion models (DDMs) have emerged as a powerful class of generative models. A forward diffusion process slowly perturbs the data, while a deep model learns to gradually denoise. Synthesis amounts to solving a differential equation (DE) defined by the learnt model. Solving the DE requires slow iterative solvers for high-quality generation. In this work, we propose \emph{Higher-Order Denoising Diffusion Solvers} (GENIE): Based on truncated Taylor methods, we derive a novel higher-order solver that significantly accelerates synthesis. Our solver relies on higher-order gradients of the perturbed data distribution, that is, higher-order score functions. In practice, only Jacobian-vector products (JVPs) are required and we propose to extract them from the first-order score network via automatic differentiation. We then distill the JVPs into a separate neural network that allows us to efficiently compute the necessary higher-order terms for our novel sampler during synthesis. We only need to train a small additional head on top of the first-order score network. We validate GENIE on multiple image generation benchmarks and demonstrate that GENIE outperforms all previous solvers. Unlike recent methods that fundamentally alter the generation process in DDMs, our GENIE solves the true generative DE and still enables applications such as encoding and guided sampling.
Project page and code: \url{https://nv-tlabs.github.io/GENIE}.
\looseness=-1
\end{abstract}

\section{Introduction} \label{s:introcution}
Denoising diffusion models (DDMs) offer both state-of-the-art synthesis quality and sample diversity in combination with a robust and scalable learning objective. 
DDMs have been used for image~\cite{ho2020,ho2021arxiv,nichol21,dhariwal2021diffusion,rombach2021highresolution} and video~\cite{yang2022video,ho2022video} synthesis, super-resolution~\cite{saharia2021image,li2021srdiff}, deblurring~\cite{whang2021deblurring,kawar2022restoration}, image editing and inpainting~\cite{meng2021sdedit,lugmayr2022repaint,rombach2021highresolution,saharia2021palette}, text-to-image synthesis~\cite{nichol2021glide,avrahami2021blended,ramesh2022dalle2}, conditional and semantic image generation~\cite{choi2021ilvr,pandey2022diffusevae,preechakul2021diffusion,liu2021more,batzolis2021conditional}, image-to-image translation~\cite{sasaki2021unitddpm,saharia2021palette,su2022dual} and for inverse problems in medical imaging~\cite{song2022solving,peng2022towards,xie2022measurement,luo2022mri,chung2021scorebased,hu2022unsupervised,chung2021comecloser}. They also enable high-quality speech synthesis~\cite{chen2021wavegrad,kong2021diffwave,jeong2021difftts,chen2021wavegrad2,popov2021gradtts,liu2022diffgantts}, 3D shape generation~\cite{luo2021diffusion,cai2020eccv,zhou20213d,lyu2022a,dupont2022from}, molecular modeling~\cite{shi2021learning,xu2022geodiff,xie2022crystal,jo2022scorebased}, maximum likelihood training~\cite{song2021maximum,kingma2021variational,vahdat2021score,huang2021perspective}, and more~\cite{niu2020permutation,yoon2021advpur,tashiro2021csdi,bortoli2022riemannian,sanchez2022diffusion,nie2022DiffPure}. In DDMs, a diffusion process gradually perturbs the data towards random noise, while a deep neural network learns to denoise. Formally, the problem reduces to learning the \textit{score function}, i.e., the gradient of the log-density of the perturbed data. 
The (approximate) inverse of the forward diffusion can be described by an ordinary or a stochastic differential equation (ODE or SDE, respectively), defined by the learned score function, and can therefore be used for generation when starting from random noise~\cite{song2020,song2021maximum}.

\begin{figure}
    \centering
    \vspace{-0.9cm}
    \includegraphics[width=0.90\textwidth]{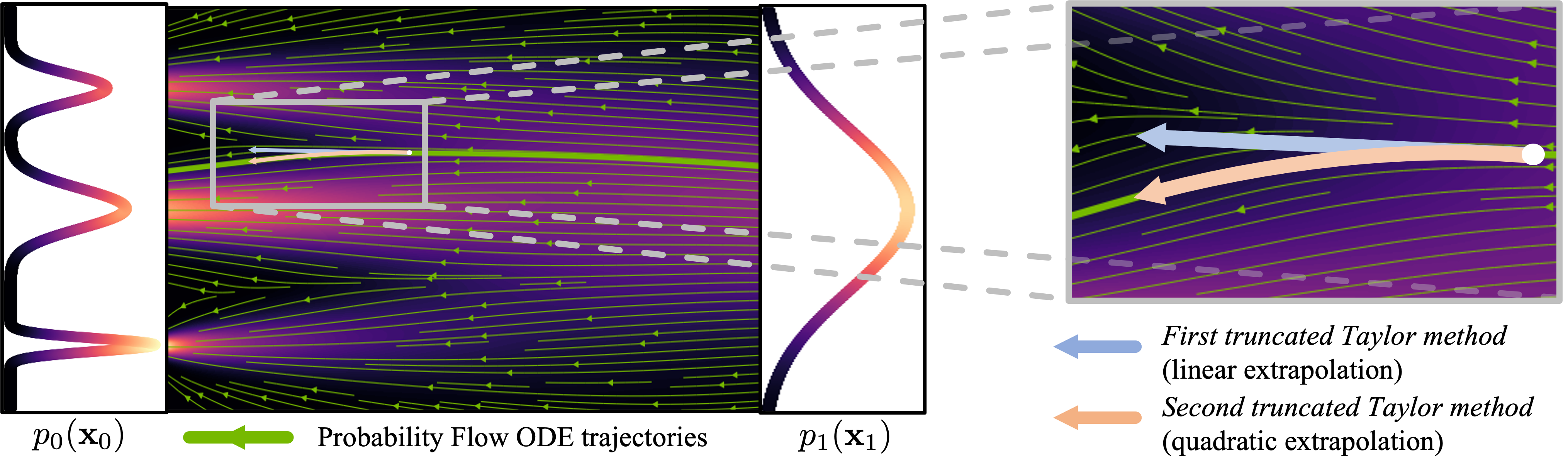}
    \vspace{-1mm}
    \caption{\small Our novel \textit{Higher-Order Denoising Diffusion Solver} (GENIE) relies on the second \textit{truncated Taylor method} (TTM) to simulate a (re-parametrized) Probability Flow ODE for sampling from denoising diffusion models. The second TTM captures the local curvature of the ODE's gradient field and enables more accurate extrapolation and larger step sizes than the first TTM (Euler's method), which previous methods such as DDIM~\cite{song2021denoising} utilize.\looseness=-1}
    \vspace{-5.5mm}
    \label{fig:pipeline}
\end{figure}

A crucial drawback of DDMs is that the generative ODE or SDE is typically difficult to solve, due to the complex score function. 
Therefore, efficient and tailored samplers are required for fast synthesis. In this work, building on the generative ODE~\cite{song2020,song2021denoising,song2021maximum}, we rigorously derive a novel second-order ODE solver using \emph{truncated Taylor methods}~\citep{kloeden1992euler}. These higher-order methods require higher-order gradients of the ODE---in our case this includes higher-order gradients of the log-density of the perturbed data, i.e., higher-order score functions. Because such higher-order scores are usually not available, existing works typically use simple first-order solvers or samplers with low accuracy~\cite{ho2020,song2020,song2021denoising,dockhorn2022scorebased}, higher-order methods that rely on suboptimal finite difference or other approximations~\cite{tachibana2021taylor,jolicoeur2021gotta,liu2022pseudo}, or alternative approaches~\cite{kong2021arxiv,bao2022analyticdpm,watson2022learning} for accelerated sampling. Here, we fundamentally avoid such approximations and directly model the higher-order gradient terms: Importantly, our novel \emph{Higher-Order Denoising Diffusion Solver (GENIE)} relies on Jacobian-vector products (JVPs) involving second-order scores. We propose to calculate these JVPs by automatic differentiation of the regular learnt first-order scores. For computational efficiency, we then distill the entire higher-order gradient of the ODE, including the JVPs, into a separate neural network. In practice, we only need to add a small head to the first-order score network to predict the components of the higher-order ODE gradient.
By directly modeling the JVPs
we avoid explicitly forming high-dimensional higher-order scores. Intuitively, the higher-order terms in GENIE capture the local curvature of the ODE and enable larger steps when iteratively solving the generative ODE (Fig.~\ref{fig:pipeline}). %

Experimentally, we validate GENIE on multiple image modeling benchmarks and achieve state-of-the-art performance in solving the generative ODE of DDMs with few synthesis steps. In contrast to recent methods that fundamentally modify the generation process of DDMs by training conditional GANs~\cite{xiao2022tackling} or by distilling the full sampling trajectory~\cite{luhman2021knowledge,salimans2022progressive}, GENIE solves the true generative ODE. Therefore, we also show that we can still encode images in the DDM's latent space, as required for instance for image interpolation, and use techniques such as guided sampling~\cite{song2020,dhariwal2021diffusion,ho2021classifierfree}.

We make the following contributions: \textbf{(i)} We introduce GENIE, a novel second-order ODE solver for fast DDM sampling. \textbf{(ii)} We propose to extract the required higher-order terms from the first-order score model by automatic differentiation. In contrast to existing works, we explicitly work with higher-order scores without finite difference approximations. To the best of our knowledge, GENIE is the first method that \textit{explicitly} uses higher-order scores for generative modeling with DDMs. 
\textbf{(iii)} We propose to directly model the necessary JVPs and distill them into a small neural network.
\textbf{(iv)}~We outperform all previous solvers and samplers for the generative differential equations of DDMs.\looseness=-1 %
\vspace{-2.5mm}
\section{Background} \label{sec:background}
\vspace{-2mm}
We consider continuous-time DDMs~\citep{song2020, sohl2015, ho2020}
whose forward process can be described by 
\begin{align}
    p_t(\rvx_t|\rvx_0) = \gN(\rvx_t; \alpha_t \rvx_0, \sigma_t^2 \mI),
\end{align}
where $\rvx_0 \sim p_0(\rvx_0)$ is drawn from the empirical data distribution and $\rvx_t$ refers to diffused data samples at time $t\in[0,1]$ along the diffusion process. The functions $\alpha_t$ and $\sigma_t$ are generally chosen such that the \emph{logarithmic signal-to-noise ratio}~\citep{kingma2021variational} $\log \frac{\alpha_t^2}{\sigma_t^2}$ decreases monotonically with $t$ and the data diffuses towards random noise, i.e., $p_1(\rvx_1)\!\approx\!\gN(\rvx_1;\bm{0}, \mI)$. We use \emph{variance-preserving}~\citep{song2020} diffusion processes for which $\sigma_t^2 = 1 - \alpha_t^2$ (however, all methods introduced in this work are applicable to more general DDMs).
The diffusion process can then be expressed by the (variance-preserving) SDE
\begin{align}
    d\rvx_t &= - \tfrac{1}{2} \beta_t \rvx_t \, dt + \sqrt{\beta_t} \, d\rvw_t,
\end{align}
where $\beta_t = -\frac{d}{dt} \log \alpha_t^2$, $\rvx_0 \sim p_0(\rvx_0)$ and $\rvw_t$ is a standard Wiener process. A corresponding reverse diffusion process that effectively inverts the forward diffusion is given by~\cite{anderson1982,haussmann1986time,song2020}
\begin{align} \label{eq:probability_flow_sde}
    d\rvx_t &= -\tfrac{1}{2} \beta_{t} \left[\rvx_t + 2\nabla_{\rvx_t} \log p_t(\rvx_t) \right]dt + \sqrt{\beta_t}\, d\rvw_t,
\end{align}
and this reverse-time generative SDE is marginally equivalent to the generative ODE~\citep{song2020,song2021maximum}
\begin{align} \label{eq:probability_flow_ode}
    d\rvx_t &= -\tfrac{1}{2} \beta_{t} \left[\rvx_t + \nabla_{\rvx_t} \log p_t(\rvx_t) \right]dt,
\end{align}
where $\nabla_{\rvx_t} \log p_t(\rvx_t)$ is the \emph{score function}. \Cref{eq:probability_flow_ode} is referred to as the \emph{Probability Flow} ODE~\citep{song2020}, an instance of continuous Normalizing flows~\citep{chen2018neuralODE, grathwohl2019ffjord}. To generate samples from the DDM, one can sample $\rvx_1 \sim \gN(\rvx_1; \bm{0}, \mI)$ and numerically simulate either the Probability Flow ODE or the generative SDE, replacing the unknown score function by a learned score model $\vs_\vtheta(\rvx_t, t) \approx \nabla_{\rvx_t} \log p_t(\rvx_t)$.

The DDIM solver~\citep{song2021denoising} has been particularly popular to simulate DDMs due to its speed and simplicity. It has been shown that DDIM is \emph{Euler's method} applied to an ODE based on a re-parameterization of the Probability Flow ODE~\citep{salimans2022progressive, song2021denoising}:
Defining $\gamma_t = \sqrt{\frac{1 - \alpha_t^2}{\alpha_t^2}}$ and $\bar \rvx_t = \rvx_t \sqrt{1 + \gamma_t^2}$, we have
\begin{align}
    \frac{d\bar \rvx_t}{d\gamma_t} = \sqrt{1+\gamma^2_t} \frac{d\rvx_t}{dt} \frac{dt}{d\gamma_t} + \rvx_t \frac{\gamma_t}{\sqrt{1+\gamma^2_t}} =  - \frac{\gamma_t}{\sqrt{1 + \gamma_t^2}} \nabla_{\rvx_t} \log p_t(\rvx_t),
\end{align}
where we inserted \Cref{eq:probability_flow_ode} for $\frac{d\rvx_t}{dt}$ and used $\beta(t) \frac{dt}{d\gamma_t} = \frac{2\gamma_t}{\gamma_t^2 + 1}$.
Letting $\vs_\vtheta(\rvx_t, t) \coloneqq - \frac{\vepsilon_\vtheta(\rvx_t, t)}{\sigma_t}$ denote a parameterization of the score model, the approximate generative DDIM ODE is then given by
\begin{align} \label{eq:ddim_ode}
    d \bar \rvx_t &= \vepsilon_\vtheta\left(\rvx_t, t\right)d\gamma_t,
\end{align}
where we used $\sigma_t = \sqrt{1-\alpha_t^2} = \frac{\gamma_t}{\sqrt{\gamma_t^2 + 1}}$ (see~\Cref{s:app_ddim_ode} for a more detailed derivation of~\Cref{eq:ddim_ode}). 
The model $\vepsilon_\vtheta(\rvx_t, t)$
can be learned by minimizing the score matching objective~\cite{vincent2011,ho2020}
\begin{align} \label{eq:score_objective}
    \min_\vtheta\, \E_{t \sim \gU[t_\mathrm{cutoff}, 1], \rvx_0 \sim p(\rvx_0), \vepsilon \sim \gN(\bm{0}, \mI)} \left[g(t) \| \vepsilon - \vepsilon_\vtheta(\rvx_t, t) \|_2^2\right], \quad \rvx_t = \alpha_t \rvx_0 + \sigma_t \vepsilon,
\end{align}
for small $0<t_\mathrm{cutoff}\ll1$. As is standard practice, we set $g(t) = 1$. Other weighting functions $g(t)$ are possible; for example, setting $g(t) = \frac{\beta_t}{2 \sigma_t^2}$ recovers maximum likelihood learning~\citep{song2021maximum,kingma2021variational,vahdat2021score,huang2021perspective}.
\vspace{-2mm}
\section{Higher-Order Denoising Diffusion Solver} \label{s:method}
\vspace{-2mm}
As discussed in~\Cref{sec:background}, the so-known DDIM solver~\citep{song2021denoising} is simply Euler's method applied to the DDIM ODE (cf.~\Cref{eq:ddim_ode}). %
In this work, we apply a higher-order method to the DDIM ODE, building on the \emph{truncated Taylor method}~(TTM)~\cite{kloeden1992euler}. The $p$-th TTM is simply the $p$-th order~\emph{Taylor polynomial} applied to an ODE. For example, for the general $\frac{d\rvy}{dt} = \vf(\rvy, t)$, the $p$-th TTM reads as 
\begin{align}\label{eq:p_ttm}
    \rvy_{t_{n+1}} = \rvy_{t_n} + h_n \frac{d\rvy}{dt} \rvert_{(\rvy_{t_n}, t_n)} + \cdots + \frac{1}{p!} h_n^p \frac{d^p\rvy}{dt^p} \rvert_{(\rvy_{t_n}, t_n)},
\end{align}
where $h_n = t_{n+1} - t_n$ (see \Cref{app:theoreticalbounds} for a truncation error analysis with respect to the exact ODE solution). Note that the first TTM is simply Euler's method. Applying the second TTM to the DDIM ODE results in the following scheme:
\begin{align} \label{eq:hsds}
    \bar \rvx_{t_{n+1}} = \bar \rvx_{t_n} + h_n \vepsilon_\vtheta(\rvx_{t_n}, t_n) + \frac{1}{2} h_n^2 \frac{d\vepsilon_\vtheta}{d\gamma_t} \rvert_{(\rvx_{t_n}, t_n)},
\end{align}
where $h_n = \gamma_{t_{n+1}} - \gamma_{t_n}$. Recall that $\gamma_t = \sqrt{\frac{1-\alpha_t^2}{\alpha_t^2}}$, where the function $\alpha_t$ is a time-dependent hyperparameter of the DDM.
The total derivative $d_{\gamma_t} \vepsilon_\vtheta \coloneqq \frac{d\vepsilon_\vtheta}{d\gamma_t}$ can be decomposed as follows
\begin{align} \label{eq:total_derivative}
    d_{\gamma_t} \vepsilon_\vtheta(\rvx_t, t) = \frac{\partial \vepsilon_\vtheta(\rvx_t, t)}{\partial \rvx_t} \frac{d\rvx_t}{d \gamma_t} + \frac{\partial \vepsilon_\vtheta(\rvx_t, t)}{\partial t} \frac{dt}{d\gamma_t},
\end{align}
where $\tfrac{\partial \vepsilon_\vtheta(\rvx_t, t)}{\partial \rvx_t}$ denotes the Jacobian of $\vepsilon_\vtheta(\rvx_t, t)$ and
\begin{align} \label{eq:space_derivative}
    \frac{d\rvx_t}{d\gamma_t} = \frac{\partial \rvx_t}{\partial \bar \rvx_t} \frac{d\bar \rvx_t}{d\gamma_t} + \frac{\partial \rvx_t}{\partial \gamma_t} = \frac{1}{\sqrt{\gamma_t^2 + 1}} \vepsilon_\vtheta(\rvx_t, t) - \frac{\gamma_t}{1 + \gamma_t^2} \rvx_t.
\end{align}
If not explicitly stated otherwise, we refer to the second TTM applied to the DDIM ODE, i.e., the scheme in~\Cref{eq:hsds}, as \textit{Higher-Order Denoising Diffusion Solver} (GENIE). Intuitively, the higher-order gradient terms used in the second TMM model the local curvature of the ODE. This translates into a Taylor formula-based extrapolation that is quadratic in time (cf.~\Cref{eq:p_ttm,eq:hsds}) and more accurate than linear extrapolation, as in Euler's method, thereby enabling larger time steps (see~\Cref{fig:pipeline} for a visualization). In~\Cref{s:app_ttm}, we also discuss the application of the third TTM to the DDIM ODE. We emphasize that TTMs are not restricted to the DDIM ODE and could just as well be applied to the Probability Flow ODE~\cite{song2020} (also see~\Cref{s:app_ttm}) or neural ODEs~\cite{chen2018neuralODE,grathwohl2019ffjord} more generally.

\begin{figure}
\vspace{-0.6cm}
    \captionsetup[subfigure]{aboveskip=-1pt,belowskip=-1pt}
    \centering
    \begin{subfigure}{0.32\textwidth}
        \centering
        \includegraphics[scale=1.]{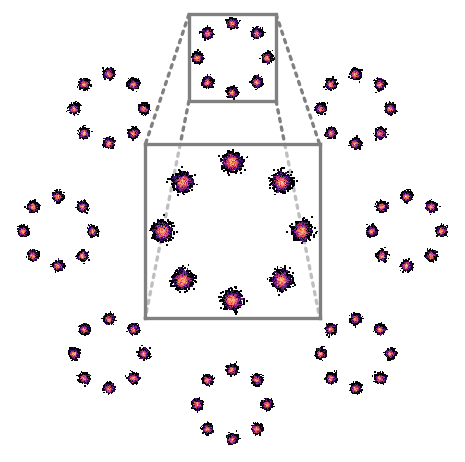}
        \caption{Ground truth}
        \label{fig:toy_distribution_ground_truth}
    \end{subfigure}
    \begin{subfigure}{0.32\textwidth}
        \centering
        \includegraphics[scale=1.]{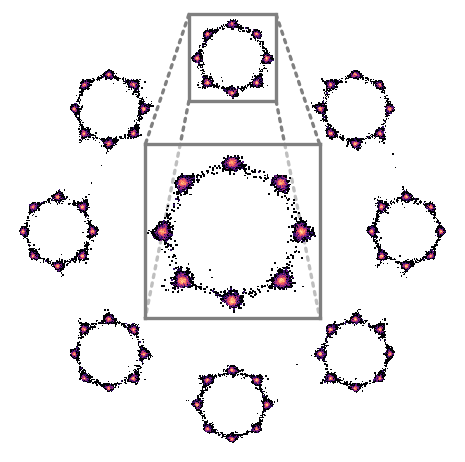}
        \caption{DDIM}
    \end{subfigure}
    \begin{subfigure}{0.32\textwidth}
        \centering
        \includegraphics[scale=1.0]{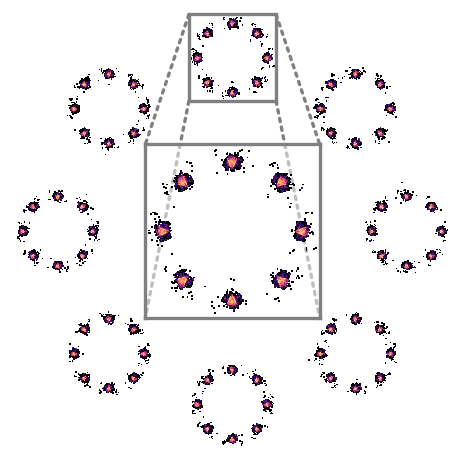}
        \caption{GENIE}
    \end{subfigure}
    \vspace{-0.1cm}
    \caption{\footnotesize Modeling a complex 2D toy distribution: Samples in \textit{(b)} and \textit{(c)} are generated via DDIM and GENIE, respectively, with 25 solver steps using the analytical score function of the ground truth distribution.}
    \vspace{-0.5cm}
    \label{fig:toy_distribution}
\end{figure}
\begin{wrapfigure}{r}{0.3\textwidth}
\vspace{-.1cm}
  \begin{center}
    \includegraphics[scale=0.41, clip=True]{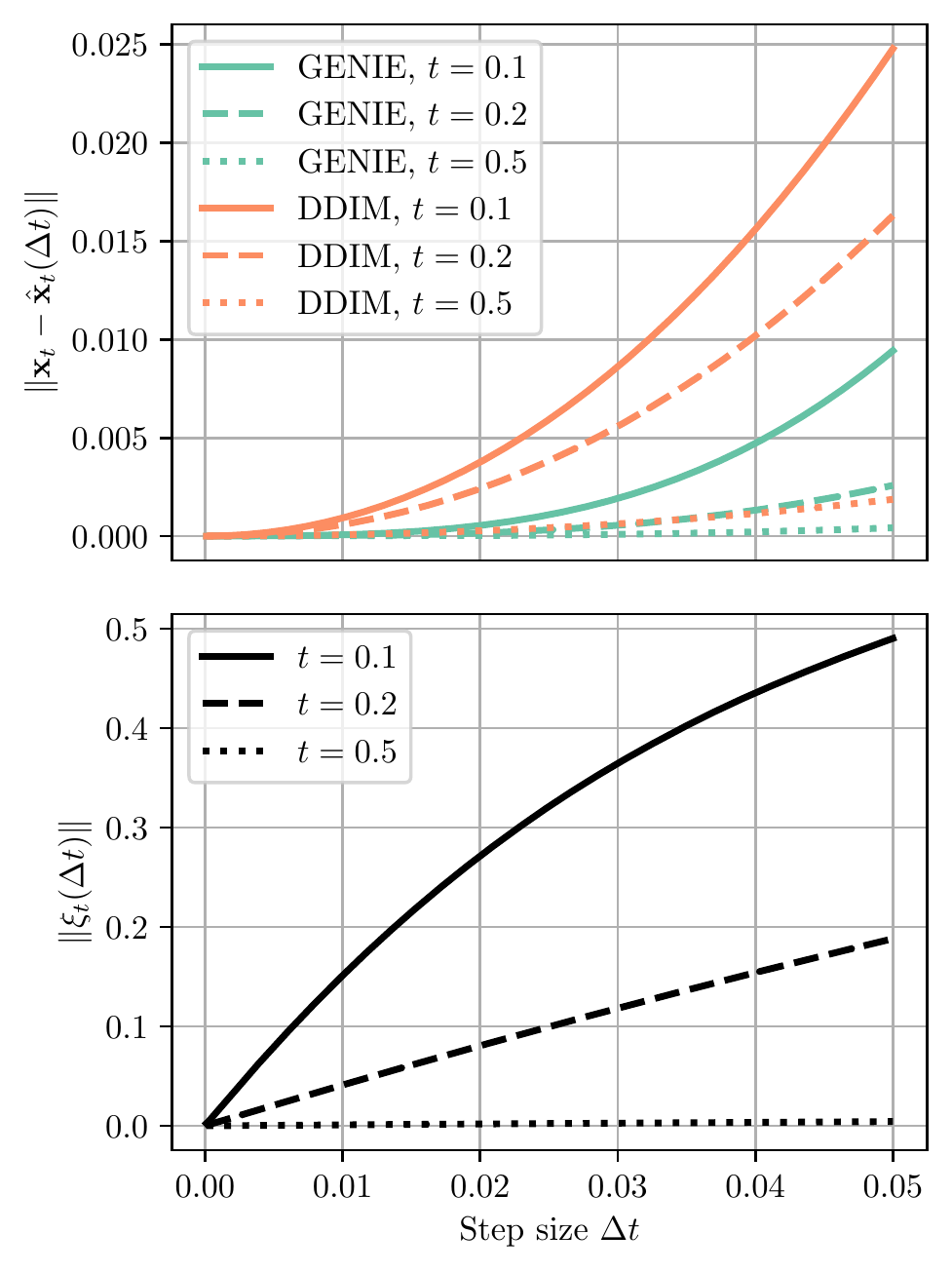}
  \end{center}
  \vspace{-0.4cm}
  \caption{\small \textit{Top}: Single step error using analytical score function. \textit{Bottom}: Norm of difference $\xi_t(\Delta t)$ between analytical and approximate derivative computed via finite difference method.\looseness=-1}
  \vspace{-0.6cm}
  \label{fig:analytical_error}
\end{wrapfigure}

\noindent \textbf{The Benefit of Higher-Order Methods:}
We showcase the benefit of higher-order methods on a 2D toy distribution~(\Cref{fig:toy_distribution_ground_truth}) for which we know the score function as well as all higher-order derivatives necessary for GENIE analytically. We generate 1k different accurate ``ground truth'' trajectories $\rvx_t$ using DDIM with 10k steps. We compare these ``ground truth'' trajectories to \emph{single} steps of DDIM and GENIE for varying step sizes $\Delta t$.
We then measure the mean $L_2$-distance of the single steps $\hat \rvx_t(\Delta t)$ to the ``ground truth'' trajectories $\rvx_t$, and we repeat this experiment for three starting points $t \in \{0.1, 0.2, 0.5\}$. We see (\Cref{fig:analytical_error} (\textit{top})) that GENIE can use larger step sizes to stay within a certain error tolerance for all starting points $t$. We further show samples for DDIM and GENIE, using 25 solver steps, in~\Cref{fig:toy_distribution}. DDIM has the undesired behavior of sampling low-density regions between modes, whereas GENIE looks like a slightly noisy version of the ground truth distribution~(\Cref{fig:toy_distribution_ground_truth}).

\noindent \textbf{Comparison to Multistep Methods:} Linear multistep methods are an alternative higher-order method to solve ODEs. \citet{liu2022pseudo} applied the well-established Adams--Bashforth~\citep[AB,][]{butcher2016numerical} method to the DDIM ODE. AB methods can be derived from TTMs by approximating higher-order derivatives $\frac{d^p \rvy}{dt^p}$ using the finite difference method~\citep{kloeden1992stochastic}. For example, the second AB method is obtained from the second TTM by replacing $\frac{d^2 \rvy}{dt^2}$ with the first-order forward difference approximation ${(f(\rvy_{t_n}, t_n) - f(\rvy_{t_{n-1}}, t_{n-1})})/{h_{n-1}}$. In~\Cref{fig:analytical_error} (\textit{bottom}), we visualize the mean $L_2$-norm of the difference $\xi_t(\Delta t)$ between the analytical derivative $d_{\gamma_t} \vepsilon_\vtheta$ and its first-order forward difference approximation for varying step sizes $\Delta t$ for the 2D toy distribution. The approximation is especially poor at small $t$ for which the score function becomes complex (\Cref{s:app_toy_experiments} for details on all toy experiments).\looseness=-1
\vspace{-2mm}
\subsection{Learning Higher-Order Derivatives}
\vspace{-2mm}
\label{s:learning_higher_order_derivatives}
The above observations inspire to apply GENIE to DDMs of more complex and high-dimensional data such as images. Regular DDMs learn a model $\vepsilon_\vtheta$ for the first-order score; however, the higher-order gradient terms required for GENIE (cf.~\Cref{eq:total_derivative}) are not immediately available to us, unlike in the toy example above. Let us insert~\Cref{eq:space_derivative} into~\Cref{eq:total_derivative} and analyze the required terms more closely:\looseness=-1
\vspace{-0.1cm}
\begin{align} \label{eq:total_derivative_full}
    d_{\gamma_t} \vepsilon_\vtheta(\rvx_t, t) = \frac{1}{\sqrt{\gamma_t^2 + 1}} \underbrace{\frac{\partial \vepsilon_\vtheta(\rvx_t, t)}{\partial \rvx_t} \vepsilon_\vtheta(\rvx_t, t)}_{\textrm{JVP}_1} - \frac{\gamma_t}{1 + \gamma_t^2} \underbrace{\frac{\partial \vepsilon_\vtheta(\rvx_t, t)}{\partial \rvx_t} \rvx_t}_{\textrm{JVP}_2} + \frac{\partial \vepsilon_\vtheta(\rvx_t, t)}{\partial t} \frac{dt}{d\gamma_t}.
\end{align}
We see that the full derivative decomposes into two JVP terms and one simpler time derivative term. The term $\frac{\partial \vepsilon_\vtheta(\rvx_t, t)}{\partial \rvx_t}$ plays a crucial role in~\Cref{eq:total_derivative_full}. It can be expressed as 
\vspace{-0.1cm}
\begin{align} \label{eq:second_order_score_function}
    \frac{\partial \vepsilon_\vtheta(\rvx_t, t)}{\partial \rvx_t}=-\sigma_t\frac{\partial \rvs_\vtheta(\rvx_t,t)}{\partial \rvx_t}\approx -\sigma_t \nabla_{\rvx_t}^\top \nabla_{\rvx_t} \log p_t(\rvx_t),
\end{align}
which means that GENIE relies on \textit{second-order score functions} $\nabla_{\rvx_t}^\top \nabla_{\rvx_t} \log p_t(\rvx_t)$ under the hood. %

Given a DDM, that is, given $\vepsilon_\vtheta$, we could compute the derivative $d_{\gamma_t} \vepsilon_\vtheta$ for the GENIE scheme in~\Cref{eq:hsds} using automatic differentiation~(AD). This would, however, make a single step of GENIE at least twice as costly as DDIM, because we would need a forward pass through the $\vepsilon_\vtheta$ network to compute $\vepsilon_\vtheta(\rvx_t, t)$ itself, and another pass to compute the JVPs and the time derivative in~\Cref{eq:total_derivative_full}.
These forward passes cannot be parallelized, since the vector-part of $\textrm{JVP}_1$ in~\Cref{eq:total_derivative_full} involves $\vepsilon_\vtheta$ itself, and needs to be known before computing the JVP.
To accelerate sampling,
this overhead is too expensive.\looseness=-1

\begin{wrapfigure}{r}{0.47\textwidth}
\vspace{-0.3cm}
  \begin{center}
    \includegraphics[scale=0.2]{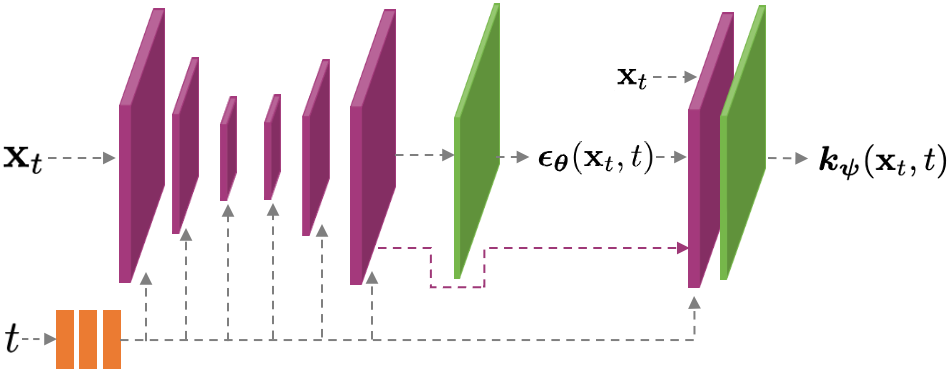}
  \end{center}
  \caption{\small Our distilled model $\vk_\vpsi$ that predicts the gradient $d_{\gamma_t} \vepsilon_\vtheta$ is implemented as a small additional output head on top of the first-order score model $\vepsilon_\vtheta$. Purple layers are used both in $\vepsilon_\vtheta$ and $\vk_\vpsi$; green layers are specific for $\vepsilon_\vtheta$ and $\vk_\vpsi$.}
  \label{fig:blocks}
  \vspace{-0.3cm}
\end{wrapfigure}
\vspace{-1mm}
\textbf{Gradient Distillation:} To avoid this 
overhead,
we propose to first distill $d_{\gamma_t} \vepsilon_\vtheta$ into a separate neural network. During distillation training, we can use the slow AD-based calculation of $d_{\gamma_t} \vepsilon_\vtheta$, but during synthesis we call the trained neural network. 
We build on the observation that the internal representations of the neural network modeling $\vepsilon_\vtheta$ (in our case a U-Net~\cite{ronneberger2015u} architecture) can be used for downstream tasks~\citep{choi2022perception, baranchuk2022labelefficient}: specifically, we provide the last feature layer from the $\vepsilon_\vtheta$ network together with its time embedding as well as $\rvx_t$ and the output $\vepsilon_\vtheta(\rvx_t,t)$ to a small prediction head $\vk_\vpsi(\rvx_t,t)$ that models the different terms in~\Cref{eq:total_derivative_full} (see~\Cref{fig:blocks}). 
The overhead generated by $\vk_\vpsi$ is small, for instance less than 2\% for our CIFAR-10 model (also see~\Cref{sec:experiments}), and we found this approach to provide excellent performance. Note that in principle we could also train an independent deep neural network, which does not make use of the internal representations of $\vepsilon_\vtheta$ and could therefore theoretically be run in parallel to the $\vepsilon_\vtheta$ model.
We justify using small prediction heads over independent neural networks because AD-based distillation training is slow:
in each training iteration we first need to call the $\vepsilon_\vtheta$ network, then calculate the JVP terms, and only then can we call the distillation model. By modeling $d_{\gamma_t} \vepsilon_\vtheta$ via small prediction heads, while reusing the internal representation of the score model, we can make training relatively fast: we only need to train $\vk_\vpsi$ for up to 50k iterations. In contrast, training score models from scratch takes roughly an order of magnitude more iterations. We leave training of independent networks to predict $d_{\gamma_t} \vepsilon_\vtheta$ to future work.\looseness=-1

\vspace{-1mm}
\textbf{Mixed Network Parameterization:} We found that learning $d_{\gamma_t} \vepsilon_\vtheta$ directly as single output of a neural network can be challenging. Assuming a single data point distribution $p_0(\rvx_0) = \delta(\rvx_0=\mathbf{0})$, for which we know the diffused score function and all higher-order derivatives analytically, we found that the terms in~\Cref{eq:total_derivative_full} all behave very differently within the $t \in [0, 1]$ interval (for instance, the prefactor of $\textrm{JVP}_1$ in~\Cref{eq:total_derivative_full} approaches $1$ as $t\rightarrow0$, while $\textrm{JVP}_2$'s prefactor vanishes). As outlined in detail in~\Cref{s:app_mixed_network_parameterization}, this simple single data point assumption implies an effective \textit{mixed network parameterization}, an approach inspired by the ``mixed score parametrizations'' in~\citet{vahdat2021score} and \citet{dockhorn2022scorebased}. In particular, we model
\vspace{-0.1cm}
\begin{align} \label{eq:mixed_score}
    \vk_\vpsi = -\frac{1}{\gamma_t} \vk^{(1)}_\vpsi + \frac{\gamma_t}{1 + \gamma_t^2} \vk^{(2)}_\vpsi + \frac{1}{\gamma_t (1 + \gamma_t^2)} \vk^{(3)}_\vpsi \approx d_{\gamma_t}\vepsilon_\vtheta,
\end{align}
where $\vk_\vpsi^{(i)}(\rvx_t,t)$, $i\in \{1,2,3\}$, are different output channels of the neural network (i.e. the additional head on top of the $\vepsilon_\vtheta$ network). The three terms in~\Cref{eq:mixed_score} exactly correspond to the three terms of~\Cref{eq:total_derivative_full}, in the same order.
We 
show the superior performance of this parametrization 
in~\Cref{s:ablations}.

\vspace{-1mm}
\textbf{Learning Objective:} Ideally, we would like our model $\vk_\vpsi$ to match $d_{\gamma_t}\vepsilon_\vtheta$ exactly, for all $t \in [0, T]$ and $\rvx_t$ 
in the diffused data distribution, which the generative ODE trajectories traverse. This suggests a simple (weighted) $L_2$-loss, similar to regular score matching losses for DDMs~\cite{ho2020,song2020}: 
\begin{align} \label{eq:objective}
    \min_\vpsi \E_{t \sim \gU[t_\mathrm{cutoff}, 1], \rvx_0 \sim p(\rvx_0), \vepsilon \sim \gN(\bm{0}, \mI)} \left[g_\mathrm{d}(t) \| \vk_\psi(\alpha_t \rvx_0 + \sigma_t \vepsilon, t) - d_{\gamma_t}\vepsilon_\vtheta(\alpha_t \rvx_0 + \sigma_t \vepsilon, t) \|_2^2\right]
\end{align}
for diffused data points 
$\alpha_t \rvx_0 + \sigma_t \vepsilon$
and 
$g_\mathrm{d}(t) = \gamma_t^2$ to counteract the 
$1/\gamma_t$ 
in the first and third terms of 
\Cref{eq:mixed_score}. 
This leads to a roughly constant loss over different time values $t$. During training we 
compute $d_{\gamma_t}\vepsilon_\vtheta$ via AD;
however, at inference time we use the learned prediction head $\vk_\vpsi$ to approximate $d_{\gamma_t}\vepsilon_\vtheta$. %
In~\Cref{s:app_ph_pseudocode}, we provide pseudo code for training and sampling with
heads $\vk_\vpsi$. 
Note that our distillation objective is consistent and principled: if $\vk_\psi$ matches $d_{\gamma_t}\vepsilon_\vtheta$
exactly, the resulting GENIE algorithm recovers the second TTM \emph{exactly} (extended discussion in App.~\ref{app:consistency}). 

\vspace{-1mm}
\textbf{Alternative Learning Approaches:} As shown in~\Cref{eq:second_order_score_function}, GENIE relies on second-order score functions. Recently, \citet{meng2021estimating} directly learnt such higher-order scores with higher-order score matching objectives. Directly applying these techniques has the downside that we would need to explicitly form the higher-order score terms $\nabla^\top_{ \rvx_t}\vepsilon_\vtheta(\rvx_t, t)$, which are very high-dimensional for data such as images. Low-rank approximations are possible, but potentially insufficient for high performance. In our approach, we are avoiding this complication by directly modeling the lower-dimensional JVPs. We found that the methods from \citet{meng2021estimating} can be modified to provide higher-order score matching objectives for the JVP terms required for GENIE and we briefly explored this (see~\Cref{s:app_meng}). However, our distillation approach with AD-based higher-order gradients
worked much better. Nevertheless, this is an interesting direction for future research. To the best of our knowledge, GENIE is the first solver for the generative differential equations of DDMs that \textit{directly} uses higher-order scores (in the form of the distilled JVPs) for generative modeling without finite difference or other approximations.\looseness=-1

\vspace{-3mm}
\section{Related Work} \label{s:related_work}
\vspace{-3mm}
\textbf{Accelerated Sampling from DDMs.} Several previous works address the slow sampling of DDMs:
One line of work reduces and readjusts the timesteps~\cite{nichol21,kong2021arxiv} used in time-discretized DDMs~\citep{ho2020, sohl2015}. This can be done systematically by grid search~\citep{chen2021wavegrad} or dynamic programming~\citep{watson2021arxiv}.
\citet{bao2022analyticdpm} speed up sampling by defining a new DDM with optimal reverse variances. DDIM~\citep{song2021denoising}, discussed in Sec.~\ref{sec:background}, was also introduced as a method to accelerate DDM synthesis.
Further works leverage modern %
ODE and SDE solvers for fast synthesis from (continuous-time) DDMs:
For instance, higher-order Runge--Kutta methods~\citep{dormand1980odes,song2020} and adaptive step size SDE solvers~\citep{jolicoeur2021gotta} have been used. These methods are not optimally suited for the few-step synthesis regime, in which GENIE shines; see also~\Cref{sec:experiments}.
Most closely related to our work is~\citet{liu2022pseudo}, which simulates the DDIM ODE~\citep{song2021denoising} using a higher-order linear multistep method~\citep{butcher2016numerical}. 
As shown in~\Cref{s:method}, linear multistep methods can be considered an approximation of the TTMs used in GENIE. Furthermore,~\citet{tachibana2021taylor} solve the generative SDE via a higher-order Itô--Taylor method~\cite{kloeden1992euler} and in contrast to our work, they propose to use an ``ideal derivative trick'' to approximate higher-order score functions. In~\Cref{s:app_2nd_ttm_id}, we show that applying this ideal derivative approximation to the DDIM ODE does not have any effect: the ``ideal derivatives'' are zero by construction. 
Note that in GENIE, we in fact use the DDIM ODE, rather than, for example, the regular Probability Flow ODE~\citep{song2020}, as the base ODE for GENIE.
\looseness=-1

Alternatively, sampling from DDMs can also be accelerated via learning: For instance, \citet{watson2022learning} learn parameters of a generalized family of DDMs by optimizing for perceptual output quality.
\citet{luhman2021knowledge} and \citet{salimans2022progressive} distill a DDIM sampler into a student model, which enables sampling 
in as few as a single step. \citet{xiao2022tackling} replace DDMs' Gaussian samplers with expressive generative adversarial networks, similarly allowing for few-step synthesis. GENIE can also be considered a learning-based approach, as we distill a derivative of the generative ODE into a separate neural network. However, in contrast to the mentioned methods, GENIE still solves the true underlying generative ODE, which has major advantages: for instance, it can still be used easily for classifier-guided sampling~\cite{song2020,dhariwal2021diffusion,ho2021classifierfree} and to efficiently encode data into latent space---a prerequisite for likelihood calculation~\cite{song2020,song2021maximum} and editing applications~\cite{ramesh2022dalle2}. 
Note that the learnt sampler~\citep{watson2022learning} defines a proper probabilistic generalized DDM;
however, it isn't clear how it relates to the generative SDE or ODE and therefore how compatible the method is with applications such as classifier guidance.\looseness=-1

Other approaches to accelerate DDM sampling change the diffusion itself~\citep{dockhorn2022scorebased, nachmani2021non, lam2021bilateral} or train DDMs in the latent space of a Variational Autoencoder~\citep{vahdat2021score}. GENIE is complementary to these methods.
\looseness=-1

\textbf{Higher-Order ODE Gradients beyond DDMs.}
TTMs~\citep{kloeden1992stochastic} and other methods that leverage higher-order gradients are also applied outside the scope of DDMs. For instance, higher-order derivatives can play a crucial role when developing solvers~\citep{djeumou2022taylor} and regularization techniques~\citep{kelly2020learning,finlay2020howto} for neural ODEs~\citep{chen2018neuralODE,grathwohl2019ffjord}. Outside the field of machine learning, higher-order TTMs have been widely studied, for example, to develop solvers for stiff~\citep{chang1994atomft} and non-stiff~\citep{chang1994atomft, corliss1982solving} systems.

\textbf{Concurrent Works.} \citet{zhang2022fast} motivate the DDIM ODE from an exponential integrator perspective applied to the Probability Flow ODE and propose to apply existing solvers from the numerical ODE literature, namely, Runge--Kutta and linear multistepping, to the DDIM ODE directly. \citet{lu2022dpm} similarly recognize the semi-linear structure of the Probability Flow ODE, derive dedicated solvers, and introduce new step size schedulers to accelerate DDM sampling. \citet{karras2022elucidating} propose new fast solvers, both deterministic and stochastic, specifically designed for the differential equations arising in DDMs. Both \citet{qinsheng2022gddim} and \citet{karras2022elucidating} realize that the DDIM ODE has ``straight line solution trajectories'' for spherical normal data and single data points---this exactly corresponds to our derivation that the higher-order terms in the DDIM ODE are zero in such a setting (see \Cref{s:app_2nd_ttm_id}). \citet{bao2022estimating} learn covariance matrices for DDM sampling using prediction heads somewhat similar to the ones in GENIE; in~\Cref{s:app_bao}, we thoroughly discuss the differences between GENIE and the method proposed in~\citet{bao2022estimating}.\looseness=-1
\vspace{-3mm}
\section{Experiments} \label{sec:experiments}
\vspace{-2mm}
\textbf{Datasets:} We run experiments on five datasets: CIFAR-10~\cite{krizhevsky2009learning} (resolution 32), LSUN Bedrooms~\citep{yu2015lsun} (128), LSUN Church-Outdoor~\citep{yu2015lsun} (128), (conditional) ImageNet~\citep{deng2009imagenet} (64), and AFHQv2~\citep{choi2020stargan} (512). On AFHQv2 we only consider the subset of cats; referred to as ``Cats'' in the remainder of this work.\looseness=-1

\textbf{Architectures:} Except for CIFAR-10 (we use a \href{https://drive.google.com/file/d/16_-Ahc6ImZV5ClUc0vM5Iivf8OJ1VSif/view?usp=sharing}{checkpoint} by~\citet{song2020}), we train our own score models using 
architectures introduced by previous works~\citep{ho2020,dhariwal2021diffusion}. The architecture of our prediction heads is based on (modified) BigGAN residual blocks~\citep{brock2018large,song2020}. 
To minimize computational overhead, we only use a single residual block.
See~\Cref{s:app_model_details} for training and architecture details.

\textbf{Evaluation:} We 
measure sample quality via Fr\'echet Inception Distance~\citep[FID,][]{heusel2017gans} (see \Cref{s:app_eval_metrics}).

\textbf{Synthesis Strategy:} 
We simulate the DDIM ODE from $t{=}1$ up to 
$t{=}10^{-3}$ using evaluation times
following
a quadratic function
(\emph{quadratic striding}~\citep{song2021denoising}). For variance-preserving DDMs, it 
can be
beneficial to denoise the ODE solver output at the cutoff $t{=}10^{-3}$, i.e., $\rvx_0 = \frac{\rvx_t - \sigma_t \vepsilon_\vtheta(\rvx_t, t)}{\alpha_t}$~\citep{song2020, jolicoeur-martineau2021adversarial}. Note that the denoising step involves a score model evaluation, and therefore ``loses'' a function evaluation that could otherwise be used as an additional step in the ODE solver. 
To this end, 
denoising the output of the ODE solver is left as a hyperparameter of our synthesis strategy.\looseness=-1

\textbf{Analytical First Step (AFS):} 
Every additional neural network call becomes crucial in the low number of function evaluations (NFEs) regime. We found that we can improve the performance of GENIE and all other methods evaluated on our checkpoints by replacing the learned score with the (analytical) score of $\gN(\bm{0}, \mI) \approx p_{t=1}(\rvx_t)$ in the first step of the ODE solver. The ``gained'' function evaluation can then be used as an additional step in the ODE solver. Similarly to the denoising step mentioned above, AFS is treated as a hyperparameter of our~\textbf{Synthesis Strategy}. AFS details in \Cref{s:app_afs}.

\textbf{Accounting for Computational Overhead:} GENIE has a slightly increased computational overhead compared to other solvers due to the prediction head $\vk_\vpsi$. The computational overhead is increased by 1.47\%, 2.83\%, 14.0\%, and 14.4\% on CIFAR-10, ImageNet, LSUN Bedrooms, and LSUN Church-Outdoor, respectively (see also~\Cref{s:app_measuring_computational_overhead}). This additional overhead is always accounted for implicitly: we divide the NFEs by the computational overhead and round to the nearest integer. For example, on LSUN Bedrooms, we compare baselines with 10/15 NFEs to GENIE with 9/13 NFEs.
\vspace{-2mm}
\subsection{Image Generation}
\vspace{-2mm}
In~\Cref{fig:main_fid_figure} we compare our method to the most competitive baselines. In particular, on the same score model checkpoints%
, we compare GENIE with DDIM~\citep{song2021denoising}, S-PNDM~\citep{liu2022pseudo}, and F-PNDM~\citep{liu2022pseudo}. For these four methods, we only include the best result over the two hyperparameters discussed above, namely, the denoising step and AFS (see~\Cref{s:app_extended_quant} for tables with all results). We also include three competitive results from the literature~\citep{watson2022learning, kong2021arxiv, bao2022analyticdpm} that use different checkpoints and sampling strategies: for each method, we include the best result for their respective set of hyperparameters. We do not compare in this figure with Knowledge Distillation~\citep[KD,][]{luhman2021knowledge}, Progressive Distillation~\citep[PG,][]{salimans2022progressive} and Denoising Diffusion GANs~\citep[DDGAN,][]{xiao2022tackling} as they do not solve the generative ODE/SDE and use fundamentally different sampling approaches with drawbacks discussed in~\Cref{s:related_work}. 

For NFEs $\in\{10, 15, 20, 25\}$, GENIE outperforms all baselines (on the same checkpoint) on all four datasets (see detailed results in \Cref{s:app_extended_quant} and GENIE image samples in \Cref{s:app_extended_qual}). On CIFAR-10 and (conditional) ImageNet, GENIE also outperforms these baselines for NFEs=5, whereas DDIM outperforms GENIE slightly on the LSUN datasets (see tables in~\Cref{s:app_extended_quant}). %
GENIE also performs better than the three additional baselines from the literature (which use different checkpoints and sampling strategies) with the exception of the Learned Sampler~\citep[LS,][]{watson2022learning} on LSUN Bedrooms for NFEs=20. Though LS uses a learned striding schedule on LSUN Bedrooms (whereas GENIE simply uses quadratic striding), the LS's advantage is most likely due to the different checkpoint. In~\Cref{tab:main_cifar_10}, we investigate the effect of optimizing the striding schedule, via learning (LS) or grid search (DDIM \& GENIE), on CIFAR-10 and find that its significance decreases rapidly with increased NFEs (also see~\Cref{s:app_extended_quant} for details). In~\Cref{tab:main_cifar_10}, we also show additional baseline results; however, we do not include commonly-used adaptive step size solvers in~\Cref{fig:main_fid_figure}, as they are arguably not well-suited for this low NFE regime: for example, on the same CIFAR-10 checkpoint we use for GENIE, the adaptive SDE solver introduced in~\citet{jolicoeur2021gotta} obtains an FID of 82.4 at 48 NFEs. Also on the same checkpoint, the adaptive Runge--Kutta 4(5)~\citep{dormand1980odes} method applied to the ProbabilityFlow ODE achieves an FID of 13.1 at 38 NFEs (solver tolerances set to $10^{-2}$).

The results in~\Cref{fig:main_fid_figure} suggest that higher-order gradient information, as used in GENIE, can be efficiently leveraged for image synthesis. Despite using small prediction heads 
our distillation seems to be sufficiently accurate: for reference, replacing the distillation heads with the
derivatives computed via AD, we obtain FIDs of 9.22, 4.11, 3.54, 3.46 using 10, 20, 30, and 40 NFEs, respectively (NFEs adjusted assuming an additional computational overhead of 100\%). 
As discussed in~\Cref{s:method}, linear multistep methods such as S-PNDM~\citep{liu2022pseudo} and F-PNDM~\citep{liu2022pseudo} can be considered (finite difference) approximations to TTMs as used in GENIE. These approximations can be inaccurate for large timesteps,
which potentially explains their inferior performance when compared to GENIE. When compared to DDIM, the superior performance of GENIE seems to become less significant for large NFE: this is in line with the theory, as higher-order gradients contribute less for smaller step sizes~(see the GENIE scheme in \Cref{eq:hsds}). Approaches such as FastDDIM~\citep{kong2021arxiv} and AnalyticDDIM~\citep{bao2022analyticdpm}, which adapt variances and discretizations of discrete-time DDMs, are useful; however, GENIE suggests that rigorous higher-order ODE solvers leveraging the continuous-time DDM formalism are still more powerful. To the best of our knowledge, the only methods that outperform GENIE abandon this ODE or SDE formulation entirely and train NFE-specific models~\citep{salimans2022progressive, xiao2022tackling} which are optimized for the single use-case of image synthesis.\looseness=-1

\begin{figure}[t!]
\vspace{-0.8cm}
    \centering
    \begin{minipage}[t]{0.66\textwidth}
        \begin{minipage}[b]{0.5\textwidth}
\centering
\begin{tikzpicture}[scale=0.55]
\begin{axis}[title style={below right,at={(0.747,0.965)},draw=black,fill=white}, title=CIFAR-10, xtick style={draw=none}, xtick={30, 25, 20, 15, 10, 5}, xticklabels={}, ytick={3, 4, 5, 6, 7, 8, 9, 10, 12, 14, 16, 18, 20}, yticklabels={3, 4, 5, 6, 7, 8, 9, 10, 12, 14, 16, 18, 20}, ymin=2.5, ymax=21, xmin=4, xmax=26.8, ylabel=FID, every axis plot/.append style={ultra thick}, every axis plot/.append style={mark size=3pt}, grid=both, grid style={line width=.1pt, draw=gray!10}, major grid style={line width=.2pt,draw=gray!50},ylabel near ticks]
\addplot[color=set25,mark=*,mark size=2pt,only marks] coordinates {
(25, 4.25)
(20, 4.72)
(15, 5.90)
(10, 7.86)
(5, 12.4)
};
\addplot[color=set26,mark=*,mark size=2pt,only marks] coordinates {
(20, 5.05)
(10, 9.9)
};
\addplot[color=set28,mark=*,mark size=2pt,only marks] coordinates {
(25, 9.22)
(10, 15.6)
};
\addplot[color=set21,mark=x, opacity=0.7] coordinates {
(25.37, 3.64)
(20.29, 3.94)
(15.22, 4.49)
(10.15, 5.28)
(5.073, 11.2)
};
\addplot[color=set22,mark=x, opacity=0.7] coordinates {
(5, 27.6)
(10, 11.2)
(15, 7.35)
(20, 5.87)
(25, 5.16)
};
\addplot[color=set23,mark=x, opacity=0.7] coordinates {
(5, 35.9)
(10, 10.3)
(15, 6.61)
(20, 5.20)
(25, 4.51)
};
\addplot[color=set24,mark=x, opacity=0.7] coordinates {
(15, 10.3)
(20, 5.96)
(25, 4.73)
};
\end{axis}
\end{tikzpicture}
\end{minipage}%
\begin{minipage}[b]{.5\textwidth}
\centering
\begin{tikzpicture}[scale=0.55]
\begin{axis}[title style={below right,at={(0.761,0.965)},draw=black,fill=white},title=ImageNet, xtick style={draw=none}, ytick style={draw=none}, xtick={30, 25, 20, 15, 10, 5}, xticklabels={}, ytick={3, 4, 5, 6, 7, 8, 9, 10, 12, 14, 16, 18, 20}, yticklabels={}, ymin=2.5, ymax=21, xmin=4, xmax=26.8, every axis plot/.append style={ultra thick}, every axis plot/.append style={mark size=3pt}, grid=both, grid style={line width=.1pt, draw=gray!10}, major grid style={line width=.2pt,draw=gray!50}]
\addplot[color=set21,mark=x, opacity=0.7] coordinates {
(24.67, 4.13)
(19.54, 4.48)
(15.42, 5.23)
(10.28, 7.41)
(5.142, 20.2)
};
\addplot[color=set22,mark=x, opacity=0.7] coordinates {
(5, 30.0)
(10, 10.7)
(15, 7.14)
(20, 5.83)
(25, 5.19)
};
\addplot[color=set23,mark=x, opacity=0.7] coordinates {
(5, 35.5)
(10, 11.2)
(15, 8.54)
(20, 7.52)
(25, 6.94)
};
\addplot[color=set24,mark=x, opacity=0.7] coordinates {
(15, 12.3)
(20, 9.01)
(25, 7.74)
};
\end{axis}
\end{tikzpicture}
\end{minipage}\\
\begin{minipage}[b]{.5\textwidth}
\centering
\begin{tikzpicture}[scale=0.55]
\begin{axis}[title style={below right,at={(0.606,0.963)},draw=black,fill=white},title=LSUN Bedrooms, xtick={30, 25, 20, 15, 10, 5}, xticklabels={30, 25, 20, 15, 10, 5}, ytick={3, 4, 5, 6, 7, 8, 9, 10, 12, 14, 16, 18, 20}, yticklabels={3, 4, 5, 6, 7, 8, 9, 10, 12, 14, 16, 18, 20}, ymin=4.2, ymax=21, xmin=4, xmax=26.8, xlabel=NFEs, ylabel=FID, every axis plot/.append style={ultra thick}, every axis plot/.append style={mark size=3pt}, grid=both, grid style={line width=.1pt, draw=gray!10}, major grid style={line width=.2pt,draw=gray!50},ylabel near ticks,xlabel near ticks]
\addplot[color=set25,mark=*,mark size=2pt,only marks, opacity=0.7] coordinates {
(20, 4.82)
(10, 11.0)
(5, 29.1)
};
\addplot[color=set21,mark=x, opacity=0.7] coordinates {
(25.08, 5.35)
(20.52, 5.79)
(14.82, 6.83)
(10.26, 9.29)
(4.560, 47.3)
};
\addplot[color=set22,mark=x, opacity=0.7] coordinates {
(5, 42.5)
(10, 12.5)
(15, 8.21)
(20, 6.77)
(25, 6.05)
};
\addplot[color=set23,mark=x, opacity=0.7] coordinates {
(5, 45.0)
(10, 10.8)
(15, 8.14)
(20, 7.23)
(25, 6.71)
};
\addplot[color=set24,mark=x, opacity=0.7] coordinates {
(15, 18.9)
(20, 9.27)
(25, 7.69)
};
\end{axis}
\end{tikzpicture}
\end{minipage}%
\begin{minipage}[b]{.5\textwidth}
\centering
\begin{tikzpicture}[scale=0.55]
\begin{axis}[title style={below right,at={(0.477,0.963)},draw=black,fill=white},title=LSUN Church-Outdoor, xtick={30, 25, 20, 15, 10, 5}, ytick style={draw=none}, xticklabels={30, 25, 20, 15, 10, 5}, ytick={3, 4, 5, 6, 7, 8, 9, 10, 12, 14, 16, 18, 20}, yticklabels={}, ymin=4.2, ymax=21, xmin=4, xmax=26.8, xlabel=NFEs, every axis plot/.append style={ultra thick}, every axis plot/.append style={mark size=3pt}, grid=both, grid style={line width=.1pt, draw=gray!10}, major grid style={line width=.2pt,draw=gray!50},xlabel near ticks]
\addplot[color=set25,mark=*,mark size=2pt,only marks, opacity=0.7] coordinates {
(20, 6.74)
(10, 11.6)
(5, 30.2)
};
\addplot[color=set21,mark=x, opacity=0.7] coordinates {
(25.17, 5.84)
(19.45, 6.38)
(14.87, 7.44)
(10.30, 10.5)
(4.576, 60.3)
};
\addplot[color=set22,mark=x, opacity=0.7] coordinates {
(5, 44.0)
(10, 12.8)
(15, 8.44)
(20, 6.97)
(25, 6.28)
};
\addplot[color=set23,mark=x, opacity=0.7] coordinates {
(5, 40.7)
(10, 12.9)
(15, 9.10)
(20, 7.82)
(25, 7.12)
};
\addplot[color=set24,mark=x, opacity=0.7] coordinates {
(15, 12.6)
(20, 9.29)
(25, 7.83)
};
\end{axis}
\end{tikzpicture}
\end{minipage}
    \end{minipage}
    \begin{minipage}[]{0.30\textwidth}
        \begin{tikzpicture}
\begin{customlegend}[legend columns=1,legend style={align=right, draw=none, font=\small}, legend cell align=left,
        legend entries={GENIE (ours),
                        DDIM~\citep{song2021denoising},
                        S-PNDM~\citep{liu2022pseudo},
                        F-PNDM~\citep{liu2022pseudo},
                        Learned Sampler~\citep{watson2022learning} (\textdagger),
                        FastDDIM~\citep{kong2021arxiv} (\textdagger),
                        Analytic DDIM~\citep{bao2022analyticdpm} (\textdagger)}]
        \addlegendimage{mark=x,mark size=3pt,solid,line legend,set21, thick}
        \addlegendimage{mark=x,,mark size=3pt,solid,line legend,set22, thick}
        \addlegendimage{mark=x,,mark size=3pt,solid,line legend,set23, thick}
        \addlegendimage{mark=x,mark size=3pt,solid,line legend,set24, thick}
        \addlegendimage{mark=*,mark size=2pt,solid,line legend,set25, only marks, thick}
        \addlegendimage{mark=*,mark size=2pt,solid,line legend,set26,only marks, thick}
        \addlegendimage{mark=*,mark size=2pt,solid,line legend,set28,only marks, thick}
\end{customlegend}
\end{tikzpicture}
        \caption{
\small Unconditional performance on four popular benchmark datasets. The first four methods use the same score model checkpoints, whereas the last three methods all use different checkpoints. (\textdagger): numbers are taken from literature.} %
\label{fig:main_fid_figure}
    \end{minipage}
    \vspace{-0.6cm}
\end{figure}
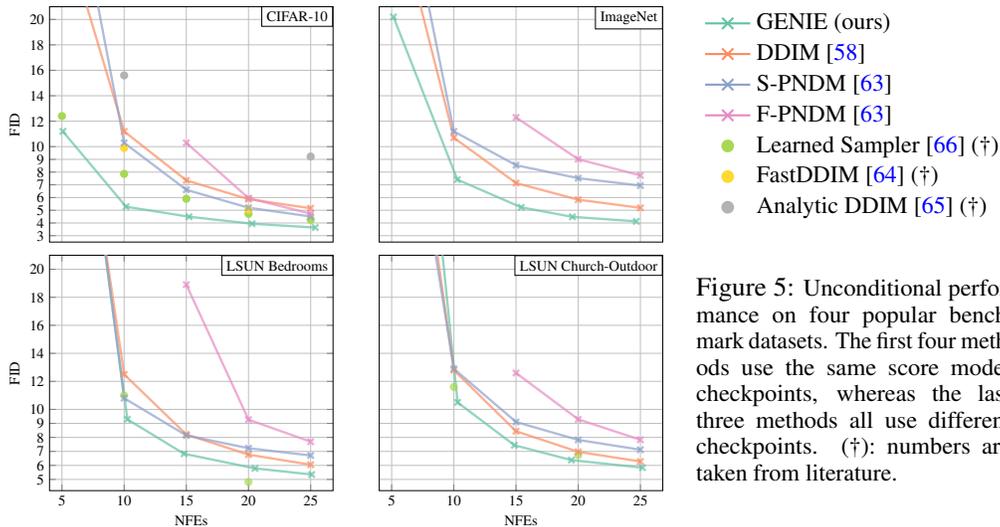
\begin{table}
    \centering
    \begin{minipage}[t]{.69\textwidth}
        \scalebox{0.7}{\begin{tabular}{l@{\hspace{0.2\tabcolsep}}c c c c c c c}
        \toprule
        Method &  NFEs=5 & NFEs=10 & NFEs=15 & NFEs=20 & NFEs=25 \\ %
        \midrule
        GENIE (ours) (\textasteriskcentered) & \textbf{11.2} & \textbf{5.28} & \textbf{4.49} & \textbf{3.94} & \textbf{3.64} \\ %
        GENIE (ours) & 13.9 & 5.97 & \textbf{4.49} & \textbf{3.94} & 3.67 \\ %
        DDIM~\citep{song2021denoising} (\textasteriskcentered) & 27.6 & 11.2 & 7.35 & 5.87 & 5.16 \\
        DDIM~\citep{song2021denoising} & 29.7 & 11.2 & 7.35 & 5.87 & 5.16 \\
        S-PNDM~\citep{liu2022pseudo} & 35.9 & 10.3 & 6.61 & 5.20 & 4.51 \\
        F-PNDM~\citep{liu2022pseudo} & N/A & N/A & 10.3 & 5.96 & 4.73 \\
        Euler--Maruyama & 325 & 230 & 164 & 112 & 80.3 \\
        \midrule
        FastDDIM~\citep{kong2021arxiv} (\textdagger) & - & 9.90 & - & 5.05 & - \\
        Learned Sampler~\citep{watson2022learning} (\textdagger / \textasteriskcentered) & 12.4 & 7.86 & 5.90 & 4.72 & 4.25 \\
        Learned Sampler~\citep{watson2022learning} (\textdagger) & 14.3 & 8.15 & 5.94 & 4.89 & 4.47 \\
        Analytic DDIM~\citep{bao2022analyticdpm} (\textdagger)& - & 14.0 & - & - & 5.71 &\\
        CLD-SGM~\citep{dockhorn2022scorebased}  & 334 & 306 & 236 & 162 & 106 \\
        VESDE-PC~\citep{song2020} & 461 & 461 & 461 & 461 & 462\\
        \bottomrule
    \end{tabular}}
    \end{minipage}\hfill
    \begin{minipage}[]{.31\textwidth}
        \caption{\footnotesize Unconditional CIFAR-10 generative performance (measured in FID). %
        Methods above the middle line use the same score model checkpoint; methods below all use different ones. (\textdagger): numbers are taken from literature. (\textasteriskcentered): methods either learn an optimal striding schedule (Learned Sampler) or do a small grid search over striding schedules (DDIM \& GENIE); also see~\Cref{s:app_extended_quant}}%
        \label{tab:main_cifar_10}
    \end{minipage}
    \vspace{-0.5cm}
\end{table}
\vspace{-0.2cm}
\subsection{Guidance and Encoding} \label{s:guidance_and_encoding}
\vspace{-0.2cm}
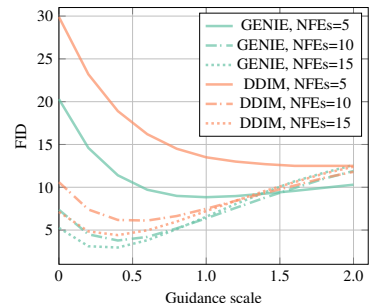
\begin{wrapfigure}{r}{0.33\textwidth}
\vspace{-1.4cm}
  \begin{center}
    \begin{tikzpicture}[scale=0.6]
\begin{axis}[, xtick={2.0, 1.5, 1.0, 0.5, 0.0}, xticklabels={2.0, 1.5, 1.0, 0.5, 0}, ytick={0, 5, 10, 15, 20, 25, 30}, yticklabels={0, 5, 10, 15, 20, 25, 30}, ymin=1, ymax=31, xmin=0, xmax=2.1, xlabel=Guidance scale, ylabel=FID, grid=both, grid style={line width=.1pt, draw=gray!10}, major grid style={line width=.2pt,draw=gray!50}, every axis plot/.append style={ultra thick}, every axis plot/.append style={mark size=3pt}, xlabel near ticks, ylabel near ticks]
\addlegendentry{GENIE, NFEs=5}
\addplot[color=set21, opacity=0.7] coordinates {
(0.0, 20.3)
(0.2, 14.6)
(0.4, 11.4)
(0.6, 9.71)
(0.8, 8.99)
(1.0, 8.84)
(1.2, 8.97)
(1.4, 9.26)
(1.6, 9.58)
(1.8, 9.94)
(2.0, 10.3)
};
\addlegendentry{GENIE, NFEs=10}
\addplot[color=set21,dash pattern={on 7pt off 2pt on 1pt off 3pt}, opacity=0.7] coordinates {
(0.0, 7.37)
(0.2, 4.50)
(0.4, 3.78)
(0.6, 4.20)
(0.8, 5.20)
(1.0, 6.37)
(1.2, 7.59)
(1.4, 8.79)
(1.6, 9.91)
(1.8, 11.0)
(2.0, 11.9)
};
\addlegendentry{GENIE, NFEs=15}
\addplot[color=set21,dotted, opacity=0.7] coordinates {
(0.0, 5.26)
(0.2, 3.11)
(0.4, 2.96)
(0.6, 3.81)
(0.8, 5.11)
(1.0, 6.56)
(1.2, 8.04)
(1.4, 9.42)
(1.6, 10.7)
(1.8, 11.7)
(2.0, 12.6)
};
\addlegendentry{DDIM, NFEs=5}
\addplot[color=set22, opacity=0.7] coordinates {
(0.0, 29.9)
(0.2, 23.2)
(0.4, 18.9)
(0.6, 16.2)
(0.8, 14.5)
(1.0, 13.5)
(1.2, 13.0)
(1.4, 12.7)
(1.6, 12.5)
(1.8, 12.5)
(2.0, 12.5)
};
\addlegendentry{DDIM, NFEs=10}
\addplot[color=set22,dash pattern={on 7pt off 2pt on 1pt off 3pt}, opacity=0.7] coordinates {
(0.0, 10.6)
(0.2, 7.40)
(0.4, 6.18)
(0.6, 6.11)
(0.8, 6.64)
(1.0, 7.49)
(1.2, 8.42)
(1.4, 9.34)
(1.6, 10.2)
(1.8, 11.1)
(2.0, 11.8)
};
\addlegendentry{DDIM, NFEs=15}
\addplot[color=set22,dotted, opacity=0.7] coordinates {
(0.0, 7.12)
(0.2, 4.84)
(0.4, 4.41)
(0.6, 4.97)
(0.8, 5.99)
(1.0, 7.22)
(1.2, 8.42)
(1.4, 9.58)
(1.6, 10.6)
(1.8, 11.6)
(2.0, 12.4)
};
\end{axis}
\end{tikzpicture}
  \end{center}
  \vspace{-0.4cm}
  \caption{\small Sample quality as a function of guidance scale on ImageNet. %
  }
  \label{fig:guided_diffusion_plot}
  \vspace{-0.3cm}
\end{wrapfigure}

As discussed in~\Cref{s:related_work}, one major drawback of approaches such as KD~\citep{luhman2021knowledge}, PG~\citep{salimans2022progressive} and DDGAN~\citep{xiao2022tackling} is that they abandon the ODE/SDE formalism, and cannot easily use methods such as classifier(-free) guidance~\citep{song2020,ho2021classifierfree} or perform image encoding. 
However, these techniques can play an important role
in synthesizing photorealistic images from DDMs~\citep{dhariwal2021diffusion, nichol2021glide, nichol21, ramesh2022dalle2}, as well as for image editing tasks~\cite{meng2021sdedit,ramesh2022dalle2}.
\looseness=-1

\vspace{-0.1cm}
\textbf{Classifier-Free Guidance~\citep{ho2021classifierfree}:} We replace the unconditional model $\vepsilon_\vtheta(\rvx_t, t)$ with $\hat \vepsilon_\vtheta(\rvx_t, t, c, w) = (1+w) \vepsilon_{\vtheta}(\rvx_t, t, c) - w \vepsilon_\vtheta(\rvx_t, t)$ in the DDIM ODE (cf~\Cref{eq:ddim_ode}), where $\vepsilon_\vtheta(\rvx_t, t, c)$ is a conditional model and $w > 1.0$ is the ``guidance scale''. 
GENIE then requires the derivative
\vspace{-0.15cm}
\begin{align}
    d_{\gamma_t} \hat \vepsilon_\vtheta(\rvx_t, t, c, w) = (1+w) d_{\gamma_t}\vepsilon_\vtheta (\rvx_t, t, c)- w d_{\gamma_t}\vepsilon_\vtheta (\rvx_t, t).
\end{align}
for guidance. Hence, we need to distill 
$d_{\gamma_t}\vepsilon_\vtheta (\rvx_t, t, c)$ and 
$d_{\gamma_t}\vepsilon_\vtheta (\rvx_t, t)$, for which we could also share parameters~\citep{ho2021classifierfree}.
We compare GENIE with DDIM on ImageNet in~\Cref{fig:guided_diffusion_plot}. 
GENIE clearly outperforms DDIM, in particular for few NFEs, and GENIE also synthesizes high-quality images (see \Cref{fig:guidance}).\looseness=-1

\vspace{-1mm}
\textbf{Image Encoding:} We can use GENIE also to solve the generative ODE in reverse to encode given images. Therefore,
we compare GENIE to DDIM on the ``encode-decode'' task, analyzing reconstructions for different NFEs (used twice for encoding and decoding): We find that GENIE reconstructs images much more accurately (see~\Cref{fig:latent_interpolation}).
For more details on this experiment as well as the guidance experiment above, see~\Cref{s:app_encoding} and~\Cref{s:app_cfg}, respectively. We also show latent space interpolations for both GENIE and DDIM in~\Cref{s:app_lsi}.\looseness=-1

\begin{figure}
\vspace{-0.8cm}
\begin{minipage}[c]{0.83\textwidth}
    \centering
    \includegraphics[width=\textwidth]{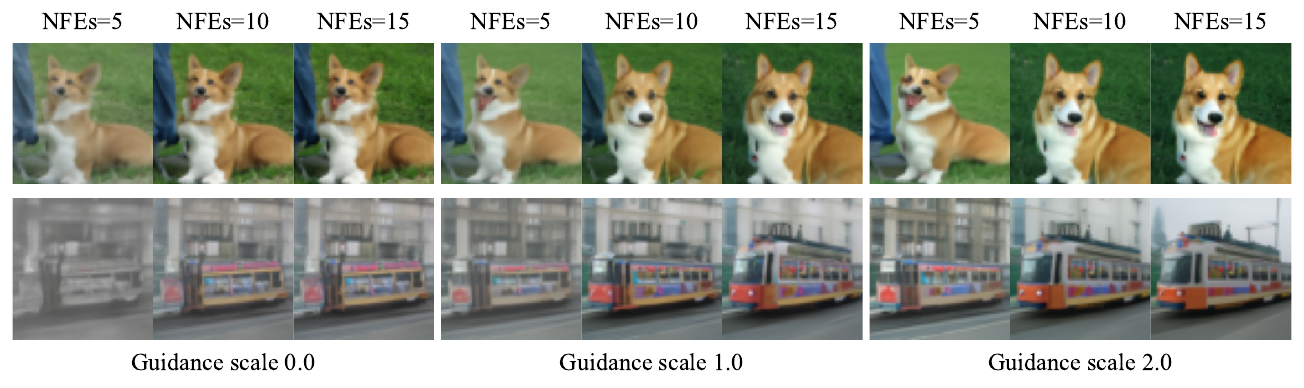}
    \end{minipage}\hfill
    \begin{minipage}[c]{0.15\textwidth}
    \caption{\small Classifier-free guidance for the ImageNet classes Pembroke Welsh Corgi (263) and Streetcar (829).}
    \label{fig:guidance}
    \end{minipage}
\vspace{-0.4cm}
\end{figure}
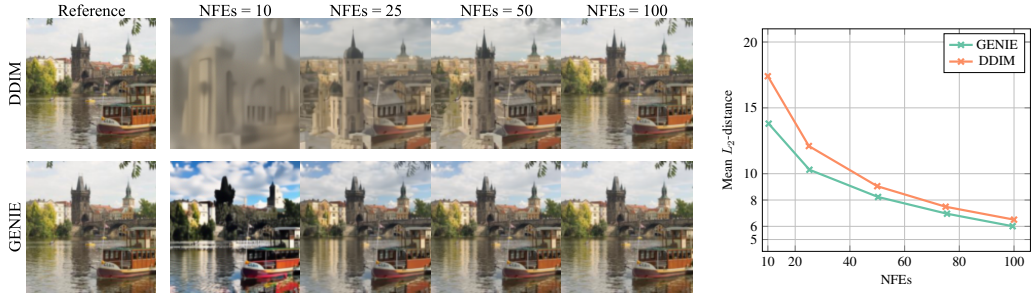
\begin{figure}
    \centering
    \begin{subfigure}{0.65\textwidth}
        \centering
        \includegraphics[scale=0.9]{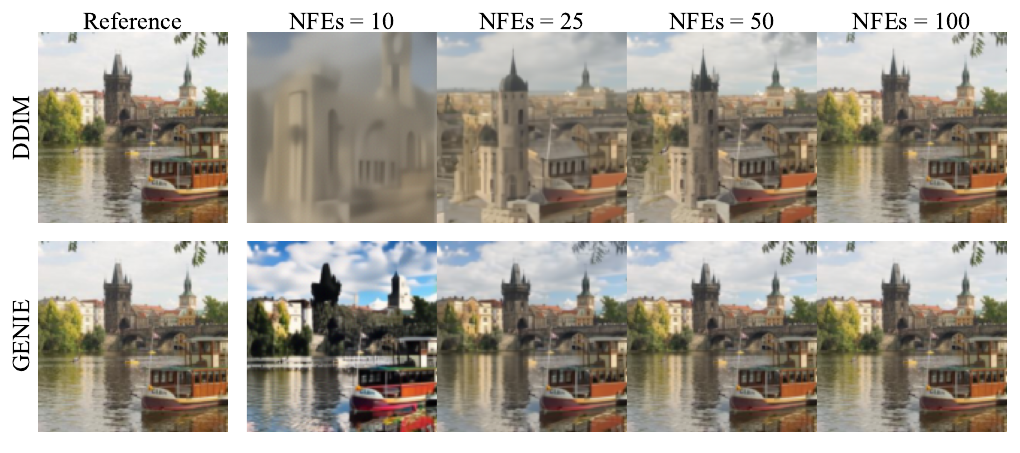}
    \end{subfigure}\hspace{1em}%
    \begin{subfigure}{0.32\textwidth}
    \centering
        \vspace{0.2cm}
        \begin{tikzpicture}[scale=0.52]
\begin{axis}[xtick={100, 80, 60, 40, 20, 10, 5}, xticklabels={100, 80, 60, 40, 20, 10, 5}, ytick={2, 4, 5, 6, 8, 10, 15, 20}, yticklabels={2, 4, 5, 6, 8, 10, 15, 20}, ymin=4.1, ymax=21, xmin=8, xmax=106, xlabel=NFEs, ylabel=Mean $L_2$-distance, grid=both, grid style={line width=.1pt, draw=gray!10}, major grid style={line width=.2pt,draw=gray!50}, every axis plot/.append style={ultra thick}, every axis plot/.append style={mark size=3pt}, xlabel near ticks, ylabel near ticks]
\addplot[color=set21,mark=x] coordinates {
(99.5, 6.01)
(75.5, 6.96)
(50.3, 8.23)
(25.2, 10.3)
(10.3, 13.8)
};
\addlegendentry{GENIE}
\addplot[color=set22,mark=x] coordinates {
(100, 6.51)
(75, 7.49)
(50, 9.05)
(25, 12.1)
(10, 17.4)
};
\addlegendentry{DDIM}
\end{axis}
\end{tikzpicture}
    \end{subfigure}
    \vspace{-0.4cm}
    \caption{\small Encoding and subsequent decoding on LSUN Church-Outdoor. \textit{Left:} Visual reconstruction. \textit{Right:} $L_2$-distance to reference in Inception feature space~\citep{szegedy2016rethinking}, averaged over 100 images.}
    \vspace{-0.6cm}
    \label{fig:latent_interpolation}
\end{figure}
\vspace{-0.2cm}
\subsection{Ablation Studies} \label{s:ablations}
\vspace{-0.2cm}
\begin{wraptable}{r}{0.54\textwidth}
    \vspace{-1.1cm}
    \centering
    \caption{\small CIFAR-10 ablation studies (measured in FID).}\vspace{-0.2cm}
    \scalebox{0.67}{
    \begin{tabular}{l c c c c c}
        \toprule
        Ablation & NFEs=5 & NFEs=10 & NFEs=15 & NFEs=20 & NFEs=25 \\ %
        \midrule
        Standard & 13.9 & 6.04 & 4.49 & 3.94 & 3.67 \\ %
        No mixed & 14.7 & 6.32 & 4.82 & 4.31 & 4.10 \\ %
        No weighting & 14.8 & 7.45 & 5.89 & 5.17 & 4.80 \\ %
        \midrule 
        Bigger model & 13.7 & 5.58 & 4.46 & 4.05 & 3.77 \\ %
        \bottomrule
    \end{tabular}
    }
    \label{tab:ablations}
    \vspace{-0.3cm}
\end{wraptable}
We perform ablation studies over architecture and training objective for the prediction heads used in GENIE: 
In~\Cref{tab:ablations}, ``No mixed'' refers to learning $d_{\gamma_t}\vepsilon_\vtheta$ directly as single network output without mixed network parameterization; ``No weighting'' refers to setting $g_\mathrm{d}(t)=1$ in~\Cref{eq:objective};
``Standard'' uses both the mixed network parameterization and the weighting function $g_\mathrm{d}(t) = \gamma_t^2$.
We can see that having both the mixed network parametrization and the weighting function is clearly beneficial. We also tested
deeper networks in the prediction heads:
for ``Bigger model'' we increased the number of residual blocks from one to two.
The performance is roughly on par with ``Standard'', and we therefore opted for the smaller head due to the lower computational overhead.\looseness=-1

\vspace{-2mm}
\subsection{Upsampling}
\begin{figure}
\vspace{-3mm}
    \centering
    \includegraphics[width=0.8\textwidth]{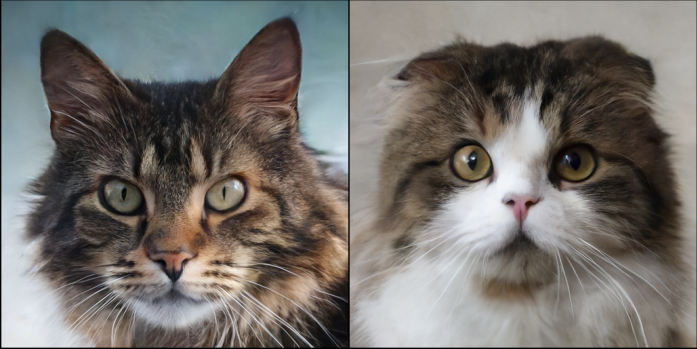}\\
    \vspace{3pt}
    \includegraphics[width=0.8\textwidth]{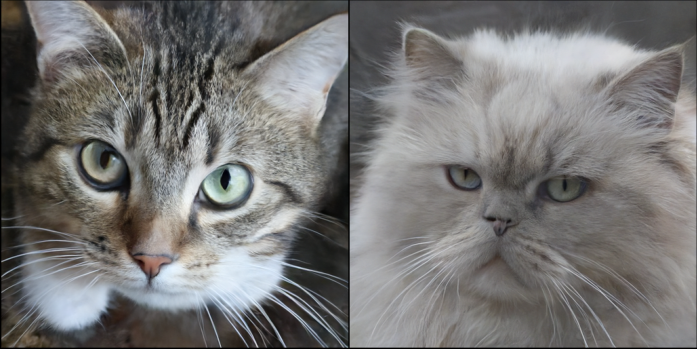}
    \caption{High-resolution images generated with the $128 \times 128 \rightarrow 512 \times 512$ GENIE upsampler using only five neural network calls. For the two images at the top, the upsampler is conditioned on test images from the Cats dataset. For the two images at the bottom, the upsampler is conditioned on samples from the $128 \times 128$ GENIE base model (generated using 25 NFEs); an upsampler neural network evaluation is roughly four times as expensive as a base model evaluation.}
    
    \label{fig:upsampling}
    \vspace{-3mm}
\end{figure}

\vspace{-2mm}
\begin{wraptable}{r}{0.4\textwidth}
\vspace{-1.2cm}
\caption{\footnotesize Cats (upsampler) generative performance (measured in FID).}
    \label{tab:upsampling}
    \centering
        \scalebox{0.7}{\begin{tabular}{l@{\hspace{0.2\tabcolsep}}c c c c c}
        \toprule
        Method &  NFEs=5 & NFEs=10 & NFEs=15 \\
        \midrule
        GENIE (ours) & \textbf{5.53} & \textbf{4.90} & \textbf{4.83} \\ 
        DDIM~\citep{song2021denoising} & 9.47 & 6.64 & 5.85 \\
        S-PNDM~\citep{liu2022pseudo} & 14.6 & 11.0 & 8.83\\
        F-PNDM~\citep{liu2022pseudo} & N/A & N/A & 11.7 \\
        \bottomrule
    \end{tabular}}
\vspace{-0.3cm}
\end{wraptable}
Cascaded diffusion model pipelines~\citep{ho2021arxiv} and DDM-based super-resolution~\citep{saharia2021image} have become crucial ingredients in DDMs for large-scale image generation~\citep{saharia2022photorealistic}. Hence, we also explore the applicability of GENIE in this setting.
We train a $128 \times 128$ base model as well as a $128 \times 128 \rightarrow 512 \times 512$ diffusion upsampler~\citep{ho2021arxiv,saharia2021image} on Cats. In~\Cref{tab:upsampling}, we compare the generative performance of GENIE to other fast samplers for the upsampler (in isolation). We find that GENIE performs very well on this task: with only five NFEs GENIE outperforms all other methods at NFEs=15. We show upsampled samples for GENIE with NFEs=5 in~\Cref{fig:upsampling}. For more quantitative and qualitative results, we refer to~\Cref{s:app_extended_quant} and~\Cref{s:app_extended_qual}, respectively. Training and inference details for the score model and the GENIE prediction head, for both base model and upsampler, can be found in~\Cref{s:app_model_details}.\looseness=-1
\vspace{-0.3cm}
\section{Conclusions}
\label{s:conclusion}
\vspace{-0.3cm}
We introduced GENIE, a higher-order ODE solver for DDMs. GENIE improves upon the commonly used DDIM solver
by capturing the local curvature of its ODE's gradient field, which allows 
for larger step sizes when solving the ODE. We further propose to distill the required higher-order derivatives 
into a small prediction head---which we can efficiently call during inference---on top of the first-order score network. A limitation of GENIE is that it is still slightly slower than approaches that abandon the differential equation framework of DDMs altogether, which, however, comes at the considerable cost 
of preventing applications such as guided sampling.
To overcome this limitation, future work could leverage even higher-order gradients
to accelerate sampling from DDMs even further (also see~\Cref{s:app_genie_x_pgd}).\looseness=-1

\vspace{-1mm}
\textbf{Broader Impact.} Fast synthesis from DDMs, the goal of GENIE, can potentially make DDMs an attractive method for promising interactive generative modeling applications, such as digital content creation or real-time audio synthesis, and also reduce DDMs' environmental footprint by decreasing the computational load during inference. Although we validate GENIE on image synthesis, it could also be utilized for other tasks, which makes its broader societal impact application-dependent. In that context, it is important that practitioners apply an abundance of caution to mitigate impacts given generative modeling can also be used for malicious purposes, discussed for instance in~\citet{vaccari2020deepfakes,nguyen2021deep,mirsky2021deepfakesurvey}.
\section*{Acknowledgements} We thank Yaoliang Yu for early discussions. Tim Dockhorn acknowledges additional funding from the Vector Institute Research Grant, which is not in direct support of this work.

\bibliographystyle{unsrtnat}
\addcontentsline{toc}{section}{References}
{\small \bibliography{iclr_2022_conference_fixed,neurips_2022}}

\clearpage

\tableofcontents
\clearpage
\appendix

\section{DDIM ODE} \label{s:app_ddim_ode}
The DDIM ODE has previously been shown~\citep{song2021denoising, salimans2022progressive} to be a re-parameterization of the Probability Flow ODE~\citep{song2020}. In this section, we show an alternative presentation to the ones given in~\citet{song2021denoising} and~\citet{salimans2022progressive}. We start from the Probability Flow ODE for variance-preserving continuous-time DDMs~\citep{song2020}, i.e., 
\begin{align} \label{eq:app_probability_flow_ode}
    d\rvx_t &= -\tfrac{1}{2} \beta_{t} \left[\rvx_t + \nabla_{\rvx_t} \log p_t(\rvx_t) \right]dt,
\end{align}
where $\beta_t = -\frac{d}{dt} \log \alpha_t^2$ and $\nabla_{\rvx_t} \log p_t(\rvx_t)$ is the \emph{score function}. Replacing the unknown score function with a learned score model $\vs_\vtheta(\rvx_t, t) \approx \nabla_{\rvx_t} \log p_t(\rvx_t)$, we obtain the approximate Probability Flow ODE
\begin{align} \label{eq:app_approximate_probability_flow_ode}
    d\rvx_t &= -\tfrac{1}{2} \beta_{t} \left[\rvx_t + \vs_\vtheta(\rvx_t, t) \right]dt.
\end{align}
Let us now define $\gamma_t = \sqrt{\frac{1 - \alpha_t^2}{\alpha_t^2}}$ and $\bar \rvx_t = \rvx_t \sqrt{1 + \gamma_t^2}$, and take the (total) derivative of $\bar \rvx_t$ with respect to $\gamma_t$:
\begin{align}
    \frac{d\bar \rvx_t}{d\gamma_t} &= \frac{\partial \bar \rvx_t}{\partial \rvx_t} \frac{d\rvx_t}{d\gamma_t} + \frac{\partial \bar \rvx_t}{\partial \gamma_t}\\
    &= \sqrt{1+\gamma^2_t} \frac{d\rvx_t}{d\gamma_t} + \frac{\gamma_t}{\sqrt{1+\gamma^2_t}} \rvx_t. \label{eq:dbarx_dgamma}
\end{align}
The derivative $\frac{d\rvx_t}{d\gamma_t}$ can be computed as follows
\begin{align}
    \frac{d\rvx_t}{d\gamma_t} &= \frac{d\rvx_t}{dt} \frac{dt}{d\gamma_t} \quad (\text{by chain rule}) \\
    &= -\frac{1}{2} \beta_{t} \left[\rvx_t + \vs_\vtheta(\rvx_t, t) \right] \frac{dt}{d\gamma_t} \quad (\text{inserting~\Cref{eq:app_approximate_probability_flow_ode}}) \\
    &= \frac{1}{2} \frac{d\log \alpha_t^2}{dt} \left[\rvx_t + \vs_\vtheta(\rvx_t, t) \right] \frac{dt}{d\gamma_t} \quad (\text{by definition of} \, \beta_t) \\
    &= \frac{1}{2} \frac{d\log \alpha_t^2}{d\gamma_t} \left[\rvx_t + \vs_\vtheta(\rvx_t, t) \right] \quad (\text{by chain rule}) \\
    &= \frac{1}{2} \frac{d\log \alpha_t^2}{d\alpha_t^2} \frac{d\alpha_t^2}{d\gamma_t} \left[\rvx_t + \vs_\vtheta(\rvx_t, t) \right] \quad (\text{by chain rule}) \\
    &= \frac{1}{2} \frac{1}{\alpha_t^2} \frac{d\alpha_t^2}{d\gamma_t} \left[\rvx_t + \vs_\vtheta(\rvx_t, t) \right]. \label{eq:app_dx_dgamma}
\end{align}
We can write $\alpha_t^2$ as a function of $\gamma_t$, i.e., $\alpha_t^2 = \left(\gamma_t^2 + 1 \right)^{-1}$, and therefore
\begin{align} \label{eq:app_dalpha_squared_dgamma}
    \frac{d\alpha_t^2}{d\gamma_t} = - \frac{2 \gamma_t}{\left(\gamma_t^2 + 1 \right)^2}.
\end{align}
Inserting~\Cref{eq:app_dalpha_squared_dgamma} into~\Cref{eq:app_dx_dgamma}, we obtain
\begin{align}
    \frac{d\rvx_t}{d\gamma_t} = - \frac{\gamma_t}{\gamma_t^2 + 1} \left[\rvx_t + \vs_\vtheta(\rvx_t, t) \right]. \label{eq:app_dx_dgamma2}
\end{align}
Lastly, inserting~\Cref{eq:app_dx_dgamma2} into~\Cref{eq:dbarx_dgamma}, we have
\begin{align}
    \frac{d\bar \rvx_t}{d\gamma_t} = - \frac{\gamma_t}{\sqrt{\gamma_t^2 + 1}} \vs_\vtheta(\rvx_t, t)
\end{align}
Letting $\vs_\vtheta(\rvx_t, t) \coloneqq - \frac{\vepsilon_\vtheta(\rvx_t, t)}{\sigma_t}$, where $\sigma_t = \sqrt{1-\alpha_t^2} = \frac{\gamma_t}{\sqrt{\gamma_t^2 + 1}}$, denote a particular parameterization of the score model, we obtain the approximate generative DDIM ODE as
\begin{align}
    \frac{d\bar \rvx_t}{d\gamma_t} &= \frac{\gamma_t}{\sqrt{\gamma_t^2 + 1}} \frac{\vepsilon_\vtheta(\rvx_t, t)}{\sigma_t} \\
    &= \vepsilon_\vtheta(\rvx_t, t). \label{eq:app_ddim_ode}
\end{align}
\newpage
\section{Synthesis from Denoising Diffusion Models via Truncated Taylor Methods} \label{s:app_ttm}
In this work, we propose Higher-Order Denoising Diffusion Solvers (GENIE). GENIE is based on the \emph{truncated Taylor method} (TTM)~\citep{kloeden1992stochastic}. As outlined in~\Cref{s:method}, the $p$-th TTM is simply the \emph{p-th order Taylor polynomial} applied to an ODE. For example, for the general $\frac{d\rvy}{dt} = \vf(\rvy, t)$, the $p$-th TTM reads as 
\begin{align}
    \rvy_{t_{n+1}} = \rvy_{t_n} + h_n \frac{d\rvy}{dt} \rvert_{(\rvy_{t_n}, t_n)} + \cdots + \frac{1}{p!} h_n^p \frac{d^p\rvy}{dt^p} \rvert_{(\rvy_{t_n}, t_n)},
\end{align}
where $h_n = t_{n+1} - t_n$. To generate samples from denoising diffusion models, we can, for example, apply the second TTM to the (approximate) Probability Flow ODE or the (approximate) DDIM ODE, resulting in the following respective schemes:
\begin{align}
    \rvx_{t_{n+1}} = \rvx_{t_n} + (t_{n+1} - t_n) \vf(\rvx_{t_n}, t_n) + \frac{1}{2} (t_{n+1} - t_n)^2 \frac{d\vf}{dt} \rvert_{(\rvx_{t_n}, t_n)},
\end{align}
where $\vf(\rvx_t, t) = - \tfrac{1}{2} \beta(t) \left[\rvx_t - \tfrac{\vepsilon_\vtheta(\rvx_t, t)}{\sigma_t} \right]$, 
and
\begin{align} \label{eq:app_genie}
    \bar \rvx_{t_{n+1}} = \bar \rvx_{t_n} + (\gamma_{t_{n+1}} - \gamma_{t_n}) \vepsilon_\vtheta(\rvx_{t_n}, t_n) + \frac{1}{2} (\gamma_{t_{n+1}} - \gamma_{t_n})^2 \frac{d\vepsilon_\vtheta}{d\gamma_t} \rvert_{(\rvx_{t_n}, t_n)}.
\end{align}
In this work, we generate samples from DDMs using the scheme in~\Cref{eq:app_genie}. We distill the derivative $d_{\gamma_t} \vepsilon_\vtheta \coloneqq \frac{d\vepsilon_\vtheta}{d\gamma_t}$ into a small neural network $\vk_\vpsi$. For training, $d_{\gamma_t} \vepsilon_\vtheta$ is computed via automatic differentiation, however, during inference, we can efficiently query the trained network $\vk_\vpsi$. 
\subsection{Theoretical Bounds for the Truncated Taylor Method}\label{app:theoreticalbounds}
Consider the $p$-TTM for a general ODE $\frac{d\rvy}{dt} = \vf(\rvy, t)$:
\begin{align}
    \rvy_{t_{n+1}} = \rvy_{t_n} + h_n \frac{d\rvy}{dt} \rvert_{(\rvy_{t_n}, t_n)} + \cdots + \frac{1}{p!} h_n^p \frac{d^p\rvy}{dt^p} \rvert_{(\rvy_{t_n}, t_n)}.
\end{align}
We represent, the exact solution $\rvy(t_{n+1})$ using the $(p+2)$-th Taylor expansion
\begin{align}
    \rvy(t_{n+1}) = \rvy(t_n) + h_n \frac{d\rvy}{dt} \rvert_{(\rvy_{t_n}, t_n)} + \cdots + \frac{1}{p!} h_n^p \frac{d^p\rvy}{dt^p} \rvert_{(\rvy_{t_n}, t_n)} + \frac{1}{(p+1)!} h_n^{p+1} \frac{d^{p+1}\rvy}{dt^{p+1}} \rvert_{(\rvy_{t_n}, t_n)} + \gO(h_n^{p+2}).
\end{align}
The local truncation error (LTE) introduced by the $p$-th TTM is given by the difference between the two equations above
\begin{align}
    \|\rvy_{t_{n+1}} - \rvy(t_{n+1})\| = \|\frac{1}{(p+1)!} h_n^{p+1} \frac{d^{p+1}\rvy}{dt^{p+1}} \rvert_{(\rvy_{t_n}, t_n)} + \gO(h_n^{p+2})\|.
\end{align}
For small $h_n$, the LTE is proportional to $h_n^{p+1}$. Consequently, using higher orders $p$ implies lower errors, as $h_n$ usually is a small time step.

In conclusion, this demonstrates that it is preferable to use higher-order methods with lower errors when aiming to accurately solve ODEs like the Probability Flow ODE or the DDIM ODE of diffusion models.
\subsection{Approximate Higher-Order Derivatives via the ``Ideal Derivative Trick''} \label{s:app_2nd_ttm_id}
\citet{tachibana2021taylor} sample from DDMs using (an approximation to) a higher-order Itô-Taylor method~\citep{kloeden1992euler}. In their scheme, they approximate higher-order score functions with the ``ideal derivative trick'', essentially assuming simple single-point ($\rvx_0$) data distributions, for which higher-order score functions can be computed analytically (more formally, their approximation corresponds to ignoring the expectation over the full data distribution when learning the score function. They assume that for any $\rvx_t$, there is a single unique $\rvx_0$ from the input data to be predicted with the score model). In that case, further assuming the score model $\vepsilon_\vtheta(\rvx_t, t)$ is learnt perfectly (i.e., it perfectly predicts the noise that was used to generate $\rvx_t$ from $\rvx_0$), one has 
\begin{align} \label{eq:app_tachibana_assump}
    \vepsilon_\vtheta(\rvx_t, t) \approx \frac{\rvx_t - \alpha_t \rvx_0}{\sigma_t}.
\end{align}
This expression can now be used to analytically calculate approximate spatial and time derivatives (also see App. F.1 and App. F.2 in~\citet{tachibana2021taylor}):
\begin{align} \label{eq:app_tachibana_spatial}
    \frac{\partial \vepsilon_\vtheta(\rvx_t, t)}{\partial \rvx_t} \approx \frac{\partial }{\partial \rvx_t} \left(\frac{\rvx_t - \alpha_t \rvx_0}{\sigma_t} \right) = \frac{1}{\sigma_t} \mI,
\end{align}
and 
\begin{align} \label{eq:app_tachibana_time}
    \frac{\partial \vepsilon_\vtheta(\rvx_t, t)}{\partial t} \approx \frac{\partial}{\partial t} \left(\frac{\rvx_t - \alpha_t \rvx_0}{\sigma_t} \right) = - \frac{\rvx_t - \alpha_t \rvx_0}{\sigma_t^2} \frac{d\sigma_t}{dt} - \frac{\rvx_0}{\sigma_t} \frac{d\alpha_t}{dt}.
\end{align}
Rearranging \Cref{eq:app_tachibana_assump}, we have
\begin{align}
    \rvx_0 \approx \frac{\rvx_t - \sigma_t \vepsilon_\vtheta(\rvx_t, t)}{\alpha_t}.
\end{align}
Inserting this expression,
~\Cref{eq:app_tachibana_time} becomes
\begin{align} \label{eq:app_tachibana_time2}
    \frac{\partial \vepsilon_\vtheta(\rvx_t, t)}{\partial t} \approx  \frac{\frac{d \log \alpha_t^2}{dt}}{2 \sigma_t} \left(\frac{\vepsilon_\vtheta(\rvx_t, t)}{\sigma_t}  - \rvx_t \right).
\end{align}
We will now proceed to show that the ``ideal derivative trick'', i.e. using the approximations in~\Cref{eq:app_tachibana_spatial,eq:app_tachibana_time2}, results in $d_{\gamma_t} \vepsilon_\vtheta = \bm{0}$.

As in~\Cref{s:method}, the total derivative $d_{\gamma_t} \vepsilon_\vtheta$ is composed as 
\begin{align} \label{eq:app_total_derivative}
    d_{\gamma_t} \vepsilon_\vtheta(\rvx_t, t) = \frac{\partial \vepsilon_\vtheta(\rvx_t, t)}{\partial \rvx_t} \frac{d\rvx_t}{d \gamma_t} + \frac{\partial \vepsilon_\vtheta(\rvx_t, t)}{\partial t} \frac{dt}{d\gamma_t}.
\end{align}
Inserting the ``ideal derivative trick'', the above becomes
\begin{align} \label{eq:app_ttm_with_ideal_derivative}
    d_{\gamma_t} \vepsilon_\vtheta(\rvx_t, t) \approx \frac{1}{\sigma_t} \left(\frac{1}{2} \frac{1}{\alpha_t^2} \frac{d\alpha_t^2}{d\gamma_t} \left[\rvx_t - \frac{\vepsilon_\theta(\rvx_t,t)}{\sigma_t} \right] \right) + \left( \frac{\frac{d \log \alpha_t^2}{dt}}{2 \sigma_t} \left(\frac{\vepsilon_\vtheta(\rvx_t, t)}{\sigma_t}  - \rvx_t \right)\right) \frac{dt}{d\gamma_t},
\end{align}
where we have inserted~\Cref{eq:app_dx_dgamma} for $\tfrac{d\rvx_t}{d\gamma_t}$ and used the usual parameterization $\vs_\vtheta(\rvx_t, t) \coloneqq - \frac{\vepsilon_\vtheta(\rvx_t, t)}{\sigma_t}$. Using $\frac{d\log \alpha_t^2}{dt} = \frac{1}{\alpha_t^2} \frac{d\alpha_t^2}{dt}$ and $\frac{d\alpha_t^2}{dt} \frac{dt}{d\gamma_t} = \frac{d\alpha_t^2}{d\gamma_t}$, we can see that the right-hand side of~\Cref{eq:app_ttm_with_ideal_derivative} is $\bm{0}$. Hence, applying the second TTM to the DDIM ODE and using the ``ideal derivative trick'' is equivalent to the first TTM (Euler's method) applied to the DDIM ODE. We believe that this is potentially a reason why the DDIM solver~\citep{song2021denoising}, Euler's method applied to the DDIM ODE, shows such great empirical performance: it can be interpreted as an approximate (``ideal derivative trick'') second order ODE solver. On the other hand, our derivation also implies that the ``ideal derivative trick'' used in the second TTM for the DDIM ODE does not actually provide any benefit over the standard DDIM solver, because all additional second-order terms vanish. Hence, to improve upon regular DDIM, the ``ideal derivative trick'' is insufficient and we need to learn the higher-order score terms more accurately without such coarse approximations, as we do in our work.

Furthermore, it is interesting to show 
that we do not obtain the same cancellation effect when applying the ``ideal derivative trick'' to the Probability Flow ODE in~\Cref{eq:app_approximate_probability_flow_ode}: Let $\vf(\rvx_t, t) = - \frac{1}{2} \beta(t) \left[ \rvx_t - \tfrac{\vepsilon_\vtheta(\rvx_t, t)}{\sigma_t} \right]$ (right-hand side of Probability Flow ODE), then
\begin{align}
    \frac{d\vf}{dt}\rvert_{(\rvx_t, t)} &= \frac{\beta^\prime(t)}{\beta(t)} \vf(\rvx_t, t) - \frac{1}{2} \beta(t) \frac{d}{dt} \left[ \rvx_t - \frac{\vepsilon_\vtheta(\rvx_t, t) }{\sigma_t} \right] \\
    &= \left[\frac{\beta^\prime(t)}{\beta(t)} - \frac{1}{2} \beta(t) \right] \vf(\rvx_t, t) + \frac{1}{2} \beta(t) \left(\frac{\tfrac{d\vepsilon_\vtheta(\rvx_t, t)}{dt} }{\sigma_t} - \sigma_t^{-2} \frac{d\sigma_t}{dt} \vepsilon_\vtheta(\rvx_t, t)\right),
\end{align}
where $\beta^\prime(t) \coloneqq \tfrac{d\beta(t)}{dt}$. Using the ``ideal derivative trick'', we have $\tfrac{d\vepsilon_\vtheta}{dt} = d_{\gamma_t} \vepsilon_\vtheta \, d_t \gamma_t \approx \bm{0}$, and therefore the above becomes
\begin{align}
    \frac{d\vf}{dt} \rvert_{(\rvx_t, t)} \approx \left[\frac{\beta^\prime(t)}{\beta(t)} - \frac{1}{2} \beta(t) \right] \vf(\rvx_t, t) - \frac{\beta(t)}{2 \sigma_t^2} \frac{d\sigma_t}{dt} \vepsilon(\rvx_t, t).
\end{align}
The derivative $\frac{d\sigma_t}{dt}$ can be computed as follows
\begin{align}
    \frac{d\sigma_t}{dt} &= \frac{1}{2\sigma_t}\frac{d\sigma_t^2}{dt} \\
    &= \frac{1}{2\sigma_t} \frac{d}{dt} \left(1-e^{-\int_0^t \beta(t^\prime) \,dt^\prime}\right) \\
    &= \frac{\beta(t) e^{-\int_0^t \beta(t^\prime) \,dt^\prime}}{2\sigma_t}.
\end{align}
Putting everything back together, we have
\begin{align}
    \frac{d\vf}{dt} \rvert_{(\rvx_t, t)} = \left[\frac{\beta^\prime(t)}{2\sigma_t} + \frac{\beta^2(t)}{4\sigma_t} -  \frac{\beta^2(t) e^{-\int_0^t \beta(t^\prime) \,dt^\prime}}{4\sigma_t^3}\right] \vepsilon_\vtheta(\rvx_t, t) + \left[-\frac{\beta^\prime(t)}{2} + \frac{\beta^2(t)}{4} \right] \rvx_t,
\end{align}
which is clearly not $\bm{0}$ for all $\rvx_t$ and $t$. Hence, in contrast to the DDIM ODE, applying Euler's method to the Probability Flow ODE does not lead to an approximate (in the sense of the ``ideal derivative trick'') second order ODE solver.

Note that very related observations have been made in the concurrent works \citet{karras2022elucidating} and \citet{qinsheng2022gddim}. These works notice that when the data distribution consist only of a single data point or a spherical Gaussian distribution, then the solution trajectories of the generative DDIM ODE are straight lines. In fact, this exactly corresponds to our observation that in such a setting we have $d_{\gamma_t} \vepsilon_\vtheta = \bm{0}$, as shown above in the analysis of the ``ideal derivatives approximation''. Note in that context that our above derivation considers the ``single data point'' distribution assumption,
but also applies to the setting where the data is a spherical normal distribution (only $\sigma_t$ would be different, which would not affect the derivation).

\subsection{3rd TTM Applied to the DDIM ODE}
As promised in~\Cref{s:method}, we show here how to apply the third TTM to the DDIM ODE, resulting in the following scheme:
\begin{align}
    \bar \rvx_{t_{n+1}} = \bar \rvx_{t_n} + h_n \vepsilon_\vtheta(\rvx_{t_n}, t_n) + \frac{1}{2} h_n^2 \frac{d\vepsilon_\vtheta}{d\gamma_t} \rvert_{(\rvx_{t_n}, t_n)} + \frac{1}{6} h_n^3 \frac{d^2\vepsilon_\vtheta}{d\gamma_t^2} \rvert_{(\rvx_{t_n}, t_n)},
\end{align}
where $h_n = (\gamma_{t_{n+1}} - \gamma_{t_n})$. In the remainder of this section, we derive a computable formula for $\frac{d^2\vepsilon_\vtheta}{d\gamma_t^2}$, only containing partial derivatives.

Using the chain rule, we have
\begin{align}
    \frac{d^2 \vepsilon_\vtheta}{d\gamma_t^2}\rvert_{(\rvx_t, t)} &= \frac{\partial d_\gamma \vepsilon_\vtheta(\rvx_t, t)}{\partial \rvx_t} \frac{d\rvx_t}{d \gamma_t} + \frac{\partial d_\gamma \vepsilon_\vtheta(\rvx_t, t)}{\partial t} \frac{dt}{d\gamma_t},
\end{align}
where, using~\Cref{eq:app_total_derivative},
\begin{align}
    \frac{\partial d_\gamma \vepsilon_\vtheta(\rvx_t, t)}{\partial \rvx_t} = \frac{\partial^2 \vepsilon_\vtheta(\rvx_t, t)}{\partial \rvx^2} \frac{d\rvx_t}{d \gamma_t} + \frac{\partial \vepsilon_\vtheta(\rvx_t, t)}{\partial \rvx_t} \left(\frac{1}{\sqrt{\gamma_t^2 + 1}} \frac{\partial \vepsilon_\vtheta(\rvx_t, t)}{\partial \rvx_t} - \frac{\gamma_t}{1 + \gamma_t^2} \mI \right) + \frac{\partial^2 \vepsilon_\vtheta(\rvx_t, t)}{\partial t \partial \rvx_t} \frac{dt}{d\gamma_t},
\end{align}
and
\begin{align} \label{eq:app_third_time_derivative}
     \frac{\partial d_\gamma \vepsilon_\vtheta(\rvx_t, t)}{\partial t} = \frac{\partial}{\partial t} \left(\frac{\partial \vepsilon_\vtheta(\rvx_t, t)}{\partial \rvx_t} \frac{d\rvx_t}{d\gamma_t} \right) + \frac{\partial}{\partial t} \left(\frac{\partial \vepsilon_\vtheta(\rvx_t, t)}{\partial t} \frac{dt}{d\gamma_t} \right).
\end{align}
The remaining terms in~\Cref{eq:app_third_time_derivative} can be computed as
\begin{align}
    \frac{\partial}{\partial t} \left(\frac{\partial \vepsilon_\vtheta(\rvx_t, t)}{\partial t} \frac{dt}{d\gamma_t} \right) = \frac{\partial^2 \vepsilon_\vtheta(\rvx_t, t)}{\partial t^2} \frac{dt}{d\gamma_t} + \frac{\partial \vepsilon_\vtheta(\rvx_t, t)}{\partial t} \frac{d \left(\tfrac{dt}{d\gamma_t}\right)}{d t},
\end{align}
and
\begin{align}
    \frac{\partial}{\partial t} \left(\frac{\partial \vepsilon_\vtheta(\rvx_t, t)}{\partial \rvx_t} \frac{d\rvx_t}{d\gamma_t} \right) = \frac{\partial^2 \vepsilon_\vtheta(\rvx_t, t)}{\partial t \, \partial \rvx_t}\frac{d\rvx_t}{d\gamma_t} + \frac{\partial \vepsilon_\vtheta(\rvx_t, t)}{\partial \rvx_t} \frac{\partial \left(\tfrac{d\rvx_t}{d\gamma_t}\right)}{\partial t}
\end{align}
where, inserting~\Cref{eq:app_dx_dgamma2} for $\tfrac{d\rvx_t}{d\gamma_t}$ as well as using the usual parameterization $\vs_\vtheta(\rvx_t, t) \coloneqq - \frac{\vepsilon_\vtheta(\rvx_t, t)}{\sigma_t}$,
\begin{align}
    \frac{\partial \left(\tfrac{d\rvx_t}{d\gamma_t}\right)}{\partial t} &= \frac{\partial}{\partial t} \left(- \frac{\gamma_t}{\gamma_t^2 + 1} \left[\rvx_t -\tfrac{ \vepsilon_\vtheta(\rvx_t, t)}{\sigma_t} \right]\right) \\
    &= \frac{\partial \left(\frac{1}{\sqrt{\gamma_t^2 + 1}}\right)}{\partial t} \vepsilon_\vtheta(\rvx_t, t) + \frac{1}{\sqrt{\gamma_t^2 + 1}} \frac{\partial \vepsilon_\vtheta(\rvx_t, t)}{\partial t} - \frac{\partial \left(\frac{\gamma_t}{1 + \gamma_t^2}\right)}{\partial t} \rvx_t \quad \left(\text{using \,} \sigma_t = \frac{\gamma_t}{\sqrt{\gamma_t^2 + 1}}\right) \\
    &= \left(- \frac{\gamma_t}{\left(\gamma_t^2 + 1\right)^{3/2}} \vepsilon_\vtheta(\rvx_t, t) + \frac{\gamma_t^2 - 1}{\left(\gamma_t^2 + 1\right)^2} \rvx_t\right) \frac{d\gamma_t}{dt} + \frac{1}{\sqrt{\gamma_t^2 + 1}} \frac{\partial \vepsilon_\vtheta(\rvx_t, t)}{\partial t}.
\end{align}
We now have a formula for $\frac{d^2\vepsilon_\vtheta}{d\gamma_t^2}$ containing only partial derivatives, and therefore we can compute $\frac{d^2\vepsilon_\vtheta}{d\gamma_t^2}$ using automatic differentiation. Note that we could follow the same procedure to compute even higher derivatives of $\vepsilon_\vtheta$.

\begin{figure}
    \centering
    \includegraphics[scale=0.85]{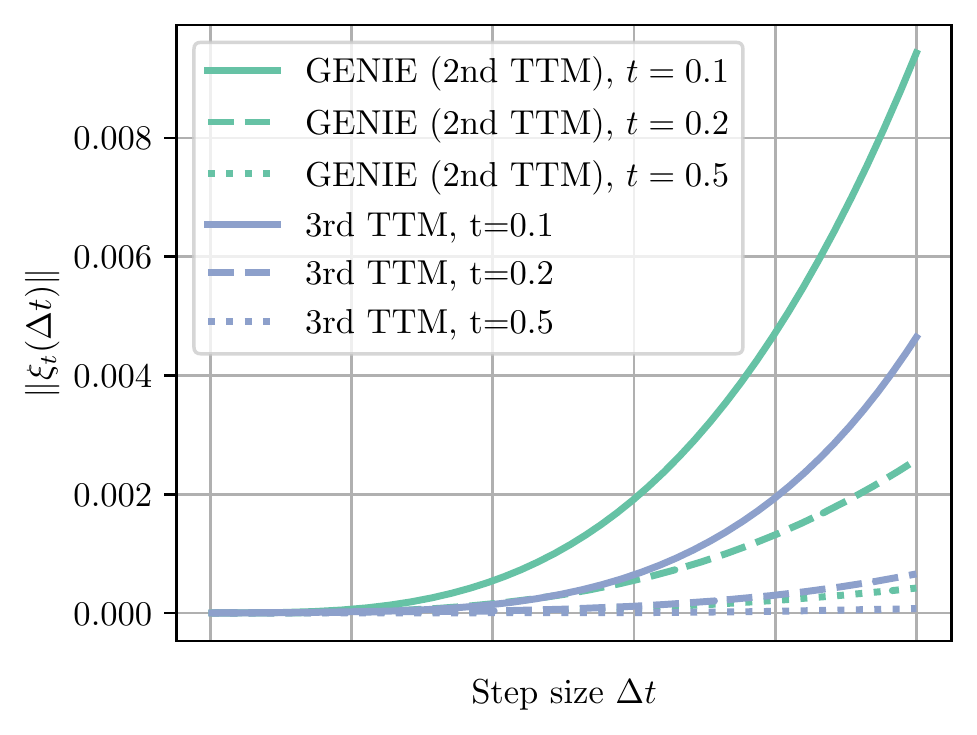}
    \caption{Single step error using analytical score function. See also~\Cref{fig:analytical_error} (\textit{top}).}
    \label{fig:app_everything_w_third_order}
\end{figure}
We repeat the 2D toy distribution single step error experiment from~\Cref{s:method} (see also~\Cref{fig:analytical_error} (\textit{top}) and~\Cref{s:app_toy_experiments} for details). As expected, in~\Cref{fig:app_everything_w_third_order} we can clearly see that the third TTM improves upon the second TTM.

\begin{figure}
    \centering
    \includegraphics[scale=0.95]{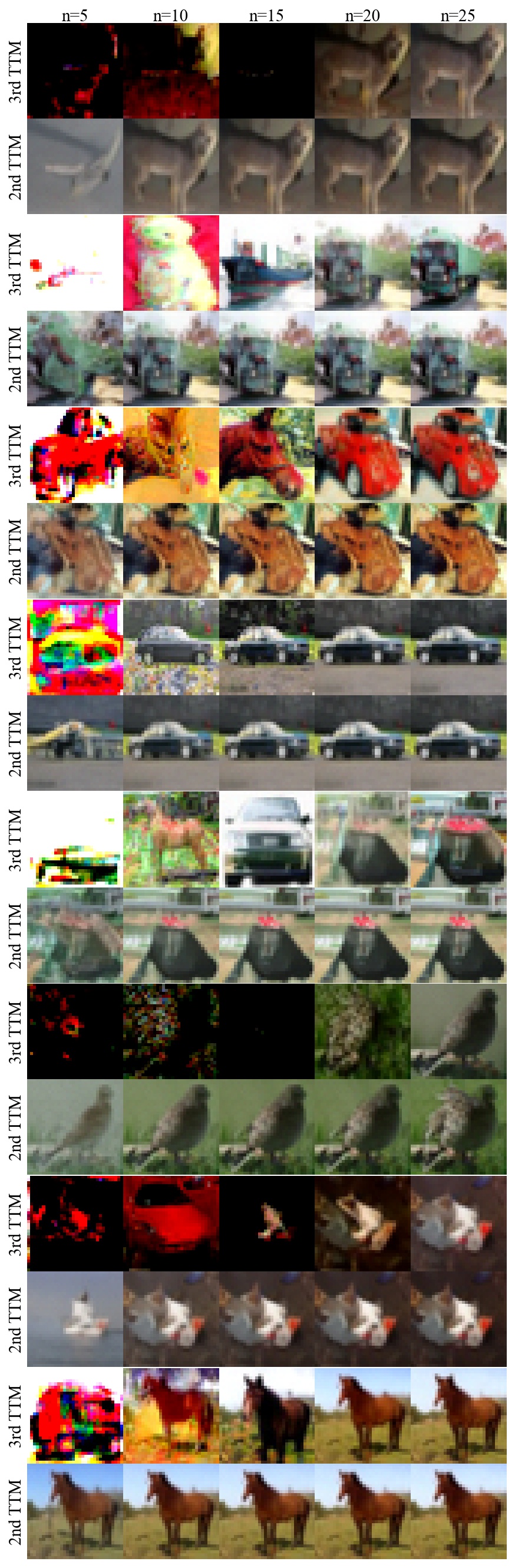}
    \caption{Qualitative comparison of the second and the third TTMs applied to the DDIM ODE on CIFAR-10 (all necessary derivatives calculated with automatic differentiation). The number of steps in the ODE solver is denoted as n.}
    \label{fig:app_cifar_backpropagation}
\end{figure}

In~\Cref{fig:app_cifar_backpropagation}, we compare the second TTM to the third TTM applied to the DDIM ODE on CIFAR-10. Both for the second and the third TTM, we compute all partial derivatives using automatic differentiation (without distillation). It appears that for using 15 or less steps in the ODE solver, the second TTM performs better than the third TTM. We believe that this could potentially be due to our score model $s_\vtheta(\rvx_t, t)$ not being accurate enough, in contrast to the above 2D toy distribution experiment, where we have access to the analytical score function. Furthermore, note that when we train $s_\vtheta(\rvx_t, t)$ via score matching, we never regularize (higher-order) derivatives of the neural network, and therefore there is no incentive for them to be well-behaved. It would be interesting to see if, besides having more accurate score models, regularization techniques such as spectral regularization~\citep{miyato2018spectral} could potentially alleviate this issue. Also the higher-order score matching techniques derived by~\citet{meng2021estimating} could help to learn higher-order derivates of the score functions more accurately. We leave this exploration to future work.

\subsection{GENIE is Consistent and Principled} \label{app:consistency}
GENIE is a consistent and principled approach to developing a higher-order ODE solver for sampling from diffusion models: GENIE's design consists of two parts: (1) We are building on the second Truncated Taylor Method (TTM), which is a well-studied ODE solver (see \citet{kloeden1992stochastic}) with provable local and global truncation errors (see also~\Cref{app:theoreticalbounds}). Therefore, if during inference we had access to the ground truth second-order ODE derivatives, which are required for the second TTM, GENIE would simply correspond to the \textit{exact} second TTM.
    
(2) In principle, we could calculate the exact second-order derivatives \textit{during inference} using automatic differentiation. However, this is too slow for competitive sampling speeds, as it requires additional backward passes through the first-order score network. Therefore, in practice, we use the learned prediction heads $\mathbf{k}_\psi(\mathbf{x}_t, t)$.

Consequently, if $\mathbf{k}_\psi(\mathbf{x}_t, t)$ modeled the ground truth second-order derivatives exactly, i.e.  $\mathbf{k}_\psi(\mathbf{x}_t, t) = d_{\gamma_t} \mathbf{\epsilon}_\theta(\mathbf{x}_t, t)$ for all $\mathbf{x}_t$ and $t$, we would obtain a rigorous second-order solver based on the TTM, following (1) above.
    
In practice, distillation will not be perfect. However, given the above analysis, optimizing a neural network $\mathbf{k}_\psi(\mathbf{x}_t, t)$ towards $d_{\gamma_t} \mathbf{\epsilon}_\theta(\mathbf{x}_t, t)$ is well motivated and theoretically grounded. In particular, \textit{during training} we are calculating \textit{exact} ODE gradients using automatic differentiation on the first-order score model as distillation targets. Therefore, in the limit of infinite neural network capacity and perfect optimization, we could in theory minimize our distillation objective function (\Cref{eq:objective}) perfectly and obtain $\mathbf{k}_\psi(\mathbf{x}_t, t) = d_{\gamma_t} \mathbf{\epsilon}_\theta(\mathbf{x}_t, t)$.
    
Also recall that regular denoising score matching itself, on which all diffusion models rely, follows the exact same argument. In particular, denoising score matching also minimizes a ``simple'' (weighted) $L_2$-loss between a trainable score model $\mathbf{s}_\theta(\mathbf{x}_t, t)$ and the spatial derivative of the log-perturbation kernel, i.e., $\nabla_{\mathbf{x}_t} \log p_t(\mathbf{x}_t \mid \mathbf{x}_0)$. From this perspective, denoising score matching itself also simply tries to ``distill'' (spatial) derivatives into a model. If we perfectly optimized the denoising score matching objective, we would obtain a diffusion model that models the data distribution exactly, but in practice, similar to GENIE, we never achieve that due to imperfect optimization and finite-capacity neural networks. Nevertheless, denoising score matching similarly is a well-defined and principled method, precisely because of that theoretical limit in which the distribution can be reproduced exactly.

We would also like to point out that other, established higher-order methods for diffusion model sampling with the generative ODE, such as linear multistep methods~\citep{liu2022pseudo}, make approximations, too, which can be worse in fact. In particular, multistep methods \emph{always} approximate higher-order derivatives in the TTM using finite differences which is crude for large step sizes, as can be seen in Fig.~3 (\textit{bottom}). From this perspective, if our distillation is sufficiently accurate, GENIE can be expected to be more accurate than such multistep methods.
\newpage
\section{Model and Implementation Details} \label{s:app_model_details}

\subsection{Score Models} \label{s:app_training_score_models}
We train \emph{variance-preserving} DDMs~\citep{song2020} for which $\sigma_t^2 = 1 - \alpha_t^2$. We follow~\citet{song2020} and set $\beta(t) = 0.1 + 19.9 t$; note that $\alpha_t = e^{-\tfrac{1}{2} \int_0^t \beta(t^\prime) \, dt^\prime}$. All score models are parameterized as either $\vs_\vtheta(\rvx_t, t) \coloneqq - \frac{\vepsilon_\vtheta(\rvx_t, t)}{\sigma_t}$ ($\vepsilon$-prediction) or $\vs_\vtheta(\rvx_t, t) \coloneqq - \frac{\alpha_t \rvv_\vtheta(\rvx_t, t) + \sigma_t \rvx_t}{\sigma_t}$ ($\rvv$-prediction), where $\vepsilon_\vtheta(\rvx_t, t)$ and $\rvv_\vtheta(\rvx_t, t)$ are U-Nets~\citep{ronneberger2015u}. The $\vepsilon$-prediction model is trained using the following score matching objective~\citep{ho2020}
\begin{align}
    \min_\vtheta\, \E_{t \sim \gU[t_\mathrm{cutoff}, 1], \rvx_0 \sim p(\rvx_0), \vepsilon \sim \gN(\bm{0}, \mI)} \left[\| \vepsilon - \vepsilon_\vtheta(\rvx_t, t) \|_2^2\right], \quad \rvx_t = \alpha_t \rvx_0 + \sigma_t \vepsilon.
\end{align}
The $\rvv$-prediction model is trained using the following score matching objective~\citep{salimans2022progressive}
\begin{align}
    \min_\vtheta\, \E_{t \sim \gU[t_\mathrm{cutoff}, 1], \rvx_0 \sim p(\rvx_0), \vepsilon \sim \gN(\bm{0}, \mI)} \left[\| \tfrac{\vepsilon - \sigma_t \rvx_t}{\alpha_t} - \rvv_\vtheta(\rvx_t, t) \|_2^2\right], \quad \rvx_t = \alpha_t \rvx_0 + \sigma_t \vepsilon,
\end{align}
which is referred to as ``SNR+1'' weighting~\citep{salimans2022progressive}. The neural network $\rvv_\vtheta$ is now effectively tasked with predicting $\rvv \coloneqq \alpha_t\vepsilon-\sigma_t\rvx_0$.

\textbf{CIFAR-10:} On this dataset, we do not train our own score model, but rather use a checkpoint\footnote{The checkpoint can be found at~\url{https://drive.google.com/file/d/16_-Ahc6ImZV5ClUc0vM5Iivf8OJ1VSif/view?usp=sharing}.} provided by~\citet{song2020}. The model is based on the DDPM++ architecture introduced in~\citet{song2020} and predicts $\vepsilon_\vtheta$. 

\textbf{LSUN Bedrooms and LSUN Church-Outdoor:} Both datasets use exactly the same model structure. The model structure is based on the DDPM architecture introduced in~\citet{ho2020} and predicts $\vepsilon_\vtheta$.

\textbf{ImageNet:} This model is based on the architecture introduced in~\citet{dhariwal2021diffusion}. We make a small change to the architecture  and replace its sinusoidal time embedding by a Gaussian Fourier projection time embedding~\citep{song2020}. The model is class-conditional and we follow~\citet{dhariwal2021diffusion} and simply add the class embedding to the (Gaussian Fourier projection) time embedding. The model predicts $\vepsilon_\vtheta$.

\textbf{Cats (Base):} This model is based on the architecture introduced in~\citet{dhariwal2021diffusion}. We make a small change to the architecture  and replace its sinusoidal time embedding by a Gaussian Fourier projection time embedding~\citep{song2020}. The model predicts $\rvv_\vtheta$.

\textbf{Cats (Upsampler):} This model is based on the architecture introduced in~\citet{dhariwal2021diffusion}. We make a small change to the architecture  and replace its sinusoidal time embedding by a Gaussian Fourier projection time embedding~\citep{song2020}. The upsampler is conditioned on noisy upscaled lower-resolution images, which are concatenated to the regular channels that form the synthesized outputs of the diffusion model. Therefore, we expand the number of input channels from three to six. We use augmentation conditioning~\citep{saharia2022photorealistic} to noise the lower-resolution image. In particular, we upscale $\alpha_{t^\prime} \rvx_\mathrm{low} + \sigma_{t^\prime} \rvz$, where $\rvx_\mathrm{low}$ is the clean lower-resolution image. During training $t^\prime$ is sampled from $\gU[t_\mathrm{cutoff}, 1]$. During inference, $t^\prime$ is a hyper-parameter which we set to $0.1$ for all experiments.

We use two-independent Gaussian Fourier projection embeddings for $t$ and $t^\prime$ and concatenate them before feeding them into the layers of the U-Net.

\textbf{Model Hyperparameters and Training Details}: All model hyperparameters and training details can be found in~\Cref{tab:app_score_model_hyperparameter}.
\begin{table}
    \centering
    \caption{Model hyperparameters and training details. The CIFAR-10 model is taken from~\citet{song2020}; all other models are trained by ourselves.}
    \scalebox{0.65}{\begin{tabular}{l c c c c c c}
        \toprule
         Hyperparameter & CIFAR-10 & LSUN Bedrooms & LSUN Church-Outdoor & ImageNet & Cats (Base) & Cats (Upsampler)\\
         \midrule
         \textbf{Model} & \\
         Data dimensionality (in pixels) & 32 & 128 & 128 & 64 & 128 & 512\\
         Residual blocks per resolution & 8 & 2 & 2 & 3 & 2 & 2\\
         Attention resolutions & 16 & 16 & 16 & 8 & (8, 16) & (8, 16) \\
         Base channels & 128 & 128 & 128 & 192 & 96 & 192\\
         Channel multipliers & 1,2,3,4 & 1,1,2,2,4,4,4 & 1,1,2,2,4,4,4 & 1,2,3,4 & 1,2,2,3,3 &
         1,1,2,2,3,3,4\\
         EMA rate & 0.9999 & 0.9999 & 0.9999 & 0.9999 & 0.9999 & 0.9999 \\
         \# of head channels & N/A & N/A & N/A & 64 & 64 & 64 \\
         \# of parameters & 107M & 148M & 148M & 283M & 200M & 80.2M \\
         Base architecture & DDPM++~\citep{song2020} & DDPM~\citep{ho2020} & DDPM~\citep{ho2020} & \citep{dhariwal2021diffusion} & \citep{dhariwal2021diffusion} & \citep{dhariwal2021diffusion} \\
         Prediction & $\vepsilon$ & $\vepsilon$ & $\vepsilon$ & $\vepsilon$ & $\rvv$ & $\rvv$ \\
         \midrule
         \textbf{Training} & \\
         \# of iterations & 400k & 300k & 300k & 400k & 400k & 150k \\
         \# of learning rate warmup iterations & 100k & 100k & 100k & 100k & 100k & 100k \\
         Optimizer & Adam & Adam & Adam & Adam & Adam & Adam  \\
         Mixed precision training & \xmark & \cmark & \cmark & \cmark & \cmark & \cmark \\
         Learning rate & $10^{-4}$ & $3\cdot 10^{-4}$ & $3\cdot 10^{-4}$ & $2\cdot 10^{-4}$ & $10^{-4}$ & $10^{-4}$\\
         Gradient norm clipping & 1.0 & 1.0 & 1.0 & 1.0 & 1.0 & 1.0 \\
         Dropout & 0.1 & 0.0 & 0.0 & 0.1 & 0.1 & 0.1 \\
         Batch size & 128 & 256 & 256 & 1024 & 128 & 64 \\
         $t_\mathrm{cutoff}$ & $10^{-5}$ & $10^{-3}$ & $10^{-3}$ & $10^{-3}$ & $10^{-3}$ & $10^{-3}$ \\
         \bottomrule
    \end{tabular}}
    \label{tab:app_score_model_hyperparameter}
\end{table}

\subsection{Prediction Heads}
\label{s:app_training_prediction_heads}
We model the derivative $d_{\gamma_t} \vepsilon_\vtheta$ using a small prediction head $\vk_\vpsi$ on top of the first-order score model~$\vepsilon_\vtheta$. In particular, we provide the last feature layer from the $\vepsilon_\vtheta$ network together with its time embedding as well as $\rvx_t$ and the output of $\vepsilon(\rvx_t, t)$ to the prediction head~(see~\Cref{fig:blocks} for a visualization). We found modeling $d_{\gamma_t} \vepsilon_\vtheta$ to be effective even for our Cats models that learn to predict $\rvv = \alpha_t \vepsilon - \sigma_t \rvx_0$ rather than $\vepsilon$. Directly learning $d_{\gamma_t} \rvv_\vtheta$ and adapting the mixed network parameterization (see~\Cref{s:app_mixed_network_parameterization}) could potentially improve results further. We leave this exploration to future work.

We provide additional details on our architecture next.
\subsubsection{Model Architecture}
The architecture of our prediction heads is based on (modified) BigGAN residual blocks~\citep{song2020, brock2018large}. To minimize computational overhead, we only use a single residual block. 

In particular, we concatenate the last feature layer with $\rvx_t$ as well as $\vepsilon_\vtheta(\rvx_t, t)$ and feed it into a convolutional layer. For the upsampler, we also condition on the noisy up-scaled lower resolution image. We experimented with normalizing the feature layer before concatenation. The output of the convolutional layer as well as the time embedding are then fed to the residual block.
Similar to U-Nets used in score models, we normalize the output of the residual block and apply an activation function. Lastly, the signal is fed to another convolutional layer that brings the number of channels to a desired value (in our case nine, three for each $\vk_\vpsi^{(i)}$, $i \in \{1, 2, 3\}$, in~\Cref{eq:app_mixed_network}).

All model hyperparameters can be found in~\Cref{tab:app_prediction_heads_hyperparameter}. We also include the additional computational overhead induced by the prediction heads in~\Cref{tab:app_prediction_heads_hyperparameter}; see~\Cref{s:app_measuring_computational_overhead} for details on how we measured the overhead.

\subsubsection{Training Details}

We train for 50k iterations using Adam~\citep{kingma2015adam}. We experimented with two base learning rates: $10^{-4}$ and $5 \cdot 10^{-5}$. We furthermore tried two ``optimization setups'': (linearly) warming up the learning rate in the first 10k iterations (score models are often trained by warming up the learning rate in the first 100k iterations) or, following~\citet{salimans2022progressive}, linearly decaying the learning rate to 0 in the entire 50k iterations of training; we respectively refer to these two setups as ``warmup'' and ``decay''. We measure the FID every 5k iterations and use the best checkpoint.

Note that we have to compute the Jacobian-vector products in~\Cref{eq:total_derivative_full} via automatic differentiation during training. We repeatedly found that computing the derivative $\tfrac{\partial \vepsilon_\vtheta(\rvx_t, t)}{\partial t}$ via automatic differentiation leads to numerical instability (\texttt{NaN}) for small $t$ when using mixed precision training. For simplicity, we turned off mixed precision training altogether. However, training performance could have been optimized by only turning off mixed precision training for the derivative $\tfrac{\partial \vepsilon_\vtheta(\rvx_t, t)}{\partial t}$.
 
All training details can be found in~\Cref{tab:app_prediction_heads_hyperparameter}.

\begin{table}
    \centering
    \caption{Model hyperparameters and training details for the prediction heads.}
    \scalebox{0.7}{\begin{tabular}{l c c c c c c}
        \toprule
         Hyperparameter & CIFAR-10 & LSUN Bedrooms & LSUN Church-Outdoor & ImageNet & Cats (Base) & Cats (Upsampler) \\
         \midrule
         \textbf{Model} & \\
         Data dimensionality & 32 & 128 & 128 & 64 & 128 & 512 \\
         EMA rate & 0 & 0 & 0 & 0 & 0 & 0\\
         Number of channels & 128 & 128 & 128 & 196 & 196 & 92 \\
         \# of parameters & 526k & 526k & 526k & 1.17M & 1.17M & 302k \\
         Normalize $\rvx_\mathrm{embed}$ & \xmark & \xmark & \cmark & \xmark & \xmark & \xmark \\
         \midrule
         \textbf{Training} & \\
         \# of iterations & 20k & 40k & 35k & 15k & 20k & 20k \\
         Optimizer & Adam & Adam & Adam & Adam & Adam & Adam \\
         Optimization setup & Decay & Warmup & Warmup & Warmup &  Warmup & Warmup  \\
         Mixed precision training & \xmark & \xmark & \xmark & \xmark & \xmark & \xmark \\
         Learning rate & $5\cdot 10^{-5}$ & $10^{-4}$ & $10^{-4}$ & $10^{-4}$ & $10^{-4}$ & $10^{-4}$ \\
         Gradient norm clipping & 1.0 & 1.0 & 1.0 & 1.0 & 1.0 & 1.0 \\
         Dropout & 0.0 & 0.0 & 0.0 & 0.0 & 0.0 & 0.0 \\
         Batch size & 128 & 256 & 256 & 256 & 64 & 16 \\
         $t_\mathrm{cutoff}$ & $10^{-3}$ & $10^{-3}$ & $10^{-3}$ & $10^{-3}$ & $10^{-3}$ & $10^{-3}$ \\
         \midrule
         \textbf{Inference} & \\
         Add. comp. overhead & 1.47\% & 14.0\% & 14.4\% & 2.83\% & 7.55\% & 13.3\% \\
         \bottomrule
    \end{tabular}}
    \label{tab:app_prediction_heads_hyperparameter}
\end{table}

\subsubsection{Mixed Network Parameterization} \label{s:app_mixed_network_parameterization}
Our mixed network parameterization is derived from a simple single data point assumption, i.e., $p_t(\rvx_t) = \gN(\rvx_t; \bm{0}, \sigma_t^2 \mI)$. This assumption leads to $\vepsilon_\vtheta(\rvx_t, t) \approx \frac{\rvx_t}{\sigma_t}$ which we can plug into the three terms of~\Cref{eq:total_derivative_full}:
\begin{align} \label{eq:mixed_network_deriv1}
    \frac{1}{\sqrt{\gamma_t^2 + 1}} \frac{\partial \vepsilon_\vtheta(\rvx_t, t)}{\partial \rvx_t} \vepsilon_\vtheta(\rvx_t, t) \approx \frac{1}{\sqrt{\gamma_t^2 + 1}} \frac{\rvx_t}{\sigma_t^2} = \frac{1}{\gamma_t} \frac{\rvx_t}{\sigma_t},
\end{align}
and
\begin{align}\label{eq:mixed_network_deriv2}
    - \frac{\gamma_t}{1 + \gamma_t^2} \frac{\partial \vepsilon_\vtheta(\rvx_t, t)}{\partial \rvx_t} \rvx_t \approx - \frac{\gamma_t}{\sigma_t \left(1 + \gamma_t^2\right)} \rvx_t = - \frac{\gamma_t}{1 + \gamma_t^2} \frac{\rvx_t}{\sigma_t},
\end{align}
and finally
\begin{align}\label{eq:mixed_network_deriv3}
    \frac{\partial \vepsilon_\vtheta(\rvx_t, t)}{\partial t} \frac{dt}{d\gamma_t} \approx -\frac{\rvx_t}{\sigma_t^2} \frac{d\sigma_t}{dt} \frac{dt}{d\gamma_t} = -\frac{\gamma_t^2 + 1}{\gamma_t^2} \rvx_t \frac{1}{\left(\gamma_t^2 + 1\right)^{3/2}} = - \frac{1}{\gamma_t (1 + \gamma_t^2)} \frac{\rvx_t}{\sigma_t},
\end{align}
where we have used $\sigma_t = \frac{\gamma_t}{\sqrt{\gamma_t^2 + 1}}$. This derivation therefore implies the following mixed network parameterization
\begin{align} \label{eq:app_mixed_network}
    \vk_\vpsi = -\frac{1}{\gamma_t} \vk^{(1)}_\vpsi + \frac{\gamma_t}{1 + \gamma_t^2} \vk^{(2)}_\vpsi + \frac{1}{\gamma_t (1 + \gamma_t^2)} \vk^{(3)}_\vpsi \approx d_{\gamma_t}\vepsilon_\vtheta,
\end{align}
where $\vk_\vpsi^{(i)}(\rvx_t,t)$, $i\in \{1,2,3\}$, are different output channels of the neural network (i.e. the additional head on top of the $\vepsilon_\vtheta$ network). To provide additional intuition, we basically replaced the $-\frac{\rvx_t}{\sigma_t}$ terms in \Cref{eq:mixed_network_deriv1,eq:mixed_network_deriv2,eq:mixed_network_deriv3} by neural networks. However, we know that for approximately Normal data $\frac{\rvx_t}{\sigma_t}\approx\vepsilon_\vtheta(\rvx_t, t)$, where $\vepsilon_\vtheta(\rvx_t, t)$ predicts ``noise'' values $\vepsilon$ that were drawn from a standard Normal distribution and are therefore varying on a well-behaved scale. Consequently, up to the Normal data assumption, we can also expect our prediction heads $\vk_\vpsi^{(i)}(\rvx_t,t)$ in the parameterization in \Cref{eq:app_mixed_network} to predict well-behaved output values, which should make training stable. This \textit{mixed network parameterization} approach is inspired by the mixed score parameterization from~\citet{vahdat2021score} and~\citet{dockhorn2022scorebased}.
\subsubsection{Pseudocode} \label{s:app_ph_pseudocode}
In this section, we provide pseudocode for training our prediction heads $\vk_\vpsi$ and using them for sampling with GENIE. In~\Cref{alg:app_traininig}, the analytical $\frac{dt}{d\gamma_t}$ is an implicit hyperparameter of the DDM as it depends on $\alpha_t$. For our choice of $\alpha_t = e^{-\tfrac{1}{2} \int_0^t 0.1 + 19.9t^\prime \, dt^\prime}$ (see~\Cref{s:app_training_score_models}), we have
\begin{align}
    \frac{dt}{d\gamma_t} = \frac{\frac{2 \gamma_t}{19.9 \left(\gamma_t^2 + 1\right)}}{\sqrt{\left(\frac{0.1}{19.9}\right)^2 + \frac{2 \log \left( 
    \gamma_t^2 + 1\right)}{19.9}}},
\end{align}
where $\gamma_t = \sqrt{\frac{1-\alpha_t^2}{\alpha_t^2}}$.

In~\Cref{alg:app_genie}, we are free to use any time discretization $t_0 = 1.0 > t_1 > \dots > t_{N} = t_\mathrm{cutoff}$. When referring to ``linear striding'' in this work, we mean the time discretization $t_n = 1.0 - (1.0-t_\mathrm{cutoff}) \frac{n}{N}$. When referring to ``quadratic striding'' in this work, we mean the time discretization $t_n = \left(1.0 - (1.0-\sqrt{t_\mathrm{cutoff}}) \frac{n}{N}\right)^2$.
\begin{algorithm}[H]
\small
\caption{Training prediction heads $\vk_\vpsi$%
}
\label{alg:app_traininig}
\begin{algorithmic}

\State {\bfseries Input:} Score model $\vs_\vtheta \coloneqq - \tfrac{\vepsilon_\vtheta(\rvx_t, t)}{\sigma_t}$, number of training iterations $N$.
\State {\bfseries Output:} Trained prediction head $\vk_\vpsi$.
\State
\For{$n=1$ {\bfseries to} $N$}
\State Sample $\rvx_0 \sim p_0(\rvx_0)$, $t \sim \gU[t_\mathrm{cutoff}, 1], \vepsilon \sim \gN(\bm{0}, \mI)$
\State Set $\rvx_t = \alpha_t \rvx_0 + \sigma_t \vepsilon$
\State Compute $\vepsilon_\vtheta(\rvx_t, t)$
\State Compute the exact spatial Jacobian-vector product $\mathrm{JVP}_\mathrm{s} = \frac{\partial \vepsilon_\vtheta(\rvx_t, t)}{\partial \rvx_t} \left(\frac{1}{\sqrt{\gamma_t^2 + 1}} \vepsilon_\vtheta(\rvx_t, t) - \frac{\gamma_t}{1 + \gamma_t^2} \rvx_t \right)$ via automatic differentiation
\State Compute the exact temporal Jacobian-vector product $\mathrm{JVP}_\mathrm{t} = \frac{\partial \vepsilon_\vtheta(\rvx_t, t)}{\partial t} \frac{dt}{d\gamma_t}$ via automatic differentiation ($\frac{dt}{d\gamma_t}$ can be computed analytically)
\State Compute $\vk_\vpsi(\rvx_t, t)$ using the mixed parameterization in~\Cref{eq:app_mixed_network} %
\State Update weights $\vpsi$ to minimize $\gamma_t^2 \|\vk_\vpsi(\rvx_t, t) - d_{\gamma_t} \vepsilon_\vtheta(\rvx_t, t) \|_2^2 $, where $d_{\gamma_t} \vepsilon_\vtheta(\rvx_t, t) = \mathrm{JVP}_\mathrm{s} - \mathrm{JVP}_\mathrm{t}$
\EndFor
\end{algorithmic}
\end{algorithm}

\begin{algorithm}[H]
\small
\caption{GENIE sampling%
}
\begin{algorithmic} \label{alg:app_genie}

\State {\bfseries Input:} Score model $\vs_\vtheta \coloneqq - \tfrac{\vepsilon_\vtheta(\rvx_t, t)}{\sigma_t}$, prediction head $\vk_\vpsi$, number of sampler steps $N$, time discretization $\{t_n\}_{n=0}^{N}$.
\State {\bfseries Output:} Generated GENIE output sample $\rvy$.
\State Sample $\rvx_{t_0} \sim \gN(\bm{0}, \mI)$
\State Set $\bar \rvx_{t_0} = \sqrt{1 + \gamma_{t_0}^2} \rvx_{t_0}$ \hspace{4.7cm} $\triangleright$ \textit{Note that $\bar \rvx_{t_n} = \sqrt{1 + \gamma_{t_n} ^2} \rvx_{t_n} $ for all ${t_n} $} 
\For{$n=0$ {\bfseries to} $N-1$}
\If{AFS and $n=0$}
    \State $\bar \rvx_{t_{n+1}} = \bar \rvx_{t_n} + (\gamma_{t_{n+1}} - \gamma_{t_n}) \rvx_{t_n}$
\Else
    \State $\bar \rvx_{t_{n+1}} = \bar \rvx_{t_n} + (\gamma_{t_{n+1}} - \gamma_{t_n}) \vepsilon_\vtheta(\rvx_{t_n}, t_n) + \frac{1}{2} (\gamma_{t_{n+1}} - \gamma_{t_n})^2 \vk_\vpsi(\rvx_{t_n}, t_n)$
\EndIf
\State $\rvx_{t_{n+1}} = \frac{\bar \rvx_{t_{n+1}}}{\sqrt{1 + \gamma^2_{t_{n+1}}}}$
\EndFor
\If{Denoising}
    \State $\rvy = \frac{\rvx_{t_{N}} - \sigma_{t_{N}} \vepsilon_\vtheta(\rvx_{t_{N}}, t_{N})}{\alpha_{t_{N}}}$
\Else
    \State $\rvy = \rvx_{t_N}$
\EndIf
\end{algorithmic}
\end{algorithm}

\subsubsection{Measuring Computational Overhead} \label{s:app_measuring_computational_overhead}
Our prediction heads induce a slight computational overhead since their forward pass has to occur after the forward pass of the score model. We measure the overhead as follows: first, we measure the inference time of the score model itself. We do five forward passes to ``warm-up'' the model and then subsequently synchronize via~\texttt{torch.cuda.synchronize()}. We then measure the total wall-clock time of 50 forward passes. We then repeat this process using a combined forward pass: first the score model and subsequently the prediction head. We choose the batch size to (almost) fill the entire GPU memory. In particular we chose batch sizes of 512, 128, 128, 64, 64, and 8, for CIFAR-10, LSUN Bedrooms, LSUN Church-Outdoor, ImageNet, Cats (base), and Cats (upsampler), respectively. The computational overhead for each model is reported in~\Cref{tab:app_prediction_heads_hyperparameter}. This measurement was carried out on a single NVIDIA 3080 Ti GPU.

\newpage
\section{Learning Higher-Order Gradients without Automatic Differentiation and Distillation}
\label{s:app_meng}
In this work, we learn the derivative $d_{\gamma_t} \vepsilon_\vtheta$, which includes a spatial and a temporal Jacobian-vector product, by distillation based on automatic differentiation (AD). We now derive an alternative learning objective for the spatial Jacobian-vector product (JVP) which does not require any AD. We start with the following (conditional) expectation
\begin{align}\label{eq:app_meng_starting_point}
    \E\left[\alpha_t^2 \rvx_0 \rvx_t^\top - \alpha_t \left[\rvx_0 \rvx_t^\top + \rvx_t \rvx_0^\top  \right]\mid \rvx_t, t \right] = -\rvx_t \rvx_t^\top + \sigma_t^4 \mS_2(\rvx_t, t) + \sigma_t^4 \vs_1(\rvx_t, t) \vs_1(\rvx_t, t)^\top + \sigma_t^2 \mI,
\end{align}
where $\vs_1(\rvx_t, t) \coloneqq \nabla_{\rvx_t} \log p_t(\rvx_t)$ and $\mS_2(\rvx_t, t) \coloneqq \nabla_{\rvx_t}^\top \nabla_{\rvx_t} \log p_t(\rvx_t)$. The above formula is derived in~\citet[Theorem 1,][]{meng2021estimating}. Adding $\rvx_t \rvx_t^\top$ to~\Cref{eq:app_meng_starting_point} and subsequently dividing by $\sigma_t^2$, we have
\begin{align}
    \E\left[\frac{\alpha_t^2}{\sigma_t^2} \rvx_0 \rvx_t^\top - \frac{\alpha_t}{\sigma_t^2} \left[\rvx_0 \rvx_t^\top + \rvx_t \rvx_0^\top\right]  + \frac{1}{\sigma_t^2} \rvx_t \rvx_t^\top \mid \rvx_t, t \right] &= \sigma_t^2 \mS_2(\rvx_t, t) + \sigma_t^2 \vs_1(\rvx_t, t) \vs_1(\rvx_t, t)^\top + \mI,
\end{align}
where we could pull the $\frac{1}{\sigma_t^2} \rvx_t \rvx_t^\top$ term into the expectation because it is conditioned on $t$ and $\rvx_t$.
Using $\rvx_t = \alpha_t \rvx_0 + \sigma_t \vepsilon$, we can rewrite the above as
\begin{align}
    \E\left[\vepsilon\vepsilon^\top \mid \rvx_t, t\right] = \sigma_t^2 \mS_2(\rvx_t, t) + \sigma_t^2 \vs_1(\rvx_t, t) \vs_1(\rvx_t, t)^\top + \mI.
\end{align}
For an arbitrary $\vv \coloneqq \vv(\rvx_t, t)$, we then have
\begin{align}
    \E\left[\vepsilon\vepsilon^\top \vv \mid \rvx_t, t\right] = \sigma_t^2 \mS_2(\rvx_t, t) \vv + \sigma_t^2 \vs_1(\rvx_t, t) \vs_1(\rvx_t, t)^\top \vv + \vv.
\end{align}
Therefore, we can develop a score matching-like
learning objective for the (general) spatial JVP $\vo_\vtheta(\rvx_t, t) \approx \mS_2(\rvx_, t) \vv$ as
\begin{align} \label{eq:app_meng_spatial_jvp_objective}
    \E_{t \sim \gU[t_\mathrm{cutoff}, 1], \rvx_0 \sim p(\rvx_0), \vepsilon \sim \gN(\bm{0}, \mI)} \left[ g_\mathrm{no-ad}(t) \| \vo_\vtheta(\rvx_t, t) + \vs_\vtheta(\rvx_t, t) \vs_\vtheta(\rvx_t, t)^\top \vv + \frac{1}{\sigma_t^2} \vv - \vepsilon \vepsilon^\top \vv \|_2^2\right],
\end{align}
for some weighting function $g_\mathrm{no-ad}(t)$. Setting $\vv(\rvx_t, t) = -\sigma_t \left(\frac{1}{\sqrt{\gamma_t^2 + 1}} \vepsilon_\vtheta(\rvx_t, t) - \frac{\gamma_t}{1 + \gamma_t^2} \rvx_t\right)$, would recover the spatial JVP needed for the computation of $d_{\gamma_t} \vepsilon$. In the initial phase of this project, we briefly experimented with learning the spatial JVP using this approach; however, we found that our distillation approach worked significantly better. 

\newpage
\section{Toy Experiments} \label{s:app_toy_experiments}
For all toy experiments in~\Cref{s:method}, we consider the following ground truth distribution:
\begin{align}
    p_0(\rvx_0) = \frac{1}{8} \sum_{i=1}^8 p_0^{(i)}(\rvx_0),
\end{align}
where 
\begin{align}
    p_0^{(i)}(\rvx_0) = \frac{1}{8} \sum_{j=1}^8 \gN(\rvx_0, s_1\vmu_i + s_1 s_2 \vmu_j, \sigma^2 \mI).
\end{align}
We set $\sigma = 10^{-2}$, $s_1 = 0.9$, $s_2=0.2$,  and 
\begin{align*}
    \vmu_1 &= \begin{pmatrix} 1 \\ 0 \end{pmatrix}, \quad &&\vmu_2 = \begin{pmatrix} -1 \\ 0 \end{pmatrix}, \quad &&\vmu_3 = \begin{pmatrix} 0 \\ 1 \end{pmatrix}, \quad &&\vmu_4 = \begin{pmatrix} 0 \\ -1 \end{pmatrix} \\
    \vmu_5 &= \begin{pmatrix} \tfrac{1}{\sqrt{2}} \\ \tfrac{1}{\sqrt{2}} \end{pmatrix}, \quad &&\vmu_6 = \begin{pmatrix} \tfrac{1}{\sqrt{2}} \\ -\tfrac{1}{\sqrt{2}} \end{pmatrix}, \quad &&\vmu_7 = \begin{pmatrix} -\tfrac{1}{\sqrt{2}} \\ \tfrac{1}{\sqrt{2}} \end{pmatrix}, \quad &&\vmu_8 = \begin{pmatrix} -\tfrac{1}{\sqrt{2}} \\ -\tfrac{1}{\sqrt{2}} \end{pmatrix}.
\end{align*}
The ground truth distribution is visualized in~\Cref{fig:toy_distribution_ground_truth}. Note that we can compute the score functions (and all its derivatives) analytically for Gaussian mixture distributions.

\begin{figure}
    \begin{subfigure}[b]{0.49\textwidth}
        \centering
        \includegraphics[scale=1.20]{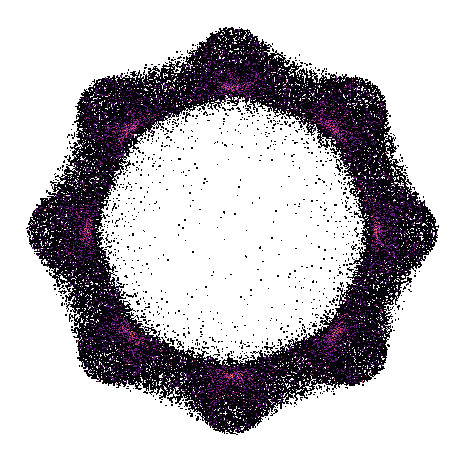}
        \caption{DDIM, $n=5$}
    \end{subfigure}
    \begin{subfigure}[b]{0.49\textwidth}
        \centering
        \includegraphics[scale=1.20]{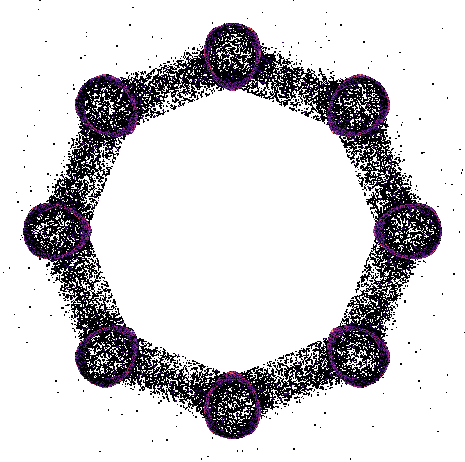}
        \caption{GENIE, $n=5$}
    \end{subfigure}
    \begin{subfigure}[b]{0.49\textwidth}
        \centering
        \includegraphics[scale=1.20]{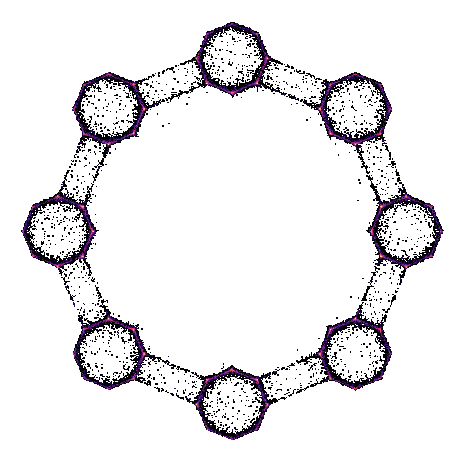}
        \caption{DDIM, $n=10$}
    \end{subfigure}
    \begin{subfigure}[b]{0.49\textwidth}
        \centering
        \includegraphics[scale=1.20]{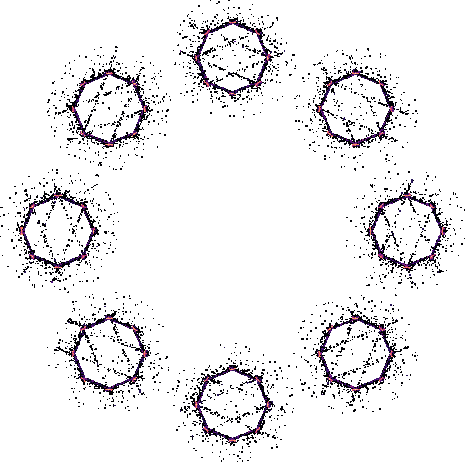}
        \caption{GENIE, $n=10$}
    \end{subfigure}
    \begin{subfigure}[b]{0.49\textwidth}
        \centering
        \includegraphics[scale=1.20]{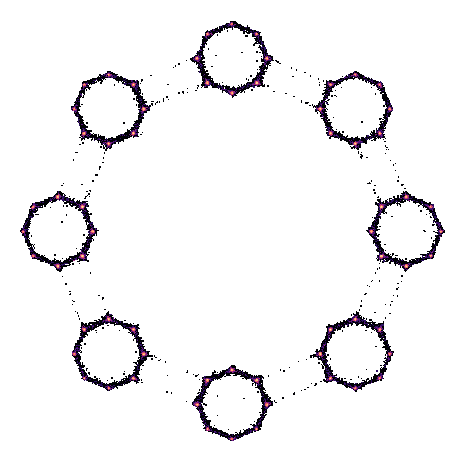}
        \caption{DDIM, $n=15$}
    \end{subfigure}
    \begin{subfigure}[b]{0.49\textwidth}
        \centering
        \includegraphics[scale=1.20]{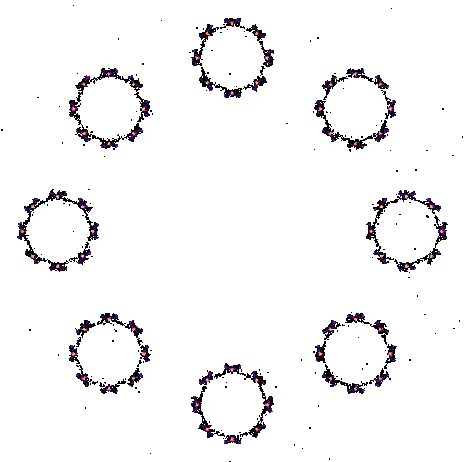}
        \caption{GENIE, $n=15$}
    \end{subfigure}
    \begin{subfigure}[b]{0.49\textwidth}
        \centering
        \includegraphics[scale=1.20]{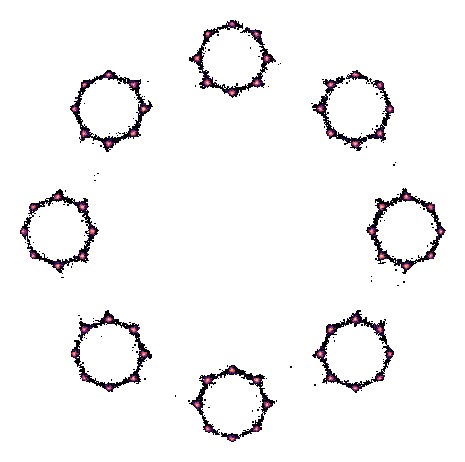}
        \caption{DDIM, $n=20$}
    \end{subfigure}
    \begin{subfigure}[b]{0.49\textwidth}
        \centering
        \includegraphics[scale=1.20]{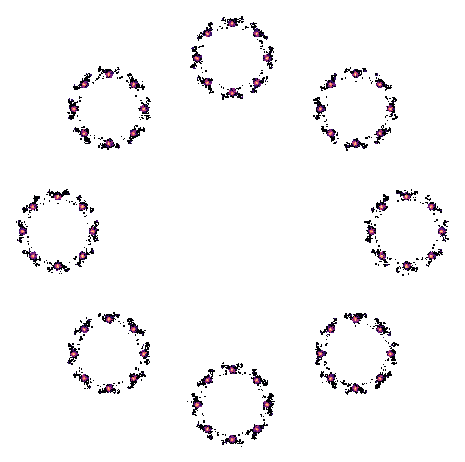}
        \caption{GENIE, $n=20$}
    \end{subfigure}
    \caption{Modeling a complex 2D toy distribution: Samples are generated with DDIM and GENIE with $n$ solver steps using the analytical score function of the ground truth distribution (visualized in~\Cref{fig:toy_distribution_ground_truth}). Zoom in for details.}
    \label{fig:app_analytical_solver}
\end{figure}

In~\Cref{fig:toy_distribution}, we compared DDIM to GENIE for sampling using the analytical score function of the ground truth distribution with 25 solver steps. In~\Cref{fig:app_analytical_solver}, we repeated this experiment for 5, 10, 15, and 20 solver steps. We found that in particular for $n=10$ both solvers generate samples in interesting patterns.
\newpage
\section{Image Experiments}
\label{s:app_image_experiments}

\subsection{Evaluation Metrics, Baselines, and Datasets} \label{s:app_eval_metrics}
\textbf{Metrics:} We quantitatively measure sample quality via Fr\'echet Inception Distance~\citep[FID,][]{heusel2017gans}. It is common practice to use 50k samples from the training set for reference statistics. We follow this practice for all datasets except for ImageNet and Cats. For ImageNet, we follow~\citet{dhariwal2021diffusion} and use the entire training set for reference statistics. For the small Cats dataset, we use the training as well as the validation set for reference statistics.

\textbf{Baselines:} We run baseline experiments using two publicly available repositories. The \href{https://github.com/yang-song/score_sde_pytorch}{\texttt{score\_sde\_pytorch}} repository is licensed according to the Apache License 2.0; see also their license file~\href{https://github.com/yang-song/score_sde_pytorch/blob/main/LICENSE}{here}. The \href{https://github.com/nv-tlabs/CLD-SGM}{\texttt{CLD-SGM}} repository is licensed according to the NVIDIA Source Code License; see also their license file~\href{https://github.com/nv-tlabs/CLD-SGM/blob/main/LICENSE}{here}.

\textbf{Datasets:} We link here the websites of the datasets used in this experiment: \href{https://www.cs.toronto.edu/~kriz/cifar.html}{CIFAR-10}, \href{https://www.yf.io/p/lsun}{LSUN datasets}, \href{https://www.image-net.org/}{ImageNet}, and \href{https://github.com/clovaai/stargan-v2/blob/master/README.md}{AFHQv2}.

\subsection{Analytical First Step (AFS)} \label{s:app_afs}
The forward process of DDMs generally converges to an analytical distribution. This analytical distribution is then used to sample from DDMs, defining the initial condition for the generative ODE/SDE. For example, for variance-preserving DDMs, we have $p_1(\rvx_1) \approx \gN(\rvx_1; \bm{0}, \mI)$.

In this work, we try to minimize the computational complexity of sampling from DDMs, and therefore operate in a low NFE regime. In this regime, every additional function evaluation makes a significant difference. We therefore experimented with replacing the learned score with the (analytical score) of $\gN(\bm{0}, \mI) \approx p_1(\rvx_1)$ in the first step of the ODE solver. This ``gained'' function evaluation can then be used as an additional step in the ODE solver later.

In particular, we have
\begin{align}
    \vepsilon_\vtheta(\rvx_1, 1) \approx \rvx_1,
\end{align}
and $\frac{d\vepsilon_\vtheta(\rvx_1, 1)}{d\gamma_1} \approx \bm{0}$ as shown below:
\begin{align}
    \frac{d\vepsilon_\vtheta(\rvx_1, 1)}{d\gamma_1} &\approx \frac{d\rvx_t}{d\gamma_t} \rvert_{t=1} \\
    &= - \frac{\gamma_t}{\gamma_t^2 + 1} \left[\rvx_t + \vs_\vtheta(\rvx_t, t)\right] \rvert_{t=1} \quad (\text{using~\Cref{eq:app_dx_dgamma2}}) \\
    &\approx \bm{0} \quad (\text{using normal assumption \,} \vs_\vtheta(\rvx_t, t) \approx - \rvx_t)
\end{align}
Given this, the AFS step becomes identical to the Euler update that uses the Normal score function for $\rvx_1$. This step is shown in the pseudocode in~\Cref{alg:app_genie}.

\subsection{Classifier-Free Guidance} \label{s:app_cfg}
As discussed in~\Cref{s:guidance_and_encoding}, to guide diffusion sampling towards particular classes, we replace $\vepsilon_\vtheta(\rvx_t, t)$ with 
\begin{align} \label{eq:app_cfg}
    \hat \vepsilon_\vtheta(\rvx_t, t, c, w) = (1+w) \vepsilon_{\vtheta}(\rvx_t, t, c) - w \vepsilon_\vtheta(\rvx_t, t),
\end{align}
where $w>1.0$ is the ``guidance scale'', in the DDIM ODE. We experiment with classifier-free guidance on ImageNet. In~\Cref{eq:app_cfg} we re-use the conditional ImageNet score model $\vepsilon_{\vtheta}(\rvx_t, t, c)$ trained before (see~\Cref{s:app_training_score_models} for details), and train an additional unconditional ImageNet score model $\vepsilon_{\vtheta}(\rvx_t, t)$ using the exact same setup (and simply setting the class embedding to zero). We also re-use the conditional prediction head trained on top of the conditional ImageNet score model and train an additional prediction head for the unconditional model. Note that for both the score models as well as the prediction heads, we could share parameters between the models to reduce computational complexity~\citep{ho2021classifierfree}. The modified GENIE scheme for classifier-free guidance is then given as 
\begin{align}
    \bar \rvx_{t_{n+1}} = \bar \rvx_{t_n} + (\gamma_{t_{n+1}} - \gamma_{t_n}) \hat \vepsilon_\vtheta(\rvx_{t_n}, t_n, c, w) + \frac{1}{2} (\gamma_{t_{n+1}} - \gamma_{t_n})^2 \hat \vk_\vpsi(\rvx_{t_n}, t_n, c, w),
\end{align}
where 
\begin{align}
    \hat \vk_\vpsi(\rvx_{t_n}, t_n, c, w) = (1+w) \vk_\vpsi(\rvx_{t_n}, t_n, c) - w \vk_\vpsi(\rvx_{t_n}, t_n).
\end{align}
\subsection{Encoding} \label{s:app_encoding}
To encode a data point $\rvx_0$ into latent space, we first ``diffuse'' the data point to $t=10^{-3}$, i.e., $\rvx_t = \alpha_t \rvx_0 + \sigma_t \vepsilon$, $\vepsilon \sim \gN(\bm{0}, \mI)$. We subsequently simulate the generative ODE (backwards) from $t=10^{-3}$ to $t=1$, obtaining the latent point $\rvx_1$.

To decode a latent point $\rvx_1$, we simulate the generative ODE (forwards) from $t=1.0$ to $t=10^{-3}$. We then denoise the data point, i.e., $\rvx_0 = \frac{\rvx_t - \sigma_t \vepsilon_\vtheta(\rvx_t, t)}{\alpha_t}$. Note that denoising is generally optional to sample from DDMs; however, for our encoding-decoding experiment we always used denoising in the decoding part to match the inital ``diffusion'' in the encoding part.
\subsection{Latent Space Interpolation} \label{s:app_lsi}
\begin{figure}
    \centering
    \includegraphics[scale=0.65]{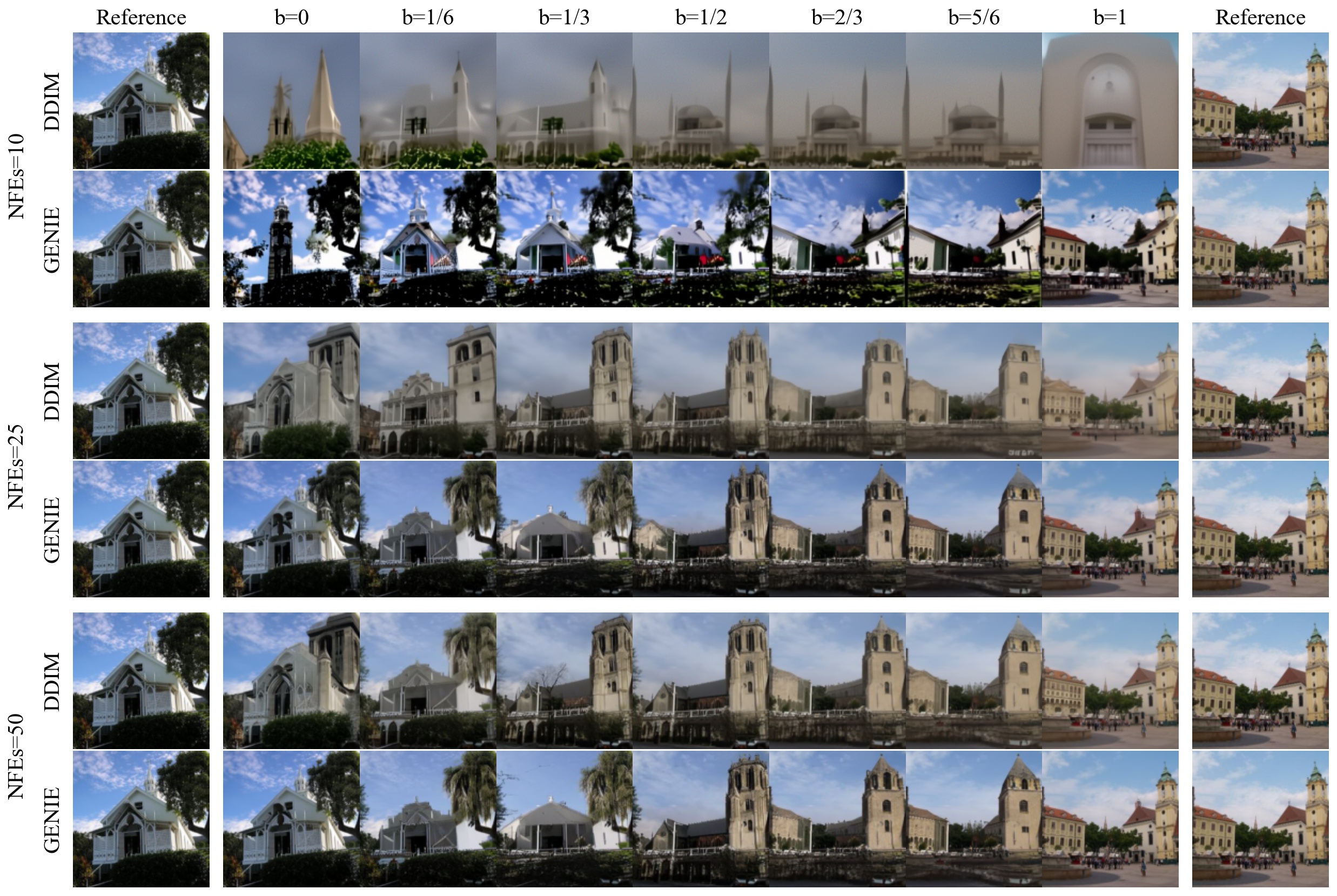}\\
    \vspace{0.5cm}
    \includegraphics[scale=0.65]{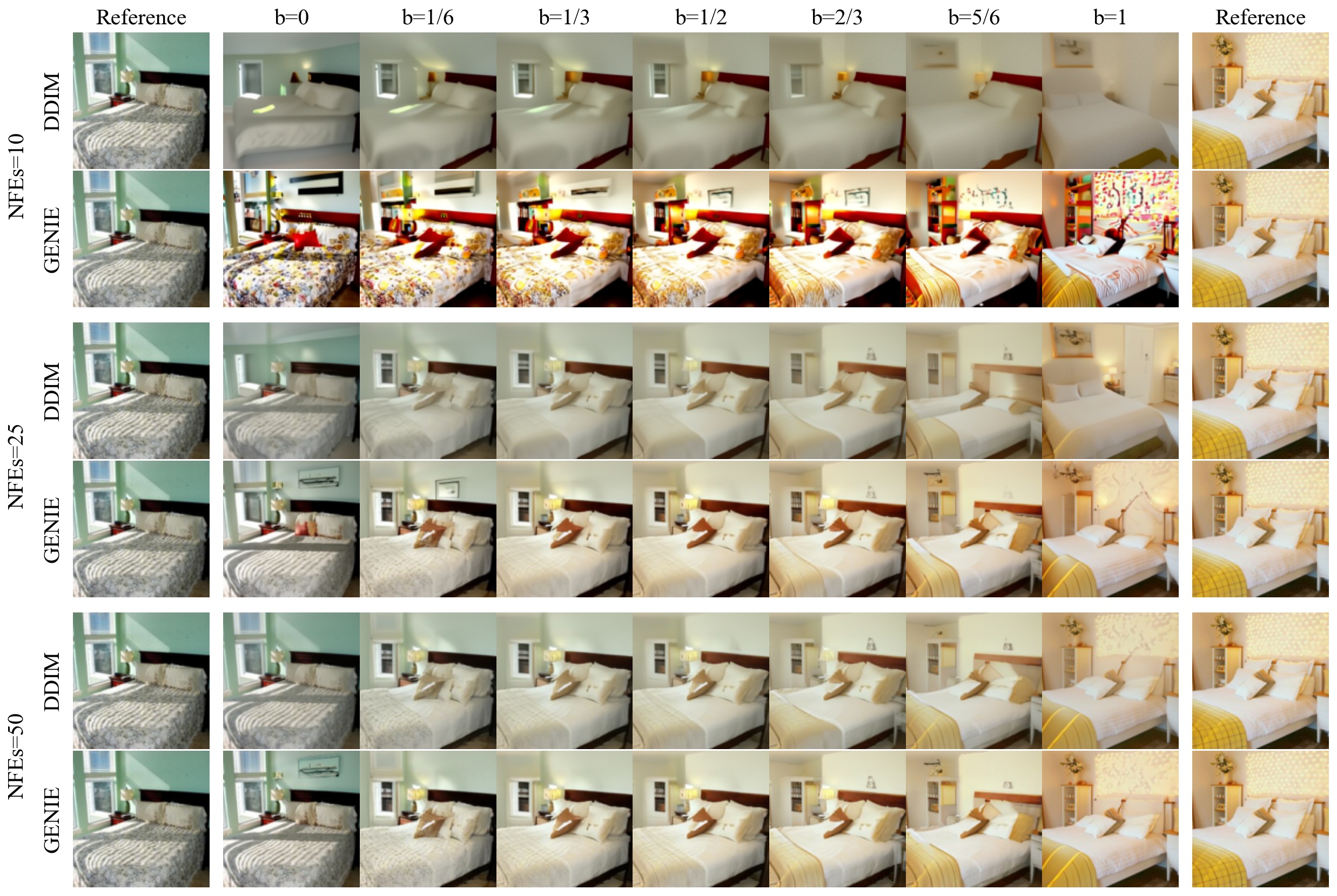}
    \caption{Latent space interpolations for LSUN Church-Outdoor (Top) and LSUN Bedrooms (Bottom). Note that $b=0$ and $b=1$ correspond to the decodings of the encoded reference images. Since this encode-decode loop is itself not perfect, the references are not perfectly reproduced at $b=0$ and $b=1$.}
    \label{fig:lsi}
\end{figure}

We can use encoding to perform latent space interpolation of two data points $\rvx_0^{(0)}$ and $\rvx_0^{(1)}$. We first encode both data points, following the encoding setup from~\Cref{s:app_encoding}, and obtain $\rvx_1^{(0)}$ and $\rvx_1^{(1)}$, respectively. We then perform spherical interpolation of the latent codes:
\begin{align}
    \rvx_1^{(b)} = \rvx_1^{(0)} \sqrt{1 - \mathrm{b}} + \rvx_2^{(1)} \sqrt{\mathrm{b}}, \quad \mathrm{b} \in [0, 1].
\end{align}
Subsequently, we decode the latent code $\rvx_1^{(b)}$ following the decoding setup from~\Cref{s:app_encoding}. In~\Cref{fig:lsi}, we show latent space interpolations for LSUN Church-Outdoor and LSUN Bedrooms.
\subsection{Extended Quantitative Results} \label{s:app_extended_quant}
In this section, we show additional quantitative results not presented in the main paper. In particular, we show results for all four hyperparameter combinations (binary choice of AFS and binary choice of denoising) for methods evaluated by ourselves. For these methods (i.e., GENIE, DDIM, S-PNDM, F-PNDM, Euler--Maruyama), we follow the~\textbf{Synthesis Strategy} outlined in~\Cref{sec:experiments}, with the exception that we use linear striding instead of quadratic striding for S-PNDM~\citep{liu2022pseudo} and F-PNDM~\citep{liu2022pseudo}. To apply quadratic striding to these two methods, one would have to derive the Adams--Bashforth methods for non-constant step sizes which is beyond the scope of our work. 

Results can be found in~\Cref{tab:app_main_cifar_10_extended,tab:app_imagenet,tab:app_lsun_bedrooms,tab:app_lsun_church_outdoor,tab:app_cats_base,tab:app_cats_upsampler}.
As expected, AFS can considerably improve results for almost all methods, in particular for NFEs $\leq 15$. Denoising, on the other hand, is more important for larger NFEs. For our Cats models, we initially found that denoising hurts performance, and therefore did not further test it in all settings.

\textbf{Recall Scores.} We quantify the sample diversity of GENIE and other fast samplers using the recall score~\citep{sajjadi2018assessing}. In particular, we follow DDGAN~\citep{xiao2022tackling} and use the improved recall score~\citep{kynkaanniemi2019improved}; results on CIFAR-10 can be found in~\Cref{tab:app_cifar_10_recall}. As expected, we can see that for all methods recall scores suffer as the NFEs decrease. Compared to the baselines, GENIE achieves excellent recall scores, being on par with F-PNDM for NFE$\geq15$. However, F-PNDM cannot be run for NFE$\leq$10 (due to its additional Runge--Kutta warm-up iterations). Overall, these results confirm that GENIE offers strong sample diversity when compared to other common samplers using the same score model checkpoint.

\begin{table}
    \centering
    \caption{Unconditional CIFAR-10 generative performance, measured in Recall (higher values are better). All methods use the same score model checkpoint.}
    \scalebox{0.85}{\begin{tabular}{l c c c c c c c}
        \toprule
        Method & AFS & Denoising &  NFEs=5 & NFEs=10 & NFEs=15 & NFEs=20 & NFEs=25 \\
        \midrule
        \multirow{4}{*}{GENIE (ours)} & \xmark & \xmark & 0.28 & 0.48 & 0.54 & 0.56 & 0.56 \\\
        & \xmark & \cmark & 0.21 & 0.45 & 0.52 & 0.56 & 0.57 \\
        & \cmark & \xmark & 0.27 & 0.47 & 0.53 & 0.56 & 0.56 \\
        & \cmark & \cmark & 0.19 & 0.46 & 0.53 & 0.55 & 0.56 \\[5pt]
        \multirow{4}{*}{DDIM~\citep{song2021denoising}} & \xmark & \xmark & 0.10 & 0.27 & 0.38 & 0.43 & 0.46 \\
        & \xmark & \cmark & 0.07 & 0.24 & 0.35 & 0.42 & 0.46 \\
        & \cmark & \xmark & 0.08 & 0.27 & 0.38 & 0.43 & 0.46 \\
        & \cmark & \cmark & 0.04 & 0.24 & 0.36 & 0.42 & 0.45 \\[5pt]
        \multirow{4}{*}{S-PNDM~\citep{liu2022pseudo}} & \xmark & \xmark & 0.06 & 0.30 & 0.43 & 0.49 & 0.52 \\
        & \xmark & \cmark & 0.02 & 0.25 & 0.39 & 0.46 & 0.50 \\
        & \cmark & \xmark & 0.11 & 0.33 & 0.45 & 0.50 & 0.53 \\
        & \cmark & \cmark & 0.06 & 0.29 & 0.41 & 0.47 & 0.51 \\[5pt]
        \multirow{4}{*}{F-PNDM~\citep{liu2022pseudo}} & \xmark & \xmark & N/A & N/A & 0.55 & 0.57 & 0.58 \\
        & \xmark & \cmark & N/A & N/A & 0.52 & 0.56 & 0.57 \\
        & \cmark & \xmark & N/A & N/A & 0.55 & 0.58 & 0.59 \\
        & \cmark & \cmark & N/A & N/A & 0.54 & 0.56 & 0.57 \\[5pt]
        \multirow{4}{*}{Euler--Maruyama} & \xmark & \xmark & 0.00 & 0.00 & 0.00 & 0.02 & 0.08 \\
        & \xmark & \cmark & 0.00 & 0.00 & 0.00 & 0.03 &  0.06 \\
        & \cmark & \xmark & 0.00 & 0.00 & 0.00 & 0.03 &  0.09 \\
        & \cmark & \cmark & 0.00 & 0.00 & 0.00 & 0.03 &  0.09 \\
        \bottomrule
    \end{tabular}}
    \label{tab:app_cifar_10_recall}
\end{table}

\textbf{Striding Schedule Grid Search.} As discussed in~\Cref{sec:experiments} the fixed quadratic striding schedule (for choosing the times $t$ for evaluating the model during synthesis under fixed NFE budgets) used in GENIE may be sub-optimal, in particular for small NFEs. To explore this, we did a small grid search over three different striding schedules. As described in \Cref{s:app_ph_pseudocode}, the quadratic striding schedule can be written as $t_n = \left(1.0 - (1.0-\sqrt{t_\mathrm{cutoff}}) \frac{n}{N}\right)^2$, and easily be generalized to
\begin{align}
    t_n = \left(1.0 - (1.0-t_\mathrm{cutoff}^{1/\rho}) \frac{n}{N}\right)^\rho, \rho > 1.
\end{align}
In particular, besides the quadratic schedule $\rho=2$, we also tested the two additional values $\rho=1.5$ and $\rho=2.5$. We tested these schedules on GENIE as well as DDIM~\citep{song2021denoising}; note that the other two comptetive baselines, S-PNDM~\citep{liu2022pseudo} and F-PNDM~\citep{liu2022pseudo}, rely on linear striding, and therefore a grid search is not applicable. We show results for GENIE and DDIM in~\Cref{tab:striding_table_genie}; for each combination of solver and NFE we applied the best synthesis strategy (whether or not we use denoising and/or the analytical first step) of quadratic striding ($\rho=2.0$) also to $\rho =1.5$ and $\rho =2.5$.
As can be seen in the table, $\rho=1.5$ improves for both DDIM and GENIE for NFE$=$5 (over the quadratic schedule $\rho=2$), whereas larger $\rho$ are preferred for larger NFE. The improvement of GENIE from 13.9 to 11.2 FID for NFE=5 is significant.

\begin{table}
    \centering
    \caption{Unconditional CIFAR-10 generative performance (measured in FID) using our GENIE and DDIM~\citep{song2021denoising} with different striding schedules using exponents $\rho \in \{1.5, 2.0, 2.5\}$.}
    \begin{tabular}{l c c c c c c}
        \toprule
        Method &$\rho$ & NFEs=5 & NFEs=10 & NFEs=15 & NFEs=20 & NFEs=25 \\
        \midrule 
        &1.5 & \textbf{11.2} & \textbf{5.28} & 5.03 & 4.35 & 3.97 \\
        GENIE &2.0 & 13.9 & 5.97 & \textbf{4.49} & \textbf{3.94} & 3.67 \\
        &2.5 & 17.8 & 7.19 & 4.57 & \textbf{3.94} & \textbf{3.64} \\
        \midrule \midrule
        &1.5 & \textbf{27.6} & 13.5 & 8.97 & 7.20 & 6.15 \\
        DDIM &2.0 & 29.7 & \textbf{11.2} & \textbf{7.35} & \textbf{5.87} & \textbf{5.16} \\
        &2.5 & 33.2 & 13.4 & 8.28 & 6.36 & 5.39 \\
        \bottomrule
    \end{tabular}
    \label{tab:striding_table_genie}
\end{table}

\textbf{Discretization Errors of GENIE compared to other Fast Samplers.} We compute discretization errors, in particular local and global truncation errors, of GENIE and compare to existing faster solvers. We are using the CIFAR-10 model. We initially sample 100 latent vectors $\mathbf{x}_T \sim \mathcal{N}(\bm{0}, \bm{I})$ and then, starting from those latent vectors, synthesize 100 approximate ground truth trajectories (GTTs) using DDIM with 1k NFEs (for that many steps, the discretization error is negligible; hence, we can treat this as a pseudo ground truth).

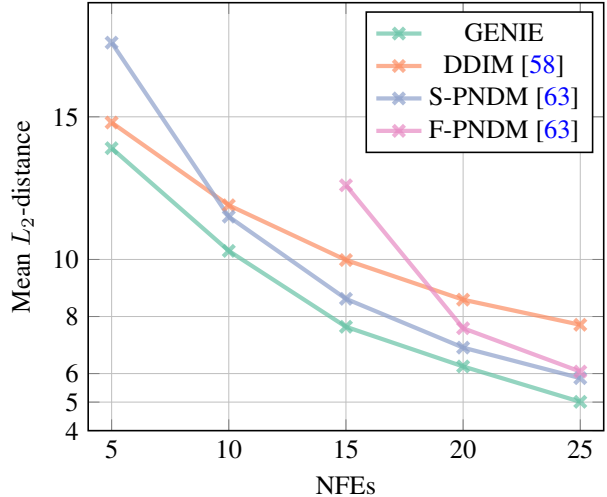
\begin{figure}
    \centering
    \begin{tikzpicture}[scale=1.0]
\begin{axis}[xtick={100, 80, 60, 40, 25, 20, 15, 10, 5}, xticklabels={100, 80, 60, 40, 25, 20, 15, 10, 5}, ytick={2, 4, 5, 6, 8, 10, 15, 20}, yticklabels={2, 4, 5, 6, 8, 10, 15, 20}, ymin=4, ymax=19, xmin=4, xmax=26, xlabel=NFEs, ylabel=Mean $L_2$-distance, grid=both, grid style={line width=.1pt, draw=gray!10}, major grid style={line width=.2pt,draw=gray!50}, every axis plot/.append style={ultra thick}, every axis plot/.append style={mark size=3pt}, xlabel near ticks, ylabel near ticks]
\addplot[color=set21,mark=x, opacity=0.7] coordinates {
(25, 5.01)
(20, 6.25)
(15, 7.64)
(10, 10.3)
(5, 13.9)
};
\addlegendentry{GENIE}
\addplot[color=set22,mark=x, opacity=0.7] coordinates {
(25, 7.71)
(20, 8.59)
(15, 9.98)
(10, 11.9)
(5, 14.8)
};
\addlegendentry{DDIM~\citep{song2021denoising}}
\addplot[color=set23,mark=x, opacity=0.7] coordinates {
(25, 5.84)
(20, 6.91)
(15, 8.62)
(10, 11.5)
(5, 17.6)
};
\addlegendentry{S-PNDM~\citep{liu2022pseudo}}
\addplot[color=set24,mark=x, opacity=0.7] coordinates {
(25, 6.07)
(20, 7.59)
(15, 12.6)
};
\addlegendentry{F-PNDM~\citep{liu2022pseudo}}
\addlegendentry{DDIM}
\end{axis}
\end{tikzpicture}
    \caption{\textbf{Global Truncation Error:} $L_2$-distance of generated outputs by the fast samplers to the (approximate) ground truth (computed using DDIM with 1k NFEs) in Inception feature space~\citep{szegedy2016rethinking}. Results are averaged over 100 samples.}
    \label{fig:global_truncation_error}
\end{figure}

We then synthesize 100 sample trajectories for DDIM~\citep{song2021denoising}, S-PNDM~\citep{liu2022pseudo}, F-PNDM~\citep{liu2022pseudo}, and GENIE (for NFEs=$\{5, 10, 15, 20, 25\}$, similar to the main experiments) using the same latent vectors as starting points that were used to generate the GTTs. DDIM, S-PNDM, and F-PNDM are training-free methods that can be run on the exact same score model, which also our GENIE relies on. Thereby, we are able to isolate discretization errors from errors in the learnt score function. We then compute the average $L_2$-distance (in Inception feature space~\citep{szegedy2016rethinking}) between the output image of the fast samplers and the ``output'' of the pseudo GTT. As can be seen in~\Cref{fig:global_truncation_error}, GENIE outperforms the three other methods on all NFEs.
    
Comparing the local truncation error (LTE) of different higher-order solvers can unfortunately not be done in a fair manner. Similar to DDIM, GENIE only needs the current value and a single NFE to predict the next step. In contrast, multistep methods rely on a history of predictions and Runge--Kutta methods rely on multiple NFEs to predict the next step. Thus, we can only fairly compare the LTE of GENIE to the LTE of DDIM. In particular, we compute LTEs at three starting times $t \in \{0.1, 0.2, .5\}$ (similar to what we did in~\Cref{fig:analytical_error}). For each $t$, we then compare one step predictions for different step sizes $\Delta t$ against the ground truth trajectory ($L_2$-distance in data space averaged over 100 predictions; since we are not operating directly in image space at these intermediate $t$, using inception feature would not make sense here). As expected, we can see in~\Cref{fig:local_truncation_error} that GENIE has smaller LTE than DDIM for all starting times $t$.
\begin{figure}
    \centering
    \pgfplotsset{scaled x ticks=false}
\begin{tikzpicture}[scale=1.0]
\begin{axis}[xtick={0.05, 0.045, 0.04, 0.035, 0.03, 0.025, 0.02, 0.015, 0.01, 0.005}, xticklabels={0.05, 0.045, 0.04, 0.035, 0.03, 0.025, 0.02, 0.015, 0.01, 0.005}, ytick={0, 0.05, 0.1, 0.15, 0.2, 0.25}, yticklabels={0, 0.05, 0.1, 0.15, 0.2, 0.25}, ymin=-0.01, ymax=0.35, xmin=0.004, xmax=0.051, xlabel=Step size $\Delta t$, ylabel=$\|\rvx_t - \hat \rvx_t(\Delta_t) \|$, grid=both, grid style={line width=.1pt, draw=gray!10}, major grid style={line width=.2pt,draw=gray!50}, every axis plot/.append style={ultra thick}, every axis plot/.append style={mark size=3pt}, xlabel near ticks, ylabel near ticks, legend pos=north west, height=.3\paperheight, width=.8\linewidth, scale only axis]
\addplot[color=set21, opacity=0.7] coordinates {
(0.05, 0.1699)
(0.045, 0.1323)
(0.04, 0.1008)
(0.035, 0.0746)
(0.03, 0.0532)
(0.025, 0.0359)
(0.02, 0.0223)
(0.015, 0.0122)
(0.01, 0.0052)
(0.005, 0.0013)
};
\addlegendentry{GENIE, $t=0.1$}
\addplot[color=set21,dash pattern={on 7pt off 2pt on 1pt off 3pt}, opacity=0.7] coordinates {
(0.05, 0.0934)
(0.045, 0.0753)
(0.04, 0.0592)
(0.035, 0.0452)
(0.03, 0.0331)
(0.025, 0.0229)
(0.02, 0.0146)
(0.015, 0.0082)
(0.01, 0.0036)
(0.005, 0.0009)
};
\addlegendentry{GENIE, $t=0.2$}
\addplot[color=set21,dotted, opacity=0.7] coordinates {
(0.05, 0.1391)
(0.045, 0.1142)
(0.04, 0.0916)
(0.035, 0.0711)
(0.03, 0.0528)
(0.025, 0.0371)
(0.02, 0.0241)
(0.015, 0.0137)
(0.01, 0.0063)
(0.005, 0.0016)
};
\addlegendentry{GENIE, $t=0.5$}
\addplot[color=set22, opacity=0.7] coordinates {
(0.05, 0.3239)
(0.045, 0.2560)
(0.04, 0.1977)
(0.035, 0.1479)
(0.03, 0.1062)
(0.025, 0.0720)
(0.02, 0.0449)
(0.015, 0.0244)
(0.01, 0.0104)
(0.005, 0.0024)
};
\addlegendentry{DDIM, $t=0.1$}
\addplot[color=set22,dash pattern={on 7pt off 2pt on 1pt off 3pt}, opacity=0.7] coordinates {
(0.05, 0.1568)
(0.045, 0.1266)
(0.04, 0.0996)
(0.035, 0.0759)
(0.03, 0.0555)
(0.025, 0.0381)
(0.02, 0.0241)
(0.015, 0.0132)
(0.01, 0.0056)
(0.005, 0.0012)
};
\addlegendentry{DDIM, $t=0.2$}
\addplot[color=set22,dotted, opacity=0.7] coordinates {
(0.05, 0.2359)
(0.045, 0.1938)
(0.04, 0.1551)
(0.035, 0.1202)
(0.03, 0.0889)
(0.025, 0.0622)
(0.02, 0.0398)
(0.015, 0.0224)
(0.01, 0.0095)
(0.005, 0.0022)
};
\addlegendentry{DDIM, $t=0.5$}
\end{axis}
\end{tikzpicture}
    \caption{\textbf{Local Truncation Error:} Single step (local discretization) error, measured in $L_2$-distance to (approximate) ground truth (computed using DDIM with 1k NFEs) in data space and averaged over 100 samples, for GENIE and DDIM for three starting time points $t \in \{0.1, 0.2, 0.5\}$ (this is, the $t$ from which a small step with size $\Delta t$ is taken).}
    \label{fig:local_truncation_error}
\end{figure}
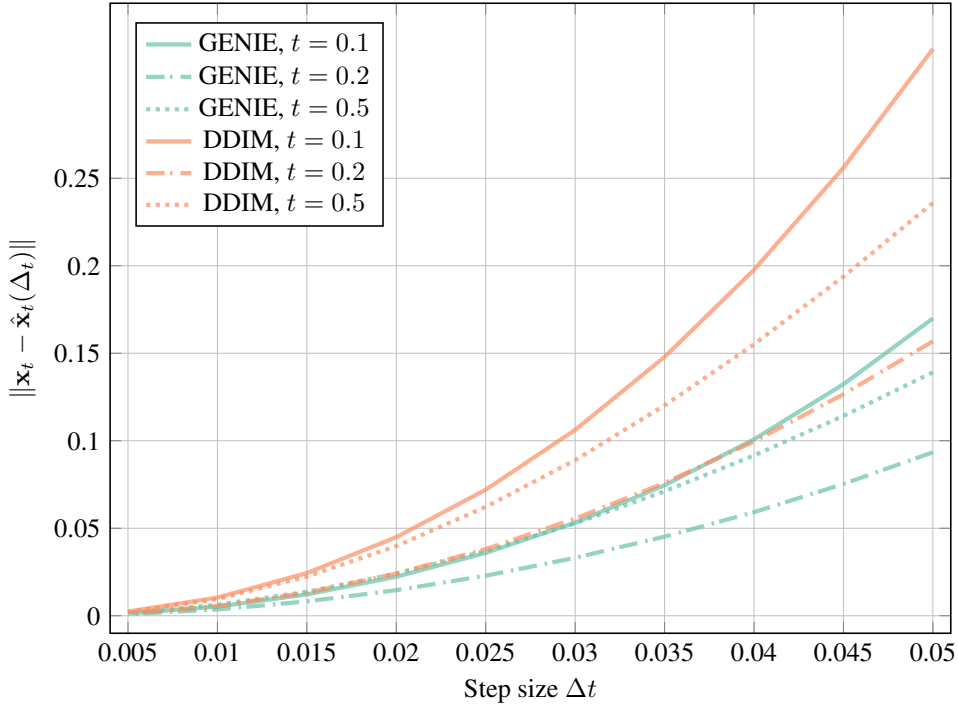
\begin{table}
    \centering
    \caption{Unconditional CIFAR-10 generative performance (measured in FID). Methods above the middle line use the same score model checkpoint; methods below all use different ones. (\textdagger): numbers are taken from literature. This table is an extension of~\Cref{tab:main_cifar_10}.}
    \scalebox{0.85}{\begin{tabular}{l c c c c c c c}
        \toprule
        Method & AFS & Denoising &  NFEs=5 & NFEs=10 & NFEs=15 & NFEs=20 & NFEs=25 \\ %
        \midrule
        \multirow{4}{*}{GENIE (ours)} & \xmark & \xmark & 15.4 & \textbf{5.97} & 4.70 & 4.30 & 4.10 \\
        & \xmark & \cmark & 23.5 & 6.91 & 4.74 & 4.02 & 3.72 \\
        & \cmark & \xmark & 13.9 & 6.04 & 4.76 & 4.33 & 4.18 \\
        & \cmark & \cmark & 17.9 & 6.27 & \textbf{4.49} & \textbf{3.94} & \textbf{3.67} \\[5pt]%
        \multirow{4}{*}{DDIM~\citep{song2021denoising}} & \xmark & \xmark & 30.1 & 11.6 & 7.56 & 6.00 & 5.27 \\
        & \xmark & \cmark & 37.9 & 13.9 & 8.76 & 6.77 & 5.76 \\
        & \cmark & \xmark & 29.7 & 11.2 & 7.35 & 5.87 & 5.16 \\
        & \cmark & \cmark & 35.2 & 12.8 & 8.17 & 6.39 & 5.49 \\[5pt]%
        \multirow{4}{*}{S-PNDM~\citep{liu2022pseudo}} & \xmark & \xmark & 60.2 & 12.1 & 7.16 & 5.48 & 4.62 \\
        & \xmark & \cmark & 101 & 17.2 & 10.8 & 8.74 & 7.62 \\
        & \cmark & \xmark & 35.9 & 10.3 & 6.61 & 5.20 & 4.51 \\
        & \cmark & \cmark & 56.8 & 14.9 & 10.2 & 8.37 & 7.35 \\[5pt]%
        \multirow{4}{*}{F-PNDM~\citep{liu2022pseudo}} & \xmark & \xmark & N/A & N/A & 12.1 & 6.58 & 4.89 \\
        & \xmark & \cmark & N/A & N/A & 19.5 & 10.6 & 8.43 \\
        & \cmark & \xmark & N/A & N/A & 10.3 & 5.96 & 4.73\\
        & \cmark & \cmark & N/A & N/A & 15.7 & 10.9 & 8.52 \\[5pt]%
        \multirow{4}{*}{Euler--Maruyama} & \xmark & \xmark & 364 & 236 & 178 & 121 & 85.0\\
        & \xmark & \cmark & 391 & 235 & 191 & 129 & 89.9 \\
        & \cmark & \xmark & 325 & 230 & 164 & 112 & 80.3 \\
        & \cmark & \cmark & 364 & 235 & 176 & 120 & 83.6 \\
        \midrule
        FastDDIM~\citep{kong2021arxiv} (\textdagger) & \xmark & \cmark & - & 9.90 & - & 5.05 & - \\
        Learned Sampler~\citep{watson2022learning} (\textdagger) & \xmark & \cmark & \textbf{12.4} & 7.86 & 5.90 & 4.72 & 4.25 \\
        Analytic DDIM (LS)~\citep{bao2022analyticdpm} (\textdagger) & \xmark & \cmark &  - & 14.0 & - & - & 5.71\\
        CLD-SGM~\citep{dockhorn2022scorebased} & \xmark & \xmark & 334 & 306 & 236 & 162 & 106 \\
        VESDE-PC~\citep{song2020} & \xmark & \cmark & 461 & 461 & 461 & 461 & 462 \\
        \bottomrule
    \end{tabular}}
    \label{tab:app_main_cifar_10_extended}
\end{table}

\begin{table}
    \centering
    \caption{Conditional ImageNet generative performance (measured in FID).}
    \scalebox{0.9}{\begin{tabular}{l c c c c c c c}
        \toprule
        Method & AFS & Denoising &  NFEs=5 & NFEs=10 & NFEs=15 & NFEs=20 & NFEs=25 \\
        \midrule
        \multirow{4}{*}{GENIE (ours)} & \xmark & \xmark &  23.4 & 8.35 & 6.13 & 5.36 & 5.00\\
        & \xmark & \cmark & 35.4 & 7.59 & \textbf{5.23} & \textbf{4.48} & \textbf{4.13} \\
        & \cmark & \xmark & 21.6 & 8.92 & 6.59 & 5.73 & 5.27 \\
        & \cmark & \cmark & \textbf{20.2} & \textbf{7.41} & 5.36 & 4.68 & 4.27 \\[5pt]
        \multirow{4}{*}{DDIM~\citep{song2021denoising}} & \xmark & \xmark & 39.0 & 14.5 & 9.47 & 7.57 & 6.64 \\
        & \xmark & \cmark & 39.8 & 11.1 & 7.17 & 5.83 & 5.19 \\
        & \cmark & \xmark & 37.4 & 14.7 & 9.73 & 7.86 & 6.92  \\
        & \cmark & \cmark & 30.0 & 10.7 & 7.14 & 5.93 & 5.35 \\[5pt]
        \multirow{4}{*}{S-PNDM~\citep{liu2022pseudo}} & \xmark & \xmark & 57.9 & 15.2 & 10.0 & 8.12 & 7.20 \\
        & \xmark & \cmark & 60.6 & 12.2 & 8.69 & 7.59 & 6.94 \\
        & \cmark & \xmark & 39.0 & 13.7 & 9.75 & 8.08 & 7.22 \\
        & \cmark & \cmark & 35.5 & 11.2 & 8.54 & 7.52 & 6.94\\[5pt]
        \multirow{4}{*}{F-PNDM~\citep{liu2022pseudo}} & \xmark & \xmark & N/A & N/A & 13.9 & 9.45 & 7.87 \\
        & \xmark & \cmark & N/A & N/A & 14.5 & 9.45 & 8.05 \\
        & \cmark & \xmark & N/A & N/A & 12.5 & 9.01 & 7.74 \\
        & \cmark & \cmark & N/A & N/A & 12.3 & 9.26 & 7.86 \\
        \bottomrule
    \end{tabular}}
    \label{tab:app_imagenet}
\end{table}

\begin{table}
    \centering
    \caption{Unconditional LSUN Bedrooms generative performance (measured in FID). Methods above the middle line use the same score model checkpoint; Learned Sampler uses a different one. (\textdagger): numbers are taken from literature.}
    \scalebox{0.9}{\begin{tabular}{l c c c c c c c}
        \toprule
        Method & AFS & Denoising &  NFEs=5 & NFEs=10 & NFEs=15 & NFEs=20 & NFEs=25 \\
        \midrule
        \multirow{4}{*}{GENIE (ours)} & \xmark & \xmark & 74.1 & 17.1 & 13.3 & 11.6 & 11.1 \\
        & \xmark & \cmark & 115 & 11.4 & 7.18 & 5.80 & \textbf{5.35} \\
        & \cmark & \xmark & 55.9 & 18.4 & 14.1 & 12.3 & 11.6\\
        & \cmark & \cmark & 47.3 & \textbf{9.29} & \textbf{6.83} & 5.79 & 5.40 \\[5pt]
        \multirow{4}{*}{DDIM~\citep{song2021denoising}} & \xmark & \xmark & 69.6 & 27.1 & 19.0 & 15.8 & 14.2 \\
        & \xmark & \cmark & 81.0 & 16.3 & 9.18 & 7.12 & 6.20 \\
        & \cmark & \xmark & 62.1 & 27.1 & 19.3 & 16.3 & 14.6 \\
        & \cmark & \cmark & 42.5 & 12.5 & 8.21 & 6.77 & 6.05 \\[5pt]
        \multirow{4}{*}{S-PNDM~\citep{liu2022pseudo}} & \xmark & \xmark & 70.4 & 22.1 & 15.7 & 13.5 & 12.4 \\
        & \xmark & \cmark & 88.9 & 12.2 & 8.40 & 7.33 & 6.80 \\
        & \cmark & \xmark & 48.0 & 20.2 & 15.2 & 13.4 & 12.4 \\
        & \cmark & \cmark & 45.0 & 10.8 & 8.14 & 7.23 & 6.71 \\[5pt]
        \multirow{4}{*}{F-PNDM~\citep{liu2022pseudo}} & \xmark & \xmark & N/A & N/A & 36.1 & 18.5 & 14.6 \\
        & \xmark & \cmark & N/A & N/A & 26.8 & 9.85 & 7.86 \\
        & \cmark & \xmark & N/A & N/A & 29.4 & 17.5 & 14.3 \\
        & \cmark & \cmark & N/A & N/A & 18.9 & 9.27 & 7.69 \\
        \midrule
        Learned Sampler~\citep{watson2022learning} (\textdagger)& \xmark & \cmark & \textbf{29.2} & 11.0 & - & \textbf{4.82} & - \\
        \bottomrule
    \end{tabular}}
    \label{tab:app_lsun_bedrooms}
\end{table}

\begin{table}
    \centering
    \caption{Unconditional LSUN Church-Outdoor generative performance (measured in FID). Methods above the middle line use the same score model checkpoint; Learned Sampler uses a different one. (\textdagger): numbers are taken from literature.}
    \scalebox{0.9}{\begin{tabular}{l c c c c c c c}
        \toprule
        Method & AFS &  Denoising &  NFEs=5 & NFEs=10 & NFEs=15 & NFEs=20 & NFEs=25 \\
        \midrule
        \multirow{4}{*}{GENIE (ours)} & \xmark & \xmark & 97.2 & 25.4 & 15.9 & 11.6 & 9.57 \\
        & \xmark& \cmark & 147 & 13.7 & 11.7 & 8.52 & 7.28 \\
        & \cmark & \xmark & 47.8 & 13.6 & 10.6 & 9.17 & 8.28 \\
        & \cmark & \cmark & 60.3 & \textbf{10.5} & \textbf{7.44} & \textbf{6.38} & \textbf{5.84} \\[5pt]
        \multirow{4}{*}{DDIM~\citep{song2021denoising}} & \xmark & \xmark & 81.5 & 28.5 & 16.7 & 11.9 & 9.9\\
        & \xmark & \cmark & 110 & 25.3 & 11.5 & 8.53 & 7.35 \\
        & \cmark & \xmark & 44.0 & 17.4 & 12.5 & 10.2 & 9.07 \\
        & \cmark & \cmark & 45.8 & 12.8 & 8.44 & 6.97 & 6.28 \\[5pt]
        \multirow{4}{*}{S-PNDM~\citep{liu2022pseudo}} & \xmark & \xmark & 59.4 & 18.7 & 13.3 & 11.4 & 10.4 \\
        & \xmark & \cmark & 87.5 & 14.8 & 9.54 & 7.98 & 7.21 \\
        & \cmark & \xmark & 40.7 & 17.0 & 12.8 & 11.2 & 10.3 \\
        & \cmark & \cmark & 48.8 & 12.9 & 9.10 & 7.82 & 7.12 \\[5pt]
        \multirow{4}{*}{F-PNDM~\citep{liu2022pseudo}} & \xmark & \xmark & N/A & N/A & 15.5 & 12.0 & 10.6 \\
        & \xmark & \cmark & N/A & N/A & 15.7 & 9.78 & 7.99 \\
        & \cmark & \xmark & N/A & N/A & 15.2 & 11.8 & 10.4 \\ 
        & \cmark & \cmark & N/A & N/A & 12.6 & 9.29 & 7.83 \\
        \midrule
        Learned Sampler~\citep{watson2022learning} (\textdagger) & \xmark & \cmark & \textbf{30.2} & 11.6 & - & 6.74 & - \\
        \bottomrule
    \end{tabular}}
    \label{tab:app_lsun_church_outdoor}
\end{table}
\begin{table}
    \centering
    \caption{Cats (base model) generative performance (measured in FID).}
    \scalebox{1.0}{\begin{tabular}{l c c c c c}
        \toprule
        Method & AFS & NFEs=10 & NFEs=15 & NFEs=20 & NFEs=25 \\
        \midrule
        \multirow{2}{*}{GENIE (ours)} & \xmark & \textbf{12.2} & \textbf{8.74} & \textbf{7.40} & 6.84 \\
        & \cmark & 13.3 & 9.07 & 7.76 & \textbf{6.76} \\[5pt]
        \multirow{2}{*}{DDIM~\citep{song2021denoising}} & \xmark & 12.7 & 9.89 &  8.66 & 7.98 \\
        & \cmark & 13.6 & 10.0 & 8.73 & 7.87 \\[5pt]
        \multirow{2}{*}{S-PNDM~\citep{liu2022pseudo}} & \xmark & 12.8 & 11.6 & 10.8 & 10.4 \\
        & \cmark &  12.5  & 11.3 & 10.7 & 10.2 \\[5pt]
        \multirow{2}{*}{F-PNDM~\citep{liu2022pseudo}} & \xmark & N/A & 12.8 & 10.4 & 10.6 \\
        & \cmark & N/A &  11.8 & 10.4 & 10.3 \\
        \bottomrule
    \end{tabular}}
    \label{tab:app_cats_base}
\end{table}
\begin{table}
    \centering
    \caption{Cats (upsampler) generative performance (measured in FID).}
    \scalebox{1.0}{\begin{tabular}{l c c c c}
        \toprule
        Method & AFS & NFEs=5 & NFEs=10 & NFEs=15 \\
        \midrule
        \multirow{2}{*}{GENIE (ours)} & \xmark & 7.03 & 4.93 & \textbf{4.83}  \\
        & \cmark & \textbf{5.53} & \textbf{4.90} & 4.91 \\[5pt]
        \multirow{2}{*}{DDIM~\citep{song2021denoising}} & \xmark & 11.3 & 7.16 & 5.99 \\
        & \cmark & 9.47 & 6.64 & 5.85  \\[5pt]
        \multirow{2}{*}{S-PNDM~\citep{liu2022pseudo}} & \xmark & 16.7 & 12.1 & 8.83 \\
        & \cmark & 14.6 & 11.0 & 9.01 \\[5pt]
        \multirow{2}{*}{F-PNDM~\citep{liu2022pseudo}} & \xmark & N/A & N/A &  12.9 \\
        & \cmark & N/A & N/A & 11.7 \\
        \bottomrule
    \end{tabular}}
    \label{tab:app_cats_upsampler}
\end{table}
\subsection{Extended Qualitative Results} \label{s:app_extended_qual}
\begin{figure}
    \begin{subfigure}[b]{\textwidth}
        \centering
        \includegraphics[scale=0.95]{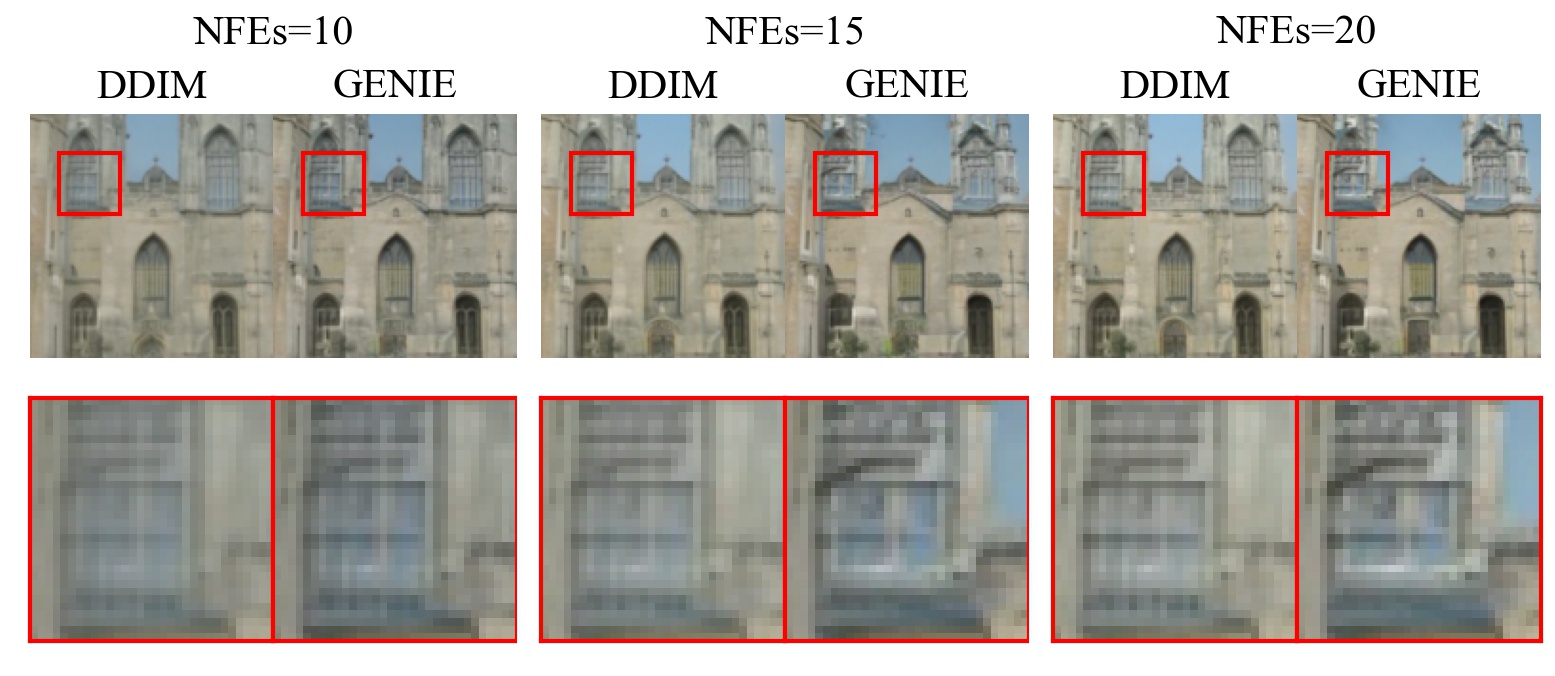}
    \end{subfigure}
    \begin{subfigure}[b]{\textwidth}
        \centering
        \includegraphics[scale=0.95]{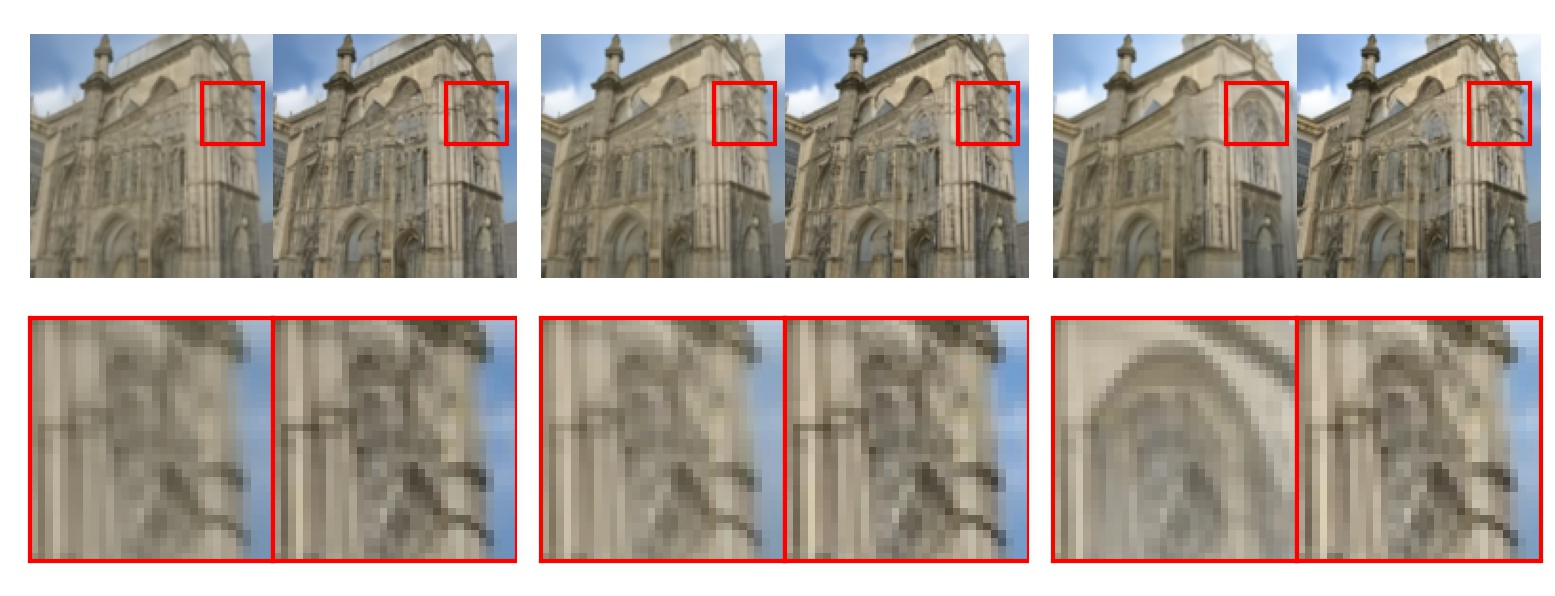}
    \end{subfigure}
    \begin{subfigure}[b]{\textwidth}
        \centering
        \includegraphics[scale=0.95]{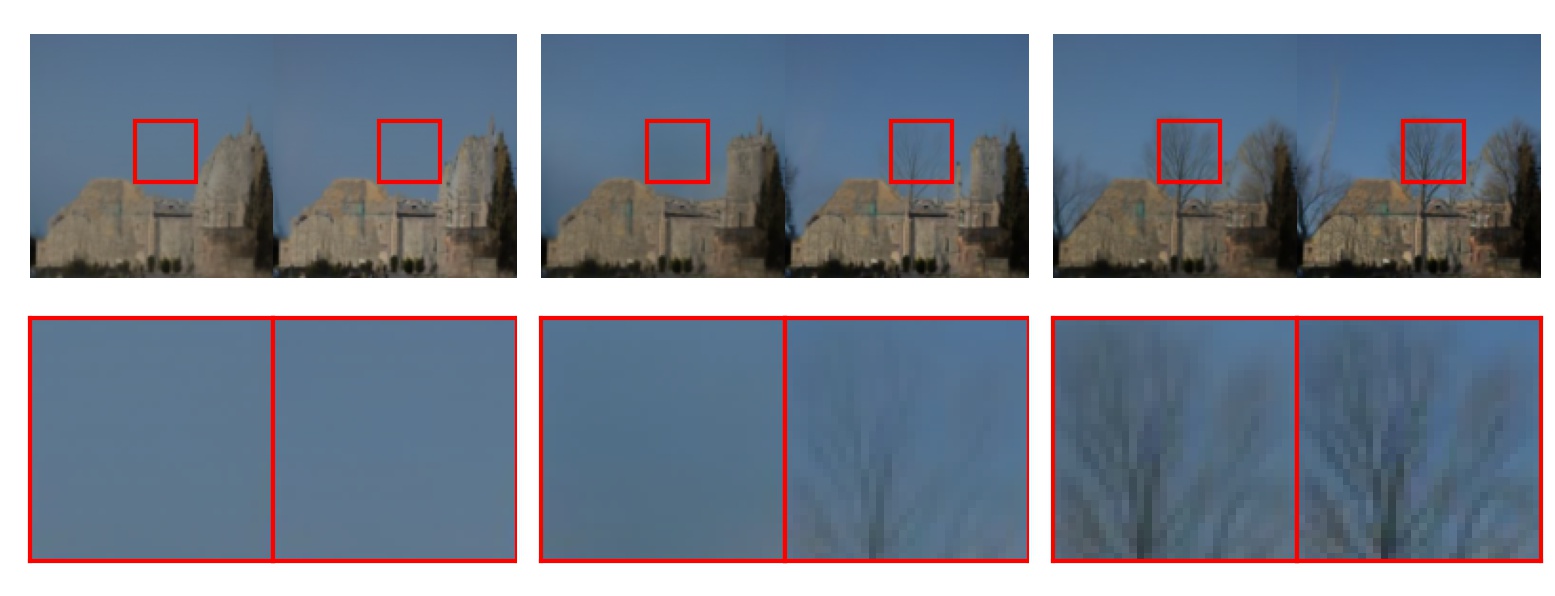}
    \end{subfigure}
    \begin{subfigure}[b]{\textwidth}
        \centering
        \includegraphics[scale=0.95]{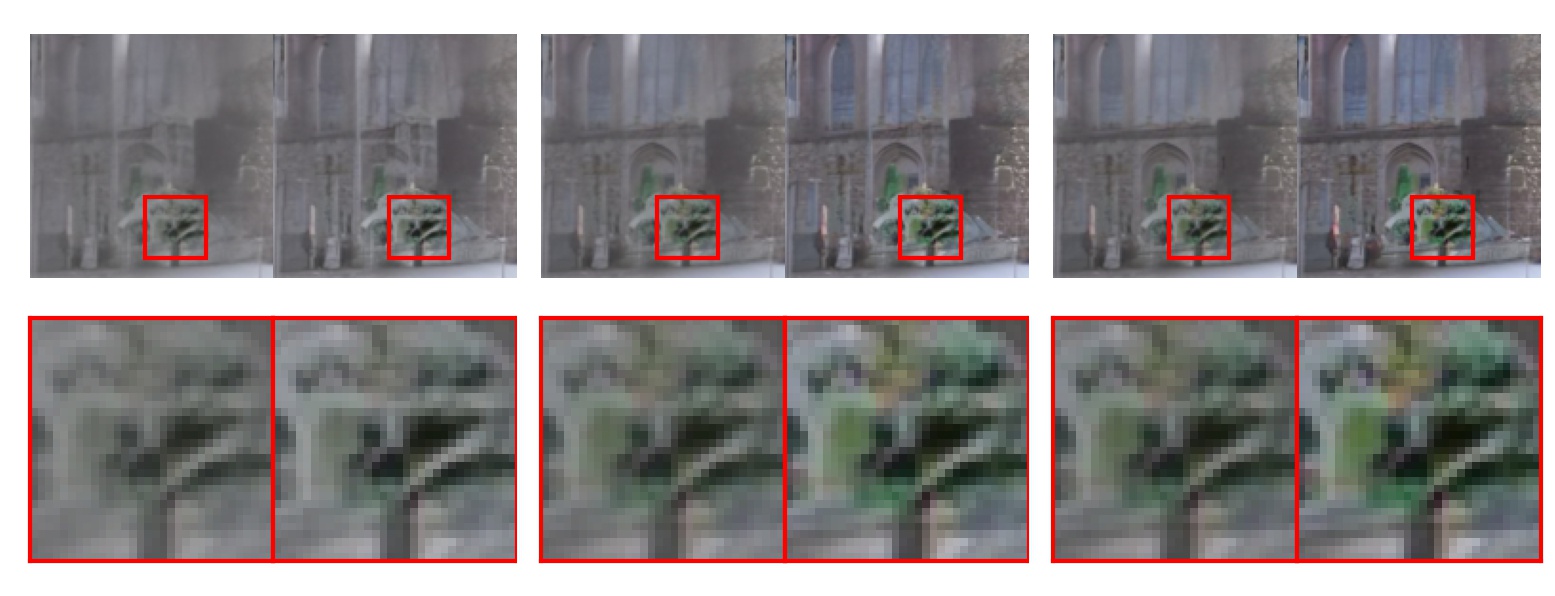}
    \end{subfigure}
    \caption{Additional samples on LSUN Church-Outdoor with zoom-in on details. GENIE often results in sharper and higher contrast samples compared to DDIM.}
    \label{fig:app_additional_church}
\end{figure}

\begin{figure}
    \begin{subfigure}[b]{\textwidth}
        \centering
        \includegraphics[scale=0.95]{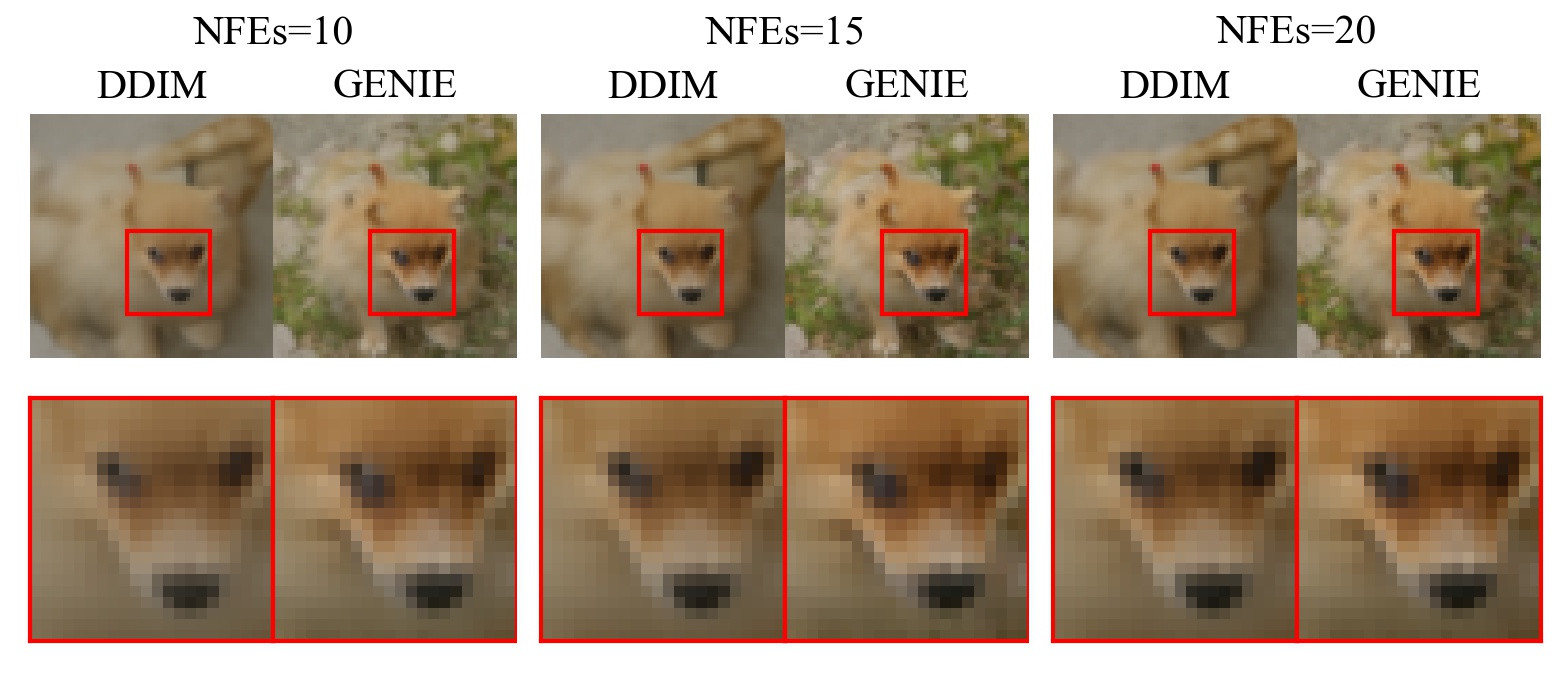}
    \end{subfigure}
    \begin{subfigure}[b]{\textwidth}
        \centering
        \includegraphics[scale=0.95]{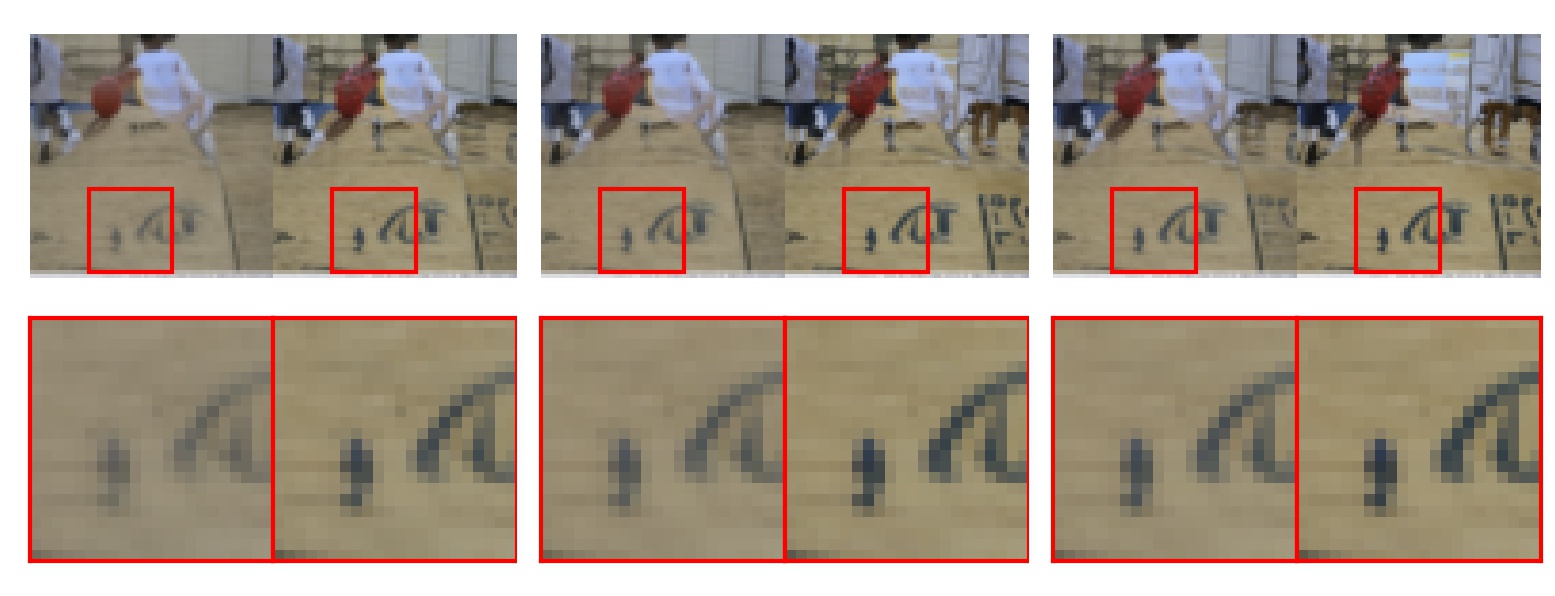}
    \end{subfigure}
    \begin{subfigure}[b]{\textwidth}
        \centering
        \includegraphics[scale=0.95]{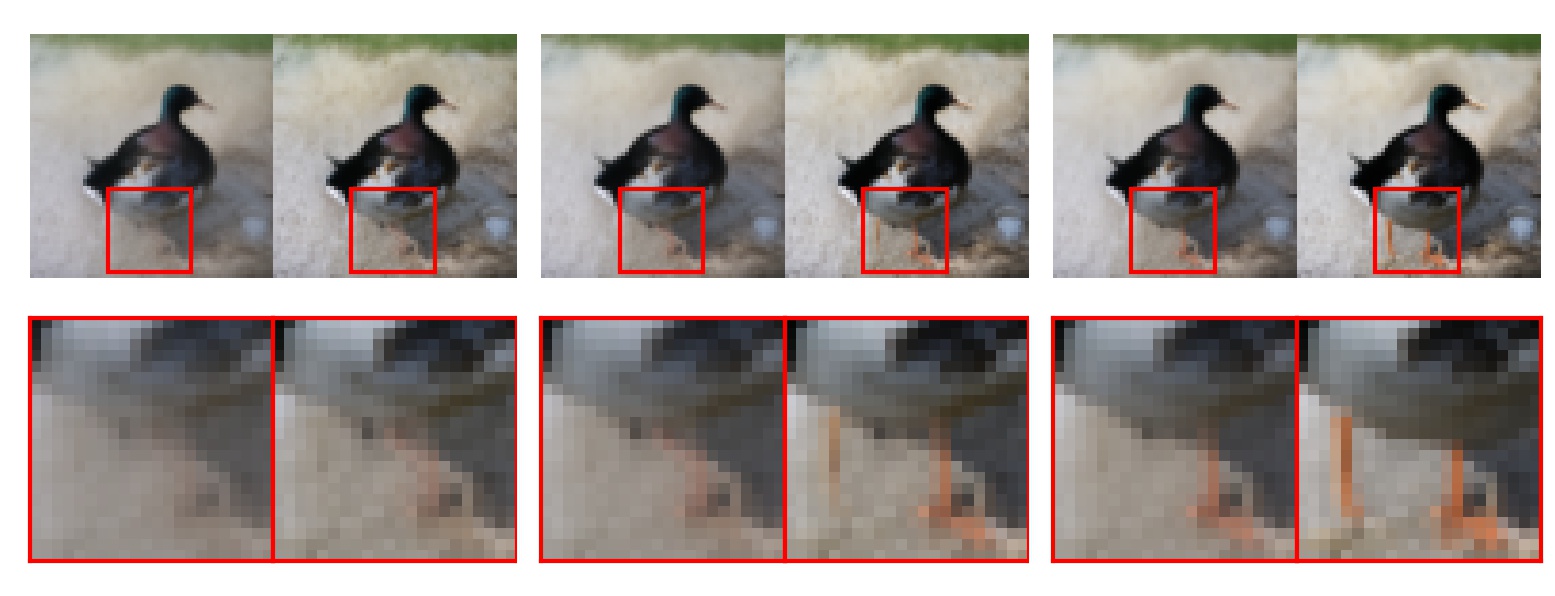}
    \end{subfigure}
    \begin{subfigure}[b]{\textwidth}
        \centering
        \includegraphics[scale=0.95]{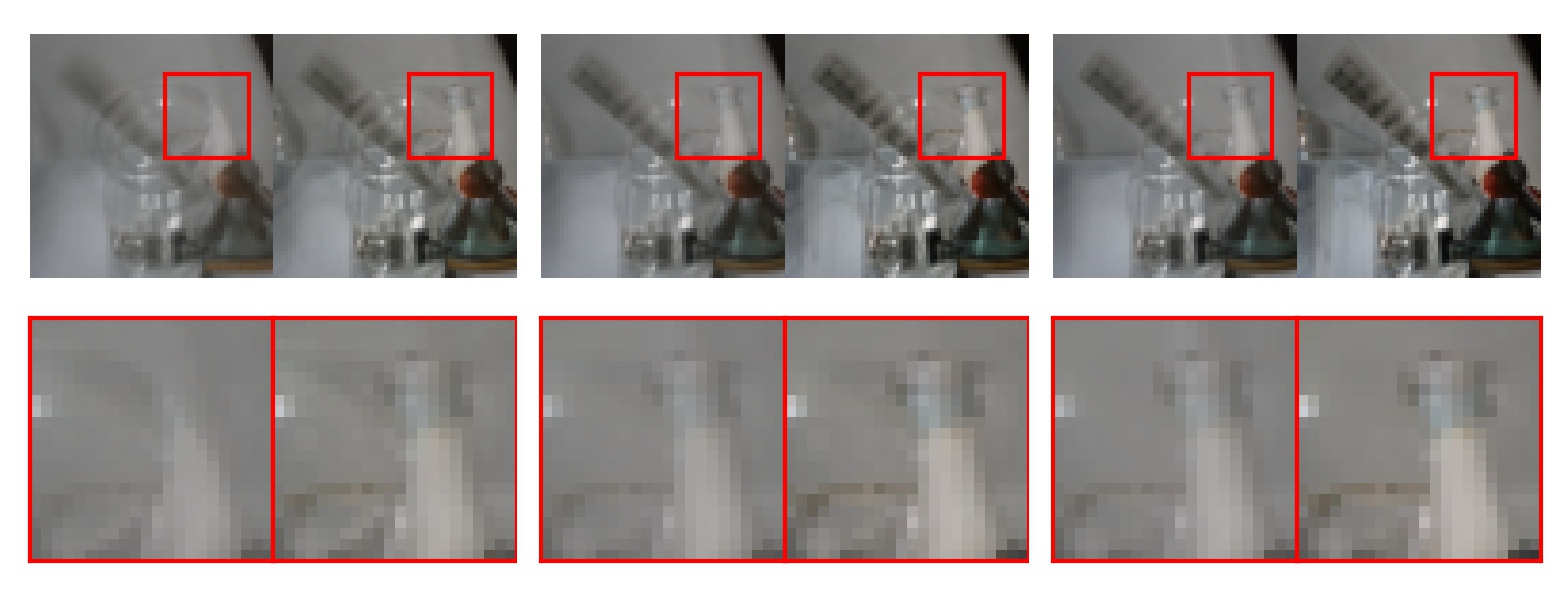}
    \end{subfigure}
    \caption{Additional samples on ImageNet with zoom-in on details. GENIE often results in sharper and higher contrast samples compared to DDIM.}
    \label{fig:app_additional_imagenet}
\end{figure}

\begin{figure}
    \begin{subfigure}[b]{\textwidth}
        \centering
        \includegraphics[scale=0.52]{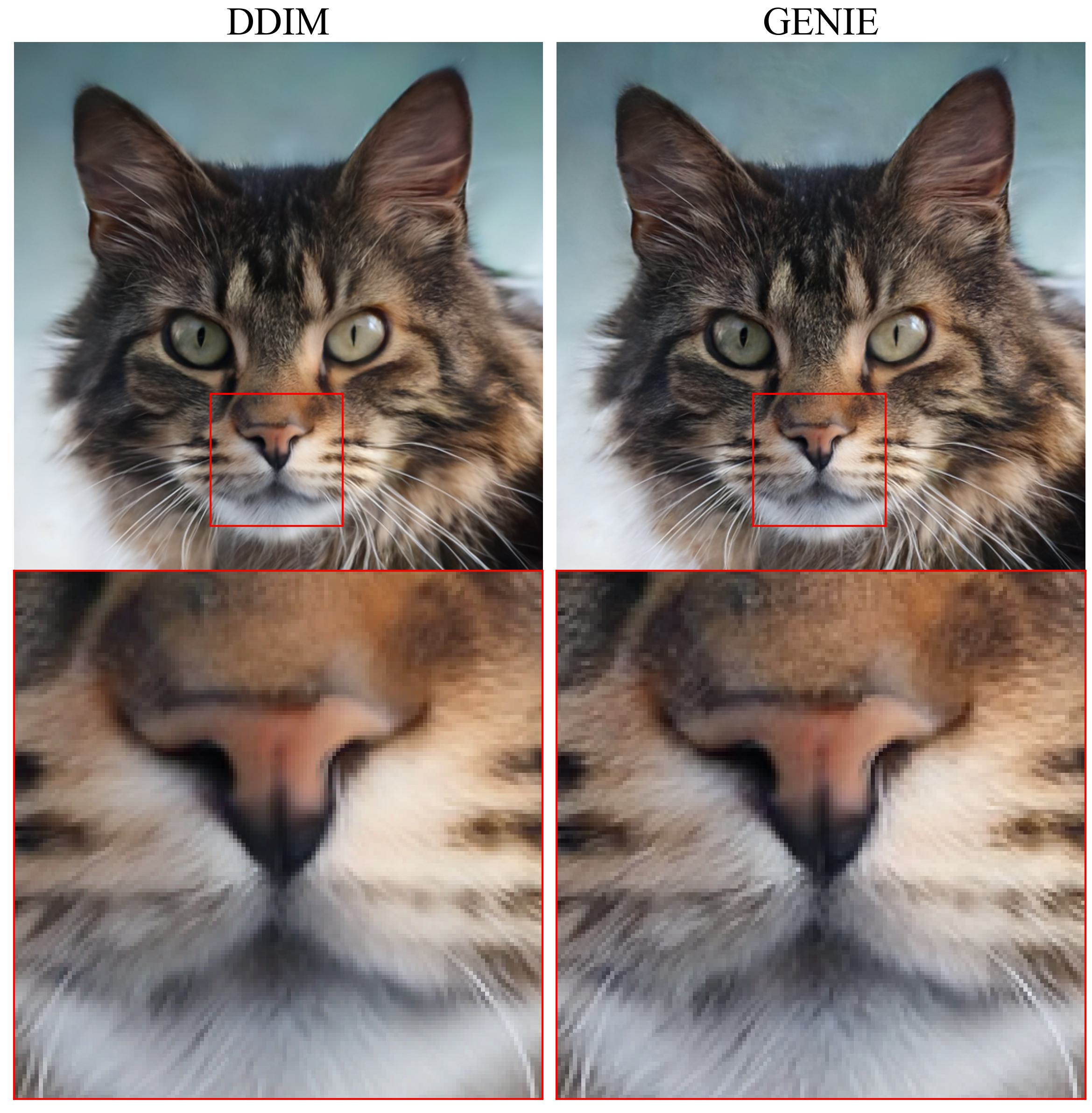}
    \end{subfigure}
    \begin{subfigure}[b]{\textwidth}
        \centering
        \includegraphics[scale=0.52]{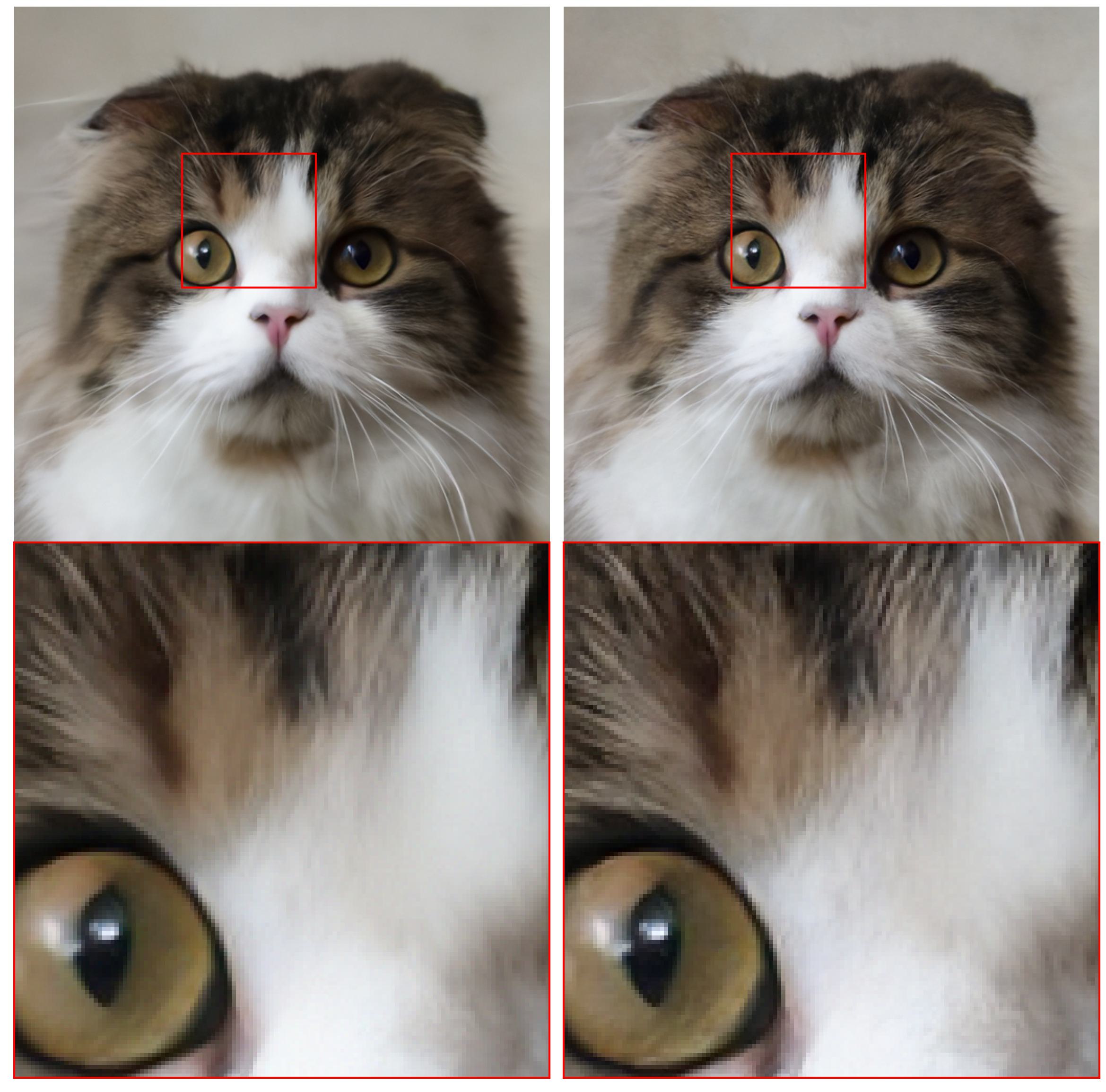}
    \end{subfigure}
    \caption{Additional samples on Cats with zoom-in on details. GENIE often results in sharper and higher contrast samples compared to DDIM.}
    \label{fig:app_additional_cats1}
\end{figure}

\begin{figure}
    \begin{subfigure}[b]{\textwidth}
        \centering
        \includegraphics[scale=0.52]{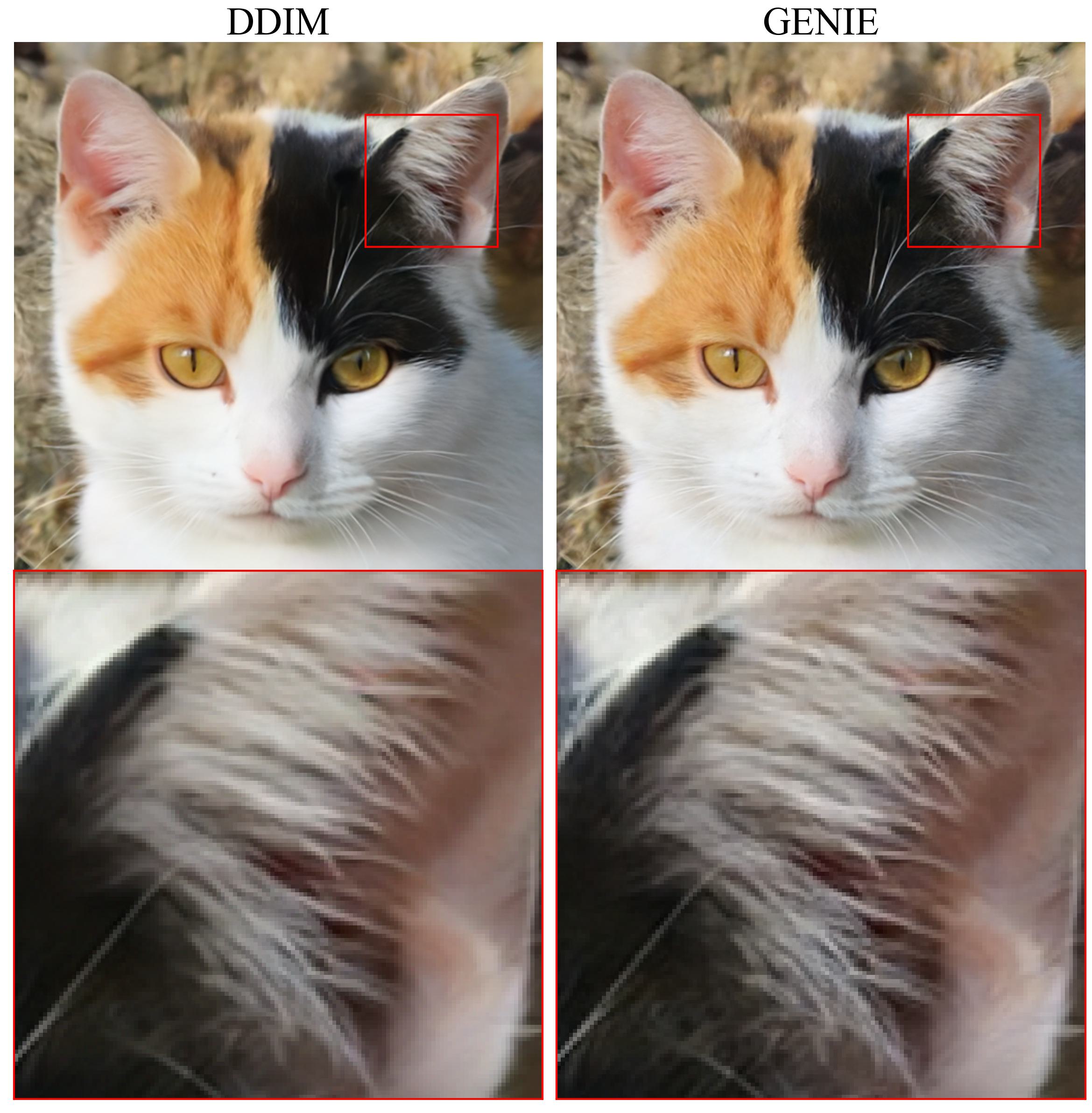}
    \end{subfigure}
    \begin{subfigure}[b]{\textwidth}
        \centering
        \includegraphics[scale=0.52]{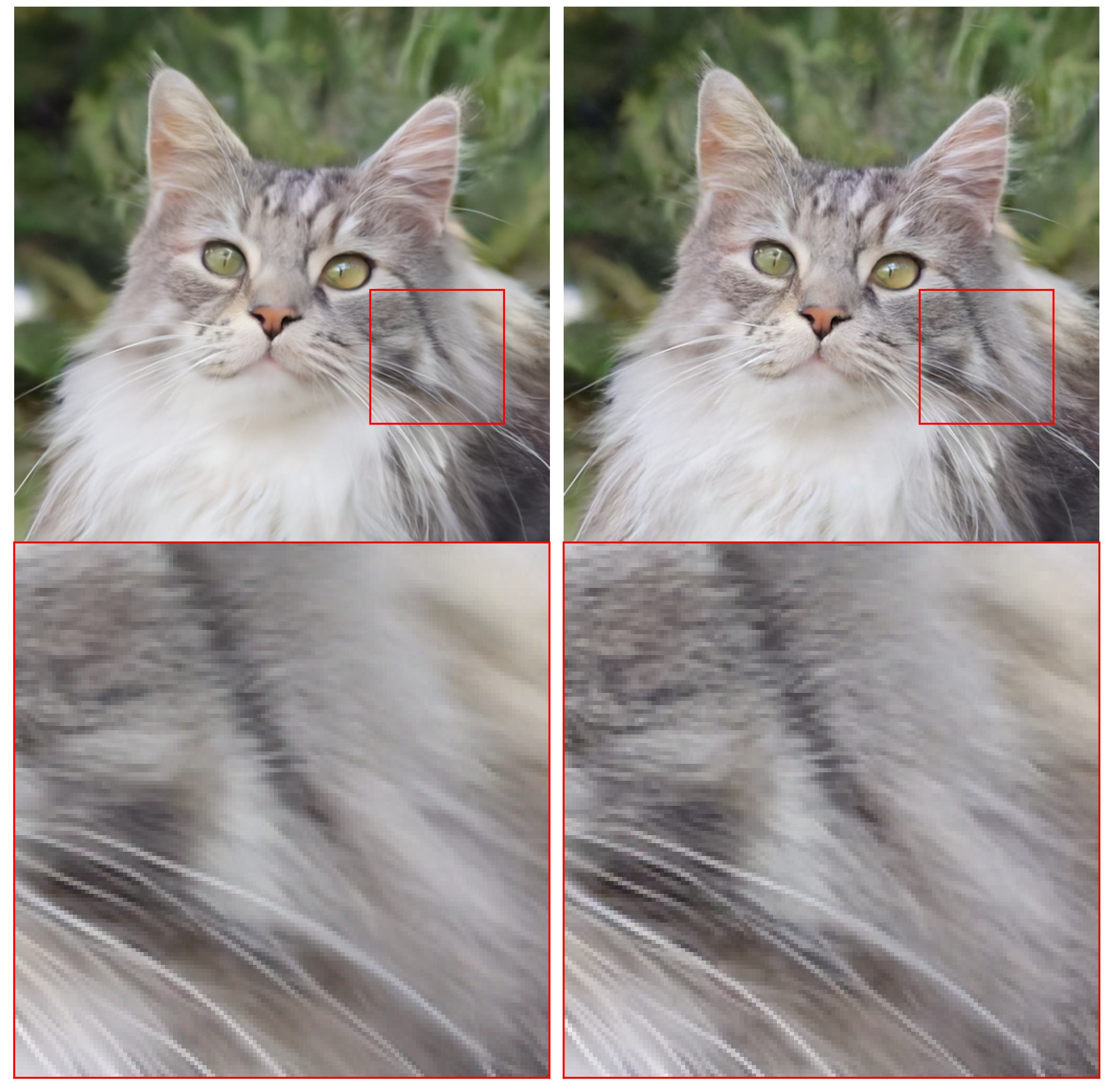}
    \end{subfigure}
    \caption{Additional samples on Cats with zoom-in on details. GENIE often results in sharper and higher contrast samples compared to DDIM.}
    \label{fig:app_additional_cats2}
\end{figure}

In this section, we show additional qualitative comparisons of DDIM and GENIE on LSUN Church-Outdoor (\Cref{fig:app_additional_church}), ImageNet (\Cref{fig:app_additional_imagenet}), and Cats (upsampler conditioned on test set images) (\Cref{fig:app_additional_cats1} and \Cref{fig:app_additional_cats2}). In all figures, we can see that samples generated with GENIE generally exhibit finer details as well as sharper contrast and are less blurry compared to standard DDIM.

\begin{figure}
    \centering
    \includegraphics[width=0.9\textwidth]{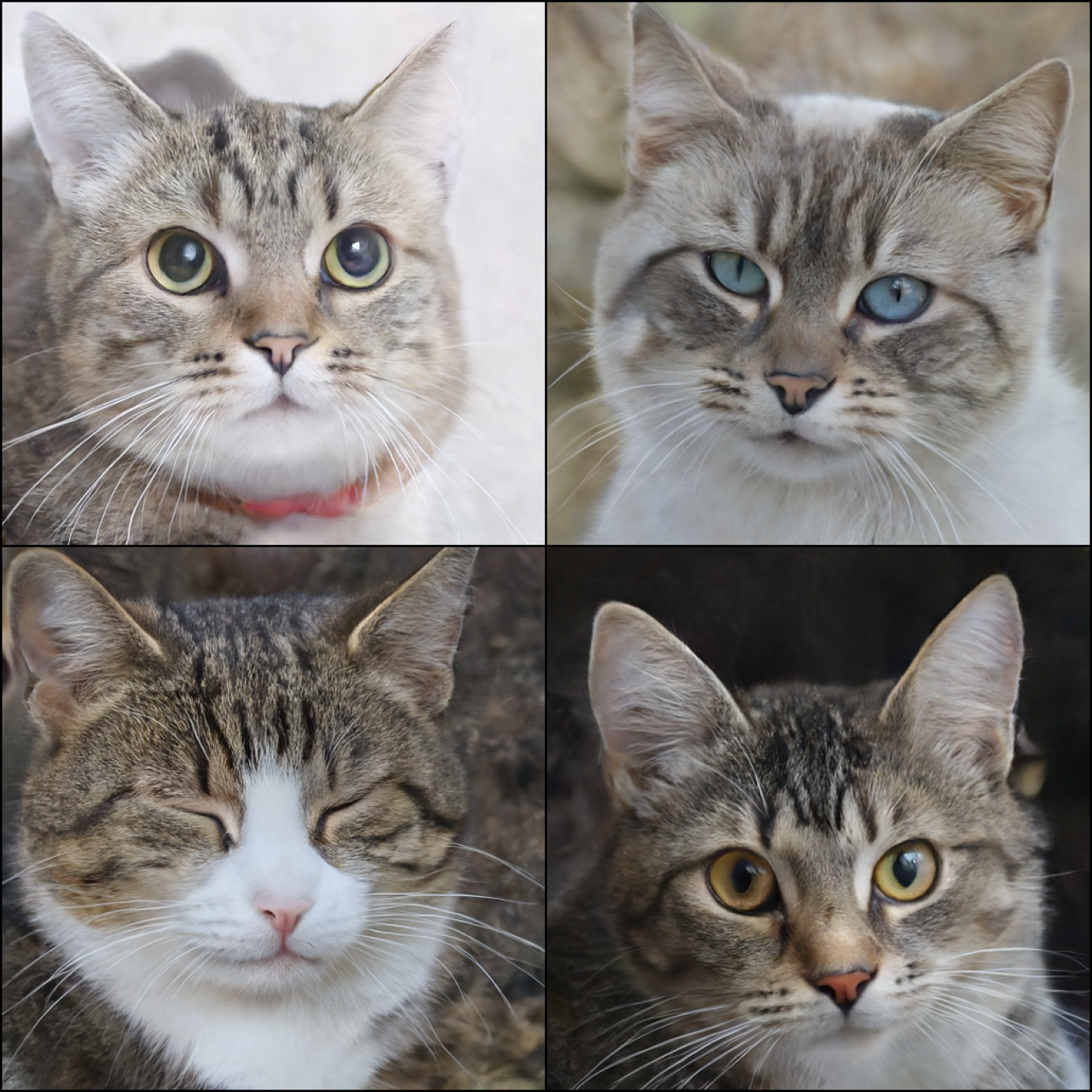}
    \caption{End-to-end samples on Cats. The GENIE base model uses 25 function evaluations and the GENIE upsampler only uses five function evaluations. An upsampler evaluation is roughly four times as expensive as a base model evaluation.}
    \label{fig:upsampling_end_to_end}
\end{figure}

\begin{figure}
    \centering
    \includegraphics[width=0.9\textwidth]{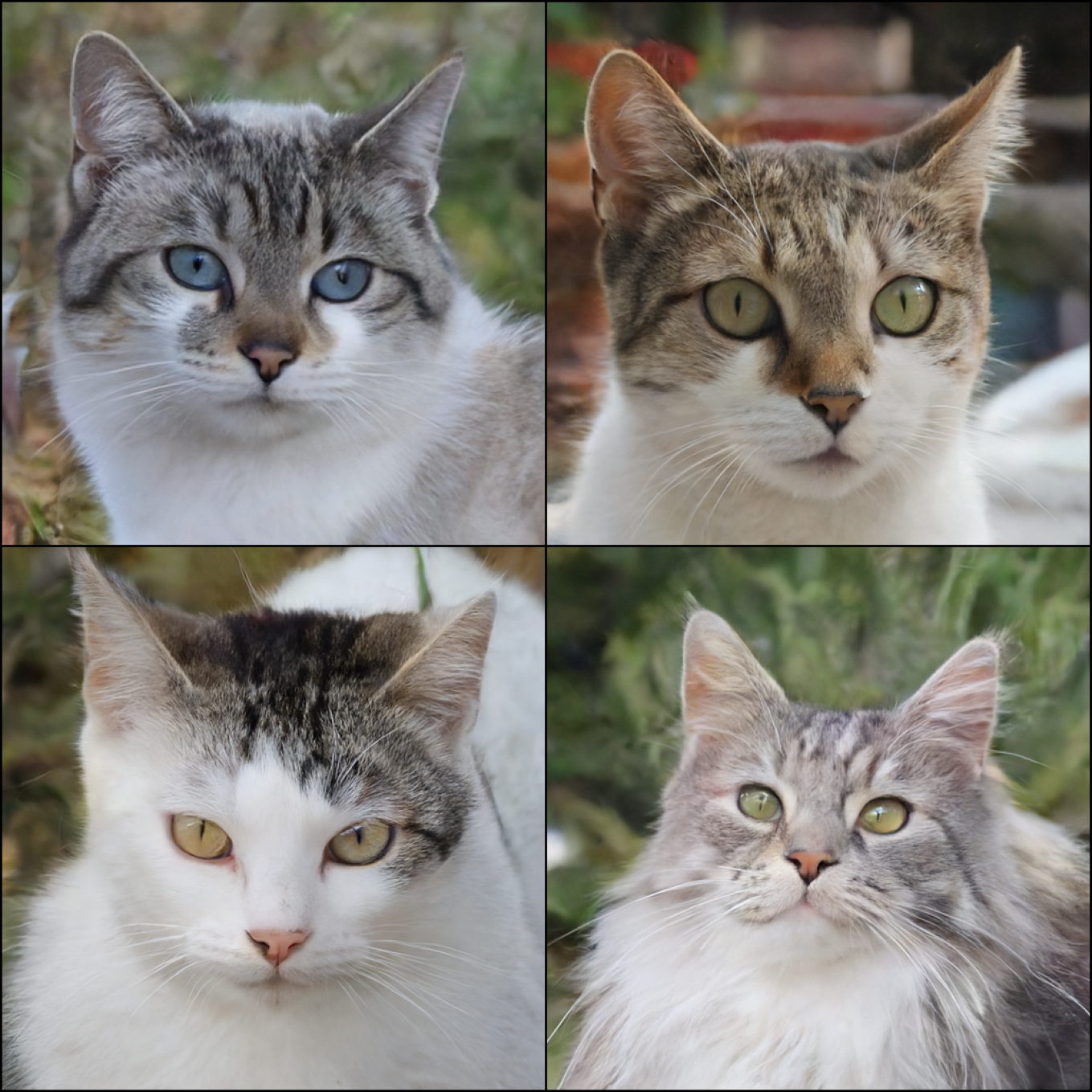}
    \caption{Upsampling $128 \times 128$ test set images using the GENIE upsampler with only five function evaluations.}
    \label{fig:upsampling_cats2}
\end{figure}

In~\Cref{fig:upsampling_end_to_end} and~\Cref{fig:upsampling_cats2}, we show 
additional high-resolution images generated with the GENIE Cats upsampler using base model samples and test set samples, respectively.

\subsection{Computational Resources} \label{s:app_comp_resources}
The total amount of compute used in this research project is roughly 163k GPU hours. We used an in-house GPU cluster of V100 NVIDIA GPUs.
\newpage
\section{Miscellaneous} \label{s:app_misc}

\subsection{Connection to~\texorpdfstring{\citet{bao2022estimating}}{}} \label{s:app_bao}
The concurrent \citet{bao2022estimating} learn covariance matrices for diffusion model sampling using prediction heads somewhat similar to the ones in GENIE. Specifically, both~\citet{bao2022estimating} and GENIE use small prediction heads that operate on top of the large first-order score predictor. However, we would like to stress multiple differences: (i) \citet{bao2022estimating} learn the DDM's sampling covariance matrices, while we learn higher-order ODE gradients. More generally, \citet{bao2022estimating} rely on stochastic diffusion model sampling, while we use the ODE formulation. (ii) Most importantly, in our case we can resort to directly learning the low-dimensional JVPs without low-rank or diagonal matrix approximations or other assumptions. Similar techniques are not directly applicable in~\citet{bao2022estimating}'s setting. In detail, this is because in their case the relevant matrices (obtained after Cholesky or another applicable decomposition of the covariance) do not act on regular vectors but random noise variables. In other words, instead of using a deterministic JVP predictor (which takes $\rvx_t$ and $t$ as inputs), as in GENIE, \citet{bao2022estimating} would require to model an entire distribution for each $\rvx_t$ and $t$ without explicitly forming high-dimensional Cholesky decomposition-based matrices, if they wanted to do something somewhat analogous to GENIE's novel JVP-based approach. As a consequence, \citet{bao2022estimating} take another route to keeping the dimensionality of the additional network outputs manageable in practice. In particular, they resort to assuming a diagonal covariance matrix in their experiments. By directly learning JVPs, we never have to rely on such potentially limiting assumptions. (iii) Experimentally, \citet{bao2022estimating} also consider fast sampling with few neural network calls. However, GENIE generally outperforms them (see, for example, their CIFAR10 results in their Table 2 for 10 and 25 NFE). This might indeed be due to the assumptions made by \citet{bao2022estimating}, which we avoid. Furthermore, their stochastic vs. our deterministic sampling may play a role, too.

\subsection{Combining GENIE with Progressive Distillation} \label{s:app_genie_x_pgd}
We speculate that GENIE could potentially be combined with Progressive Distillation~\citep{salimans2022progressive}: In every distillation stage of~\citep{salimans2022progressive}, one could quickly train a small GENIE prediction head to model higher-order ODE gradients. This would then allow for larger and/or more accurate steps, whose results represent the distillation target (teacher) in the progressive distillation protocol. This may also reduce the number of required distillation stages. Overall, this could potentially speed up the cumbersome stage-wise distillation and maybe also lead to an accuracy and performance improvement. In particular, we could replace the DDIM predictions in Algorithm 2 of~\citep{salimans2022progressive} with improved GENIE predictions. 

Note that this approach would not be possible with multistep methods as proposed by~\citet{liu2022pseudo}. Such techniques could not be used here, because they require the history of previous predictions, which are not available in the progressive distillation training scheme.

We leave exploration of this direction to future work.

\end{document}